\documentclass[twoside]{article}

\usepackage[accepted]{aistats2023arxiv}

\usepackage{natbib}

\usepackage[utf8]{inputenc} 
\usepackage[T1]{fontenc}    

\usepackage{xcolor}
\usepackage{hyperref}  
\hypersetup{
    colorlinks,
    linkcolor={red!50!black},
    citecolor={blue!50!black},
    urlcolor={blue!80!black}
}

\usepackage{url}            
\usepackage{booktabs}       
\usepackage{amsfonts}       
\usepackage{nicefrac}       
\usepackage{microtype}      
\usepackage{xcolor}         

\usepackage{microtype}
\usepackage{graphicx}
\usepackage{subfigure}

\usepackage{amsmath}
\usepackage{amssymb}
\usepackage{mathtools}
\usepackage{amsthm}

\newcommand{\E}{\mathbb{E}}

\newcommand{\va}{{\mathbf {a}}}

\newcommand{\vh}{{\mathbf {h}}}

\newcommand{\vw}{{\mathbf {w}}}

\usepackage{enumitem}

\usepackage{algorithm}
\usepackage{algorithmic}

\usepackage{graphicx}

\usepackage{amsthm}

\newtheorem{thm}{Theorem}

\usepackage{bm}
\DeclareMathAlphabet{\pazocal}{OMS}{zplm}{m}{n}
\newcommand{\Lb}{\pazocal{L}}

\newcommand\bo[1]{\textbf{ \normalfont{#1}}}
\usepackage{mathtools}
\DeclarePairedDelimiterX{\inp}[2]{\langle}{\rangle}{#1, #2}
\usepackage{tikz}
\newcommand*\circled[1]{\tikz[baseline=(char.base)]{\node[shape=circle,draw,inner sep=2pt] (char) {#1};}}
\newcommand{\norm}[1]{\left\lVert#1\right\rVert}
\usepackage{caption}

\usepackage[capitalize,noabbrev]{cleveref}

\theoremstyle{plain}
\newtheorem{theorem}{Theorem}[section]

\theoremstyle{definition}

\newtheorem{assumption}[theorem]{Assumption}
\theoremstyle{remark}

\usepackage[textsize=tiny]{todonotes}
\begin{document}

%

%

\twocolumn[

\aistatstitle{A Mini-Block Fisher Method for Deep Neural Networks}

\aistatsauthor{Achraf Bahamou \And Donald Goldfarb \And  Yi Ren }

\aistatsaddress{Department of Industrial Engineering and Operations Research \\ Columbia University } ]

\begin{abstract}
  Deep neural networks (DNNs) are currently predominantly trained using first-order methods. Some of these methods (e.g., Adam, AdaGrad, and RMSprop, and their variants) incorporate a small amount of curvature information by using a diagonal matrix to precondition the stochastic gradient. Recently, effective second-order methods, such as KFAC, K-BFGS, Shampoo, and TNT, have been developed for training DNNs, by preconditioning the stochastic gradient by layer-wise block-diagonal matrices. Here we propose a "mini-block Fisher (MBF)" "preconditioned gradient method, that lies in between these two classes of methods. Specifically, our method uses a block-diagonal approximation to the empirical Fisher matrix, where for each layer in the DNN, whether it is convolutional or feed-forward and fully connected, the associated diagonal block is itself block-diagonal and is composed of a large number of mini-blocks of modest size. Our novel approach utilizes the parallelism of GPUs to efficiently perform computations on the large number of matrices in each layer. Consequently, MBF’s per-iteration computational cost is only slightly higher than it is for first-order methods. The performance of our proposed method is compared to that of several baseline methods, on both autoencoder and CNN problems, to validate its effectiveness both in terms of time efficiency and generalization power. Finally, it is proved that an idealized version of MBF converges linearly. 
\end{abstract}

\section{Introduction}
First-order methods based on stochastic gradient descent (SGD) \citep{robbins1951stochastic}, and in particular, the class of adaptive learning rate methods, such as AdaGrad \citep{duchi2011adaptive}, RMSprop \citep{hinton2012neural}, and Adam \citep{kingma2014adam}, are currently the most widely used methods to train deep learning models (the recent paper \citep{schmidt2021descending} lists 65 methods that have “Adam” or “Ada”  as part of their names).  While these methods are easy to implement and have low computational complexity, they make use of only a limited amount of curvature information.  Standard SGD and its mini-batch variants, use none. SGD with momentum (SGD-m)  \citep{polyaksgd} and stochastic versions of Nesterov’s accelerated gradient method \citep{nesterov1998introductory}, implicitly make use of curvature by choosing step directions that combine the negative gradient with a scaled multiple of the previous step direction, very much like the classical conjugate gradient method. 

 To effectively optimize ill-conditioned functions, one usually needs to use second-order methods, which range from the Newton's method to those that use approximations to the Hessian matrix, such as BFGS quasi-Newton (QN) methods \citep{broyden1970convergence,fletcher1970new,goldfarb1970family,shanno1970conditioning}, including  limited memory variants \citep{liu1989limited}, and Gauss-Newton (GN) methods \citep{ortega1970iterative}. To handle large machine learning data sets,
 stochastic methods such as sub-sampled Newton 
 \citep{xu2019newton}), 
 QN
\citep{byrd2016stochastic,gower2016stochastic,wang2017stochastic}, GN, natural gradient (NG) 
\citep{amari2000adaptive}, Hessian-free 
\citep{martens2010deep}, and Krylov subspace methods \citep{vinyals2012krylov} have been developed. However, in all of these
methods, whether they use the Hessian or an approximation to it, the size of the matrix becomes prohibitive when the number of training parameters is huge.

Therefore, deep learning training methods have been proposed that use layer-wise block-diagonal approximations to the second-order preconditioning matrix. 
These include a Sherman-Morrison-Woodbury based variant \citep{ren2019efficient} and a low-rank variant \citep{topmomoute} of the block-diagonal Fisher matrix approximations for NG methods.   Also, Kronecker-factored matrix approximations of the diagonal blocks in Fisher matrices have been proposed to reduce the memory and computational requirements of NG methods, starting from KFAC for multilayer preceptrons (MLPs) \citep{martens2015optimizing}, which was extended to CNNs in \citep{grosse2016kronecker};
(in addition, see
\citet{heskes2000,povey2014parallel,george2018fast}).
Kronecker-factored QN methods \citep{goldfarb2020practical}, generalized GN methods \citep{botev2017practical}, an adaptive block learning rate method Shampoo \citep{gupta2018shampoo}  based on AdaGrad, and an approximate NG method TNT  \citep{ren2021tensor}, based on the assumption that the sampled tensor gradient follows a tensor-normal distribution have also been proposed.

\paragraph{Our Contributions:} We propose here a new {\it Mini-Block Fisher} (MBF) gradient method that lies in between adaptive first-order methods and block diagonal 
second-order methods. Specifically, MBF uses a block-diagonal approximation to the empirical Fisher matrix, where for each DNN layer, whether it is convolutional or feed-forward and fully-connected, the associated diagonal block is also block-diagonal and is composed of a large number of mini-blocks of modest size.

Crucially, MBF has comparable memory requirements to those of first-order methods, while its per-iteration time complexity is smaller, and in many cases, much smaller than that of popular second-order methods (e.g. KFAC) for training DNNs. Further, we prove convergence results for a variant of MBF under relatively mild conditions.

In numerical experiments on well-established Autoencoder and CNN models, MBF consistently outperformed state-of-the-art (SOTA) first-order methods (SGD-m and Adam) and performed favorably compared to popular second-order methods (KFAC and Shampoo).


\section{Notation and Definitions}
\textit{\textbf{Notation}}. 
$\text{Diag}_{i \in [L]}(A_i)$ is the block diagonal matrix with $\{A_1,...,A_L\}$ on its diagonal; $[L]:=\{1,...,L\}$; $\bm{X}=[x_1,...,x_n]^\top\in\mathbb{R}^{n\times d}$ is the input data;  $\lambda_{\min}(M), \lambda_{\max}(M)$ are the smallest and largest eigenvalues of the matrix $M$; $\otimes$ denotes the  Kronecker product;  $\Vert.\Vert_2$ denotes the Euclidean norm of a vector or matrix; and $\text{vec}(A)$ vectorizes $A$ by stacking its columns.

We consider a DNN with $L$ layers, defined by weight matrices ${W_l}$, for $l \in [L]$, that transforms the input vector $\bm{x}$ to an output $f(\bm{W}, \bm{x})$.
For a data-point $(x, y)$, the loss $\ell \left(f(\bm{W}, \bm{x}), y\right)$ between the output $f(\bm{W}, \bm{x})$ 
and $y$, is a non-convex function of
$\operatorname{vec}(\bm{W})^{\top}= \left[\operatorname{vec}\left(W_{1}\right)^{\top},...,\operatorname{vec}\left(W_{L}\right)^{\top}\right]
\in \mathbb{R}^p$,  containing all of the network’s parameters, 
and $\ell$ measures the accuracy of the prediction (e.g. squared error loss, cross entropy loss). 
The optimal parameters are obtained by minimizing the average loss $\Lb$ over the training set:
\begin{equation}
{\small
    \Lb(\bm{W}) = \frac{1}{n}\sum_{i=1}^n \ell (f(\bm{W}, \textbf{x}_i), \textbf{y}_i),
    \label{eq:loss_equation}
    }
\end{equation}
This setting is applicable to most common models in deep learning such as multilayer perceptrons (MLPs), CNNs, recurrent neural networks (RNNs), etc. In these models, the trainable parameter $W_l$ ($l = 1, \ldots,L$) come from the weights of a layer, whether it be a feed-forward, convolutional, recurrent, etc. For the weight matrix $W_l \in \mathbb{R}^{p_l}$  corresponding to layer $l$ and a subset of indices $b \subset \{1, \ldots, p_l\}$, we denote by $W_{l,b}$, the subset of parameters of $W_l$ corresponding to $b$.

The average gradient over a mini-batch of size $m$, $\bm{g}^{(m)} =  \frac{1}{m}\sum_{i=1}^{m} \frac{\partial\ell (f(\bm{W}, \textbf{x}_i), \textbf{y}_i)}{\partial \bm{W}}$, is  computed using standard back-propagation. In the full-batch case, where $m = n$, $\bm{g}^{(n)} = \bm{g} = \frac{\partial\Lb(\bm{W})}{\partial \bm{W}} = \mathcal{D} \bm{W} $. Here, we are using the notation $\mathcal{D} \bm{X} := \frac{\partial\Lb(\bm{W})}{\partial \bm{X}}$ for any subset of variables $\bm{X} \subset \bm{W}$.
 
\textbf{The Jacobian} $\bm{J}(\bm{W})$ of the loss $\Lb(\cdot)$ w.r.t the parameters $\bm{W}$ for a single output network is defined as $\bm{J} = [\bm{J}_1^{\top}, ..., \bm{J}_n^{\top}]^\top \in \mathbb{R}^{n \times p}$, where $\bm{J}_i^{\top}$ is the gradient of the loss w.r.t the parameters, i.e., $\bm{J}_i^{\top} = \operatorname{vec}(\frac{\partial\ell (f(\bm{W}, \textbf{x}_i), \textbf{y}_i)}{\partial \bm{W}})$. We use the notation ${\bm{J}_i^{X}}^\top = \operatorname{vec}(\frac{\partial\ell (f(\bm{W}, \textbf{x}_i), \textbf{y}_i)}{\partial X})$ and $\bm{J}^{X} = [{\bm{J}_1^{X}}^\top, ..., {\bm{J}_n^{X}}^\top]^\top$ for any subset of variables $X$ of $\bm{W}$.

 \textbf{The Fisher matrix} $\bm{F}(\bm{W})$ of the model's conditional distribution is defined as 
\begin{align*}
\bm{F}(\bm{W}) = \underset{{\substack{x \sim Q_x \\ y \sim p_{\bm{W}}(\cdot|x)}}}{\E} \left[ \frac{\partial \log p_{\bm{W}}(y | x)}{\partial \bm{W}}  \left( \frac{\partial \log p_{\bm{W}}(y | x)}{\partial \bm{W}} \right)^\top \right],
\label{eq_8}
\end{align*}
where $Q_x$ is the data distribution of $x$ and $p_{\bm{W}}(\cdot|x)$ is the density function of the conditional distribution defined by the model with a given input $x$. As shown in \citep{martens2020new}, $\bm{F}(\bm{W})$ is equivalent to the Generalized Gauss-Newton (GGN) matrix if the conditional distribution is in the exponential family, e.g., a categorical distribution for classification or a Gaussian distribution for regression.

\textbf{The empirical Fisher matrix (EFM)} $\tilde{\bm{F}}(\bm{W})$ defined as:
\begin{align*}
\tilde{\bm{F}} (\bm{W}) &= {\frac{1}{n}\sum_{i=1}^n \frac{\partial\ell (f(\bm{W}, \textbf{x}_i), \textbf{y}_i)}{\partial \bm{W}} \frac{\partial\ell (f(\bm{W}, \textbf{x}_i), \textbf{y}_i)}{\partial \bm{W}}^\top} \\
&= \frac{1}{n} \bm{J}(\bm{W})^\top \bm{J}(\bm{W}),
\end{align*}
is obtained by replacing the expectation over the model's distribution in  $\bm{F}(\bm{W})$ by an average over the empirical data. MBF uses
the EMF rather than the Fisher matrix, since 
doing so does not require extra
backward passes to compute additional gradients and memory to store them. We note that, as discussed in \citep{Kunstner2019} and \citep{thomas20a}, the EMF, which is an un-centered second moment of the gradient, captures less curvature information than the Fisher matrix, which coincides with the GGN matrix in many important cases, and hence is closely related to  $\nabla^2\Lb(\bm{W})$.
 To simplify notation we will henceforth drop the "tilde" $\tilde{}$ and denote the EFM by $\bm{F}$. We denote by $\bm{F}^{X} = \frac{1}{n} (\bm{J}^{X})^\top \bm{J}^{X}$, the sub-block of $\bm{F}(\bm{W})$ associated with any subset of variables $X \subset \bm{W}$, and write $(\bm{F}^{X})^{-1}$
 as $F^{-1}_{X}$.


\section{Mini-block Fisher (MBF) method}
\label{sec_1}

At each iteration, MBF preconditions the gradient direction by the inverse of a damped EFM:
\begin{equation}
{\small
    \bm{W}(k+1) = \bm{W}(k) - \alpha \left(\bm{F}(\bm{W}(k))  + \lambda \bm{I}\right)^{-1} \bm{g}(k),
    \label{eq:update_rule_fisher}
    }
\end{equation} 
where $\alpha$ is the learning rate and $\lambda$ is the damping parameter.

\begin{figure}[h]
    \centering
    \begin{minipage}{0.5\textwidth}
        \centering
        \includegraphics[width=\textwidth]{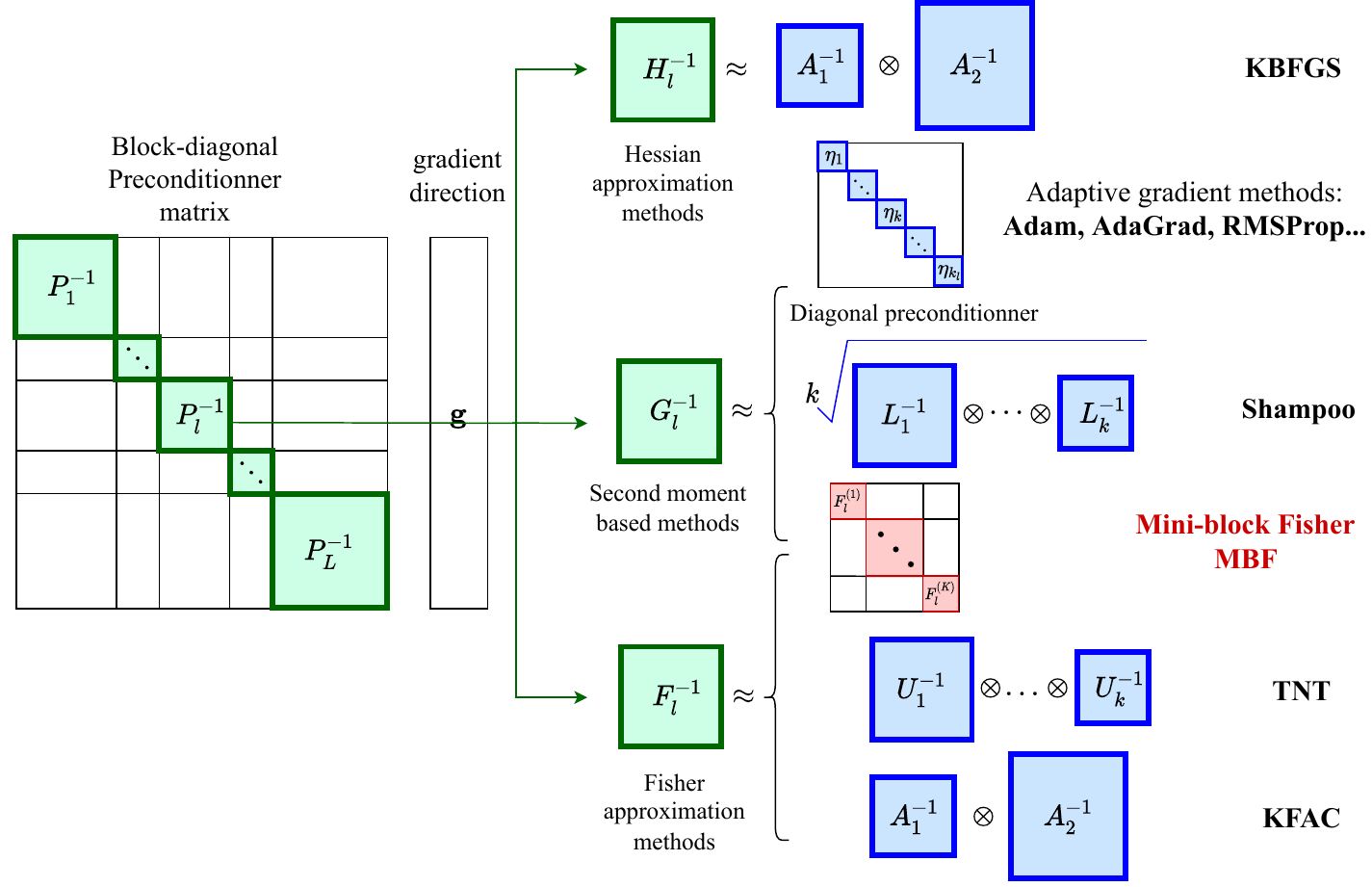}
    \end{minipage}
    \caption{
    MBF vs other block-diagonal preconditioned gradient methods
    }
    \label{precond_illustration}
\end{figure}

To avoid the work of computing and storing the inverse of the $p \times p$ damped EFM, $(\bm{F}+\lambda I)^{-1}$, where $p$ can be in the millions, we assume, as in KFAC and Shampoo, that the EFM has a block diagonal structure, where the $l_{th}$ diagonal block corresponds to the second moment of the gradient of the model w.r.t to the weights in the $l_{th}$ layer. Hence, the block-diagonal EFM is:
$$
\bm{F}(\bm{W}) \approx \text{Diag}\left( \bm{F}^{W_1}, ..., \bm{F}^{W_L}\right).
$$
Figure \ref{precond_illustration} summarizes how several existing methods further approximate these diagonal blocks.
 
MBF further approximates each of the diagonal blocks 
$F_{W_l}$
by a block-diagonal matrix, composed of a typically large number mini-blocks, depending on the nature of layer $l$, as follows:

\begin{figure}
    \centering
    \begin{minipage}{0.49\textwidth}
        \centering
        \includegraphics[width=\textwidth]{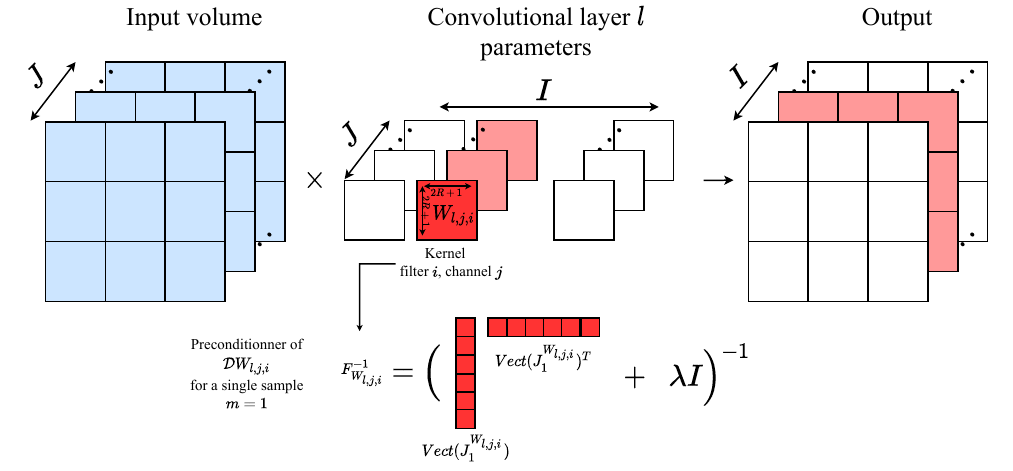} 
        \caption{Illustration of MBF's approximation for a convolutional layer.}
         \label{precond_conv_illustration}
    \end{minipage}\hfill
    \begin{minipage}{0.49\textwidth}
        \centering
        \includegraphics[width=0.8\textwidth]{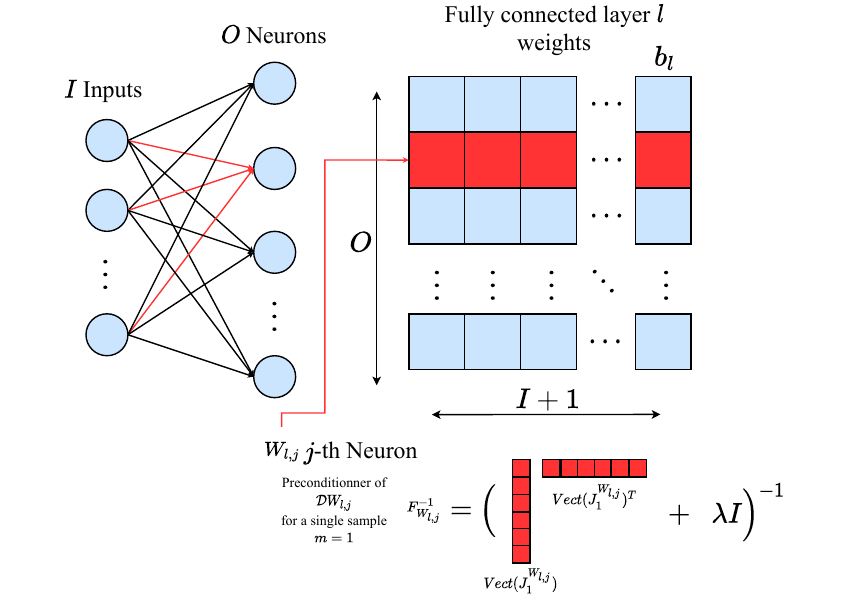} 
        \caption{Illustration of MBF's preconditionner for a feed-forward  fully-connected layer.}
        \label{precond_fcc_illustration}
    \end{minipage}
\end{figure}

\textbf{Layer $l$ is convolutionnal:} For simplicity, we assume that the convolutional layer $l$ is 2-dimensional and has $J$ input channels indexed by $j = 1, ..., J$, and $I$ output channels indexed by $i = 1, ..., I$; there are $I \times O$ kernels $W_{l, j, i}$, each of size $(2R+1) \times (2R+1)$, with spatial offsets from the centers of each filter indexed by $\delta \in {\Delta} := \{ -R, ..., R \} \times \{ -R, ..., R \}$; the stride is of length 1, and the padding is equal to $R$, so that the sets of input and output spatial locations ($t \in \mathcal{T} \subset \mathbf{R}^2$) are the same.\footnote{The derivations in this paper can also be extended to the case where the stride is greater than 1.}. For such layers, we use the following $(IJ+1) \times (IJ+1)$ block-diagonal approximation to the $l_{th}$ diagonal block $
\bm{F}^{W_l}$ of the Fisher matrix 
$$
\text{diag}\{ \bm{F}^{W_{l, 1,1}},...,
\bm{F}^{W_{l, 1,I}},...,
\bm{F}^{W_{l, J,1}},...,
\bm{F}^{W_{l, J,I}}, \bm{F}^{b_{l}}\},
$$
where each of the $IJ$ diagonal blocks  $\bm{F}^{W_{l, j,i}}$ is a $|\Delta| \times |\Delta|$ symmetric matrix corresponding to the kernel vector $W_{l, j,i}$ and where 
$\bm{F}^{b_{l}}$ is an $I \times I$ diagonal matrix corresponding the bias vector $b_l$. Therefore, the preconditioning matrix $F^{-1}_{W_{l, j,i}}$ corresponding to the kernel for input-output channel pair $(j,i)$ is given by:
$$
 F^{-1}_{W_{l, j,i}} := \left(\frac{1}{n} (\bm{J}^{W_{l, j,i}})^T\bm{J}^{W_{l, j,i}} + \lambda I \right)^{-1}
$$
A common choice in CNNs is to use  either a $3\times3$ or $5\times5$ kernel for all of the $IJ$ channel pairs in a layer. Therefore, all of these matrices are of the same (small) size, $|\Delta| \times |\Delta|$, and can be inverted efficiently by utilizing the parallelism of GPUs.
We illustrate MBF's approximation for a convolutional layer for the case of one data-point in Fig. 2.
From Fig. 2, it is apparent that kernal matrices in a convolutional layer that connect input to output channels are analagous to scalar weights that connect input to output nodes in a ff-cc layer.  Hence, MBF is analaous to using the squares of the components of the gradient in a ff-cc network, and hence is analagous to a "squared" version of an adaptive first-order method.  This observation (detailed in Appendix \ref{apx_cnn_motivation}) was in fact the motivation for our development of the MBF approach. 

\textbf{Layer $l$ is feed forward and fully connected
(ff-fc):} For a ff-fc layer with $I$ inputs and $O$ outputs, we use the following $O \times O$ block-diagonal approximation to the Fisher matrix 
$$ 
\bm{F}^{W_l} \approx \text{diag} \{ \bm{F}^{W_{l, 1}}, \ldots,\bm{F}^{W_{l, O}}\},
$$
whose $j_{th}$ diagonal block $\bm{F}^{W_{l, j}}$ is an $(I+1)\times(I+1)$ symmetric matrix corresponding to the  vector $W_{l, j}$ of $I$ weights from all of the input neurons and the bias to the 
$j_{th}$ output neuron. Therefore, the preconditioning matrix $F^{-1}_{W_{l, j}}$ corresponding to the $j_{th}$ output neuron is given by:
$$
 F^{-1}_{W_{l, j}} := \left(\frac{1}{n} (\bm{J}^{W_{l, j}})^T\bm{J}^{W_{l, j}} + \lambda I \right)^{-1}
$$
Our choice of such a mini-block subdivision was motivated by the findings presented in \citep{topmomoute}, first derived in \citep{phdthesis}, where it was  shown that the Hessian of a neural network with one hidden layer with cross-entropy loss converges during optimization to a block-diagonal matrix, where the diagonal blocks correspond to the weights linking all the input units to one hidden unit and all of the hidden units to one output unit. 

\begin{figure}
    \centering
    \begin{minipage}{0.49\textwidth}
        \centering
        \subfigure[\footnotesize First CNN layer]{\includegraphics[width=0.41\textwidth]{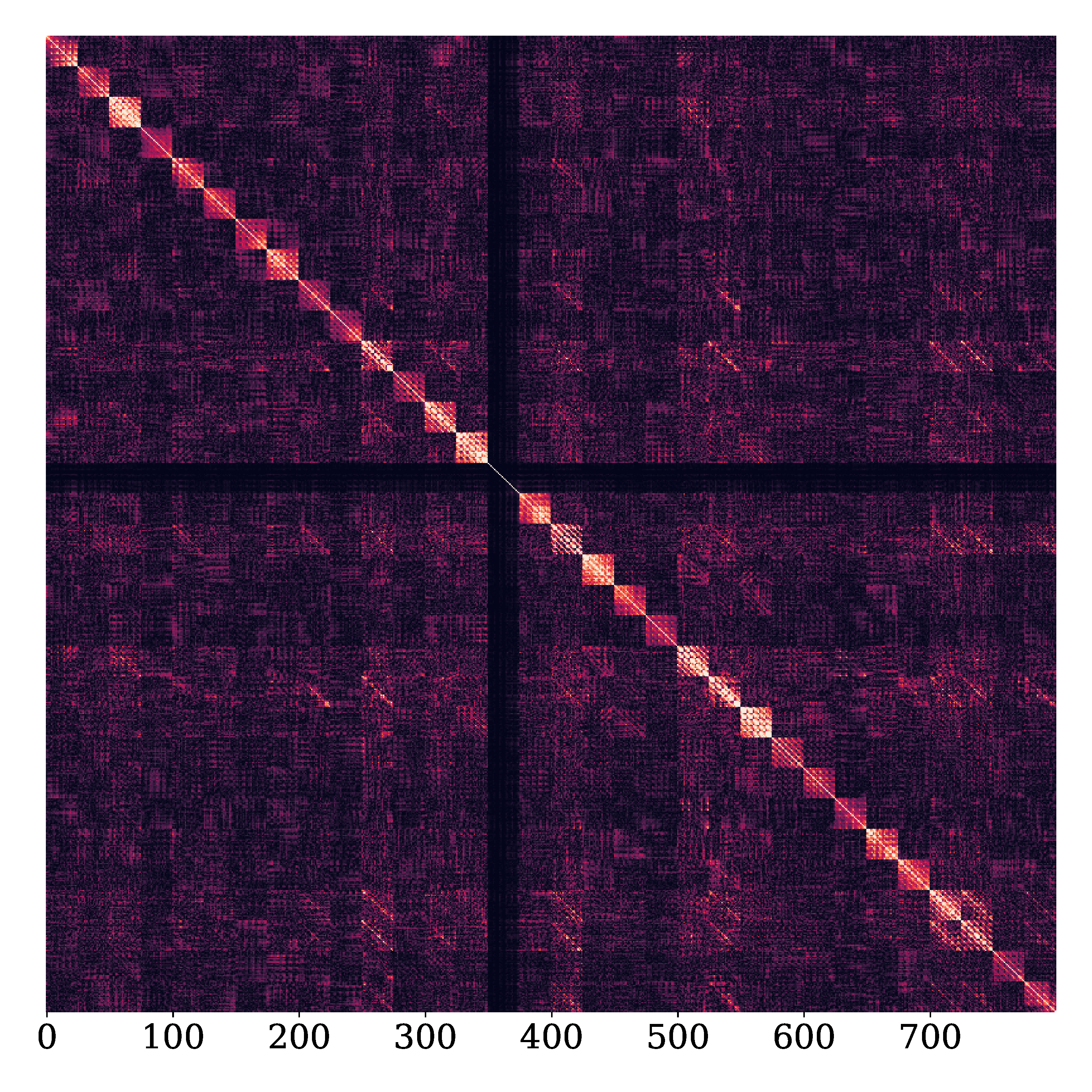}}\quad
  \subfigure[\footnotesize Zoom on first 10 blocks]{\includegraphics[width=0.49\textwidth]{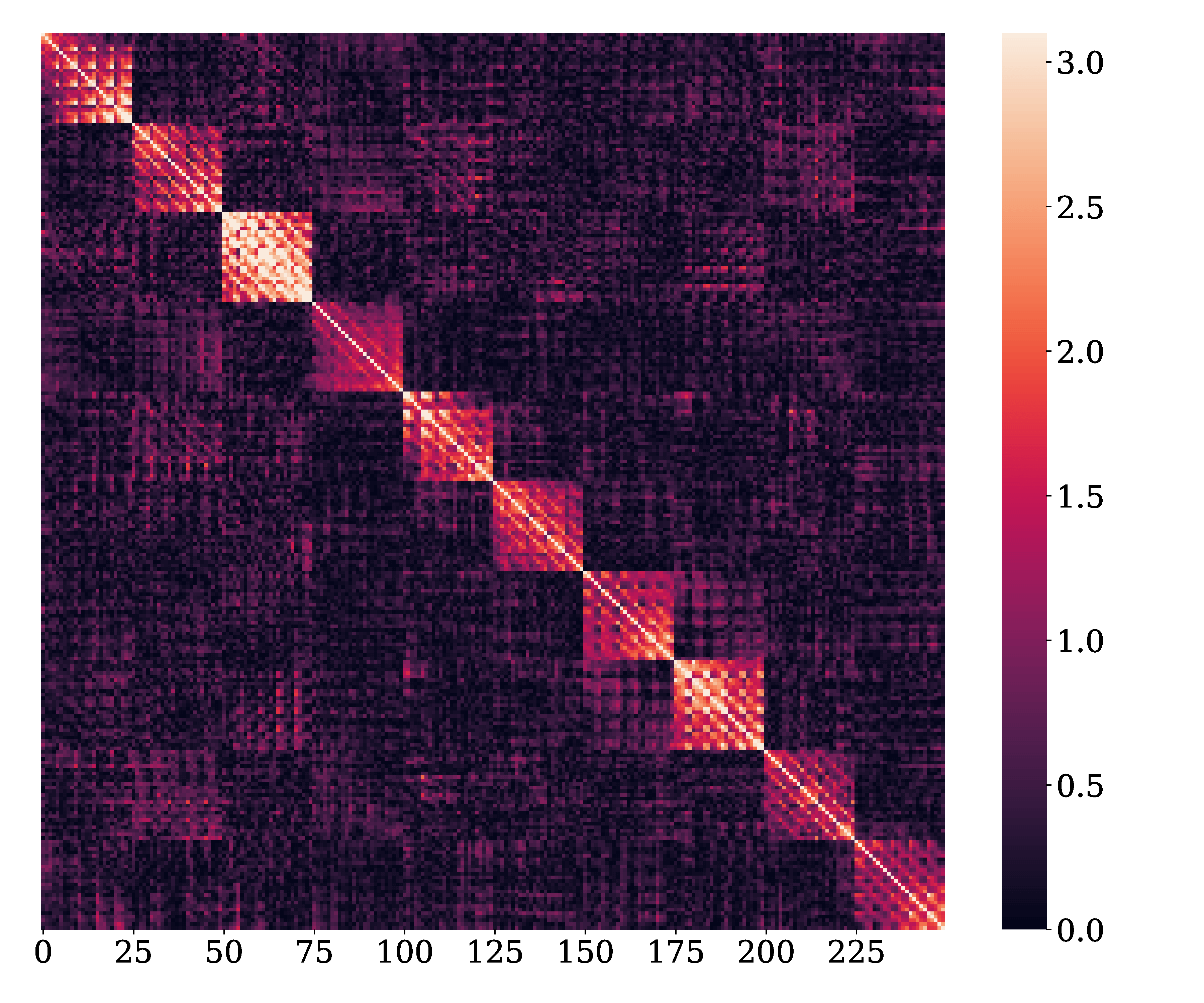}}
        \caption{Absolute EFM inverse after 10 epochs for the first convolutional layer of the Simple CNN network that uses $32$ filters of size $5 \times 5$.}
         \label{heatmap_cnn}
    \end{minipage}\hfill
    \begin{minipage}{0.49\textwidth}
        \centering
  \subfigure[\footnotesize Last layer]{\includegraphics[width=0.473\textwidth]{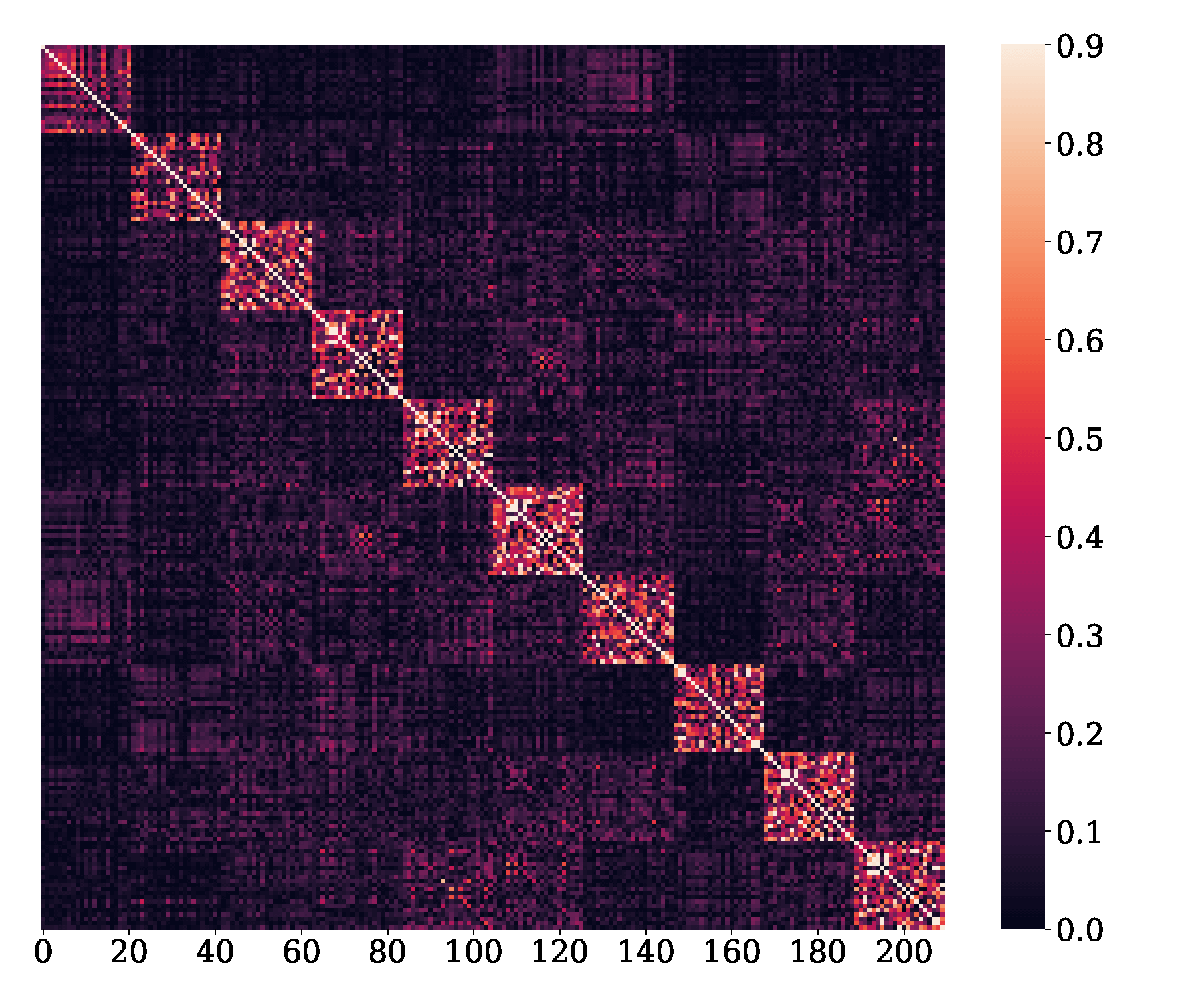}}\quad
  \subfigure[\footnotesize Middle layer]{\includegraphics[width=0.473\textwidth]{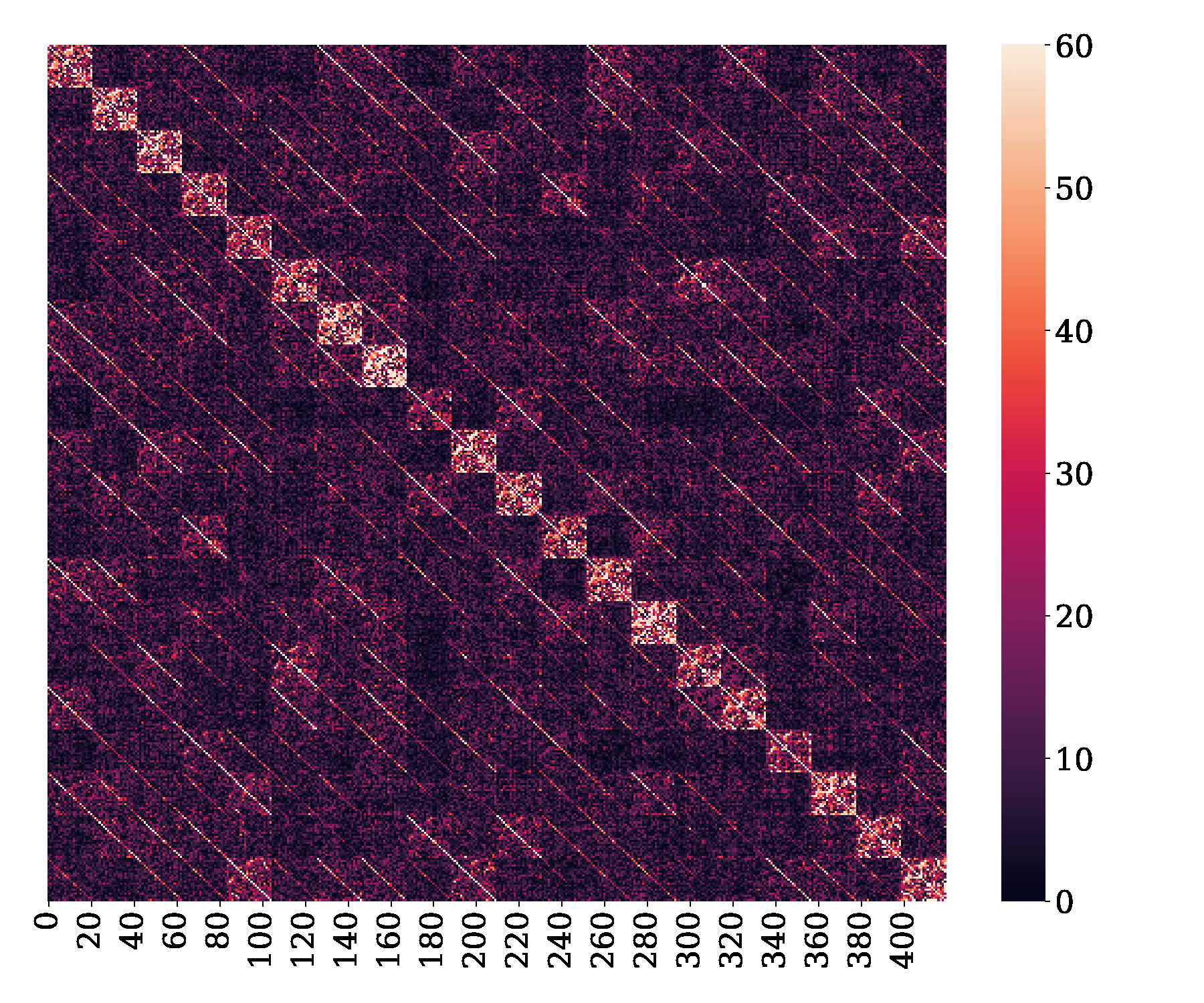}}
  \caption{Absolute EFM inverse after 50 epochs of the last and middle layers (including bias) of a small FCC-NN.}
        \label{heatmap_fcc}
    \end{minipage}
\end{figure}

This suggests that a similar block-diagonal structure applies to the Fisher matrix in the limit of a sequence of iterates produced by an optimization algorithm. The latter suggestion was indeed confirmed by findings presented in \citep{amari2018unitwise}, where the authors proved that a "unit-wise" block diagonal approximation to the Fisher information matrix is close to the full matrix 
modulo off-diagonal blocks of small magnitude, 
which provides a justification for the quasi-diagonal natural gradient method proposed in \citep{ollivier2015unitwise} and our mini-block approximation in the case of fully connected layers.
Finally, since the $O$ matrices $\bm{F}^{W_{l, j}}$, for $j = 1, \dots,O$, are all of the same size, $(I+1)\times(I+1)$, they can be inverted efficiently by utilizing the parallelism of GPUs. We illustrate MBF's approximation for a fully connected layer for the case of one data-point in Figure \ref{heatmap_fcc} for a $7$-layer (256-20-20-20-20-20-10) feed-forward DNN using $\tanh$ activations, partially trained to classify a $16 \times 16$ down-scaled version of MNIST as in \citep{martens2015optimizing}.

\begin{algorithm}[ht]
    \caption{Generic MBF training algorithm}
    \label{algo_8}
    \begin{algorithmic}[1]
    \REQUIRE Given learning rates $\{ \alpha_k \}$, damping value $\lambda$, batch size $m$
    \FOR{$k = 1, 2, ...$}
    \STATE Sample mini-batch $M$ of size $m$
    \STATE Perform a forward-backward pass over $M$ to compute stochastic gradient $\mathcal{D} W_l$ ($l = 1, ..., L$)
    \FOR{$l = 1, ..., L$}
        \FOR{mini-block $b$ in layer $l$, \textbf{in parallel}}
            \STATE $F^{-1}_{W_{l, b}} := \left(\frac{1}{m} (\bm{J}^{W_{l, b}})^T\bm{J}^{W_{l, b}} + \lambda I \right)^{-1}$
            \STATE $W_{l, b} = W_{l, b}- \alpha_k F^{-1}_{W_{l, b}} \mathcal{D} W_{l, b}$
        \ENDFOR
    \ENDFOR
    \ENDFOR
    \end{algorithmic}
\end{algorithm}

Algorithm \ref{algo_8} below gives the pseudo-code for a generic 
version of MBF.
Since updating the Fisher mini-blocks is time consuming in practice as it requires storing and computing the individual gradients, we propose in Section 7 below, a practical approach for approximating these matrices. 
However, we first present empirical results that justify and motivate both the kernel-based  and the all-to-one mini-block subdivisions described above for convolutional and ff-fc layers, respectively, followed by a discussion of the linear convergence of an idealized version of the generic MBF algorithm.

After deriving our MBF method, we became aware of the paper \citep{anil2021scalable}, which proposes using sub-layer block-diagonal preconditioning matrices for Shampoo, a tensor based DNN training method. Specifically, it considers two cases: partitioning (i) very large individual ff-fc  matrices  (illustrating this for a matrix of size $[2^9 \times 2^{11}]$ into either a $1 \times 2$ or a  $2 \times 2$ block matrix with blocks all of the same size) and (ii) ResNet-50 layer-wise matrices into sub-layer blocks of size 128. 
However, \citep{anil2021scalable} does not propose a precise method for
using mini-blocks as does MBF.

\paragraph{Motivation for MBF:}Our choice of mini-blocks for both the convolutional and ff-fc layers was motivated by the observation that most of the weight in the EFM inverse resides in diagonal blocks, and in particular in the mini-blocks described above. More specifically, to illustrate this observation for convolutional layers, we trained a simple
convolutional neural network, Simple CNN, (see Appendix \ref{apx_more_motivation} for more details) on Fashion MNIST \citep{xiao2017fashion}. Figure \ref{heatmap_cnn} shows the heatmap of the absolute value of the EFM inverse corresponding to the first convolutional layer, which uses $32$ filters of size $5 \times 5$ (thus 32 mini-blocks of size $25 \times 25$ ). One can see that the mini-block (by filter) diagonal approximation is reasonable. Figures for the 2nd convolutional layer are included in the Appendix \ref{apx_more_motivation}.
Since the ff-fc layers in the Simple-CNN model result in an EFM for those layers that is too large to work with, we chose to illustrate the mini-block structure of the EFM on a standard DNN, partially trained to classify a $16 \times 16$ down-scaled version of MNIST that was also used in \citep{martens2015optimizing}. Figure \ref{heatmap_fcc} shows the heatmap of the absolute value of the EFM inverse for the last and middle fully connected layers (including bias). One can see that the mini-block (by neuron) diagonal approximation is reasonable. A larger figure for the second fully-connected layer is included in Appendix \ref{apx_more_motivation}).

\paragraph{Comparison: directions of MBF and other methods vs. full block-diagonal EFM:}To explore how close is MBF's direction to the one obtained by a block-diagonal full EFM method (BDF), where each block corresponds to one layer's full EFM in the model, we computed the cosine similarity between these two directions. We also included SOTA first-order (SGD-m, Adam) and second-order (KFAC, Shampoo) methods for reference. The algorithms were run on a $16 \times 16$ down-scaled MNIST \citep{lecun2010mnist} dataset and a small feed-forward NN with layer widths 256-20-20-20-20-20-10 described in \citep{martens2015optimizing}. As in \citep{martens2015optimizing}, we only show the middle four layers. For all methods, we followed the trajectory obtained using the BDF method. In our implementation of the BDF method, both the gradient and the block-EMF matrices were estimated with a moving-average scheme, with the decay factors being 0.9. Note that MBF-True refers to the version of MBF in which, similarly to KFAC, the mini-block Fisher is computed by drawing one label from the model distribution for each input image as opposed to MBF, where we use the average over the empirical data. For a more detailed comparison on Autoencoder and CNN problems, see Appendix \ref{apx_exp}.

As shown in Figure \ref{fig_8}, the cosine similarity between the MBF and MBF-True and the BDF direction falls on most iterations between 0.6 to 0.7 for all four layers and not surprisingly, falls midway between the SOTA first-order and block-diagonal second order methods - always better than SGD-m and Adam, but usually lower than that of KFAC and Shampoo. Moreover, the closeness of the plots for MBF and MBF-True shows that using moving average mini-block versions of the EMF rather than the Fisher matrix does not significantly affect the effectiveness of our approach.

We also report a comparison of the performance of MBF-True and MBF on autoencoders and CNN problems in Appendix \ref{apx_mbftrue_vs_mbf}. Note that, in MBF-True, the only difference between it and MBF is that we are using the mini-batch gradient $\overline{\mathcal{D}_2 W_{l, b}}$ (denoted by $\mathcal{D}_2$) of the model on sampled labels $y_t$ from the model's distribution to update the estimate of mini-block preconditioners, using a moving average (see lines 12, 13 in Algorithm \ref{algo_MBF_true} in Appendix \ref{apx_cosine}), with a rank one outer-product, which is different from computing the true Fisher for that mini-block. 

\begin{figure*}
    \centering
    \includegraphics[width=\textwidth]{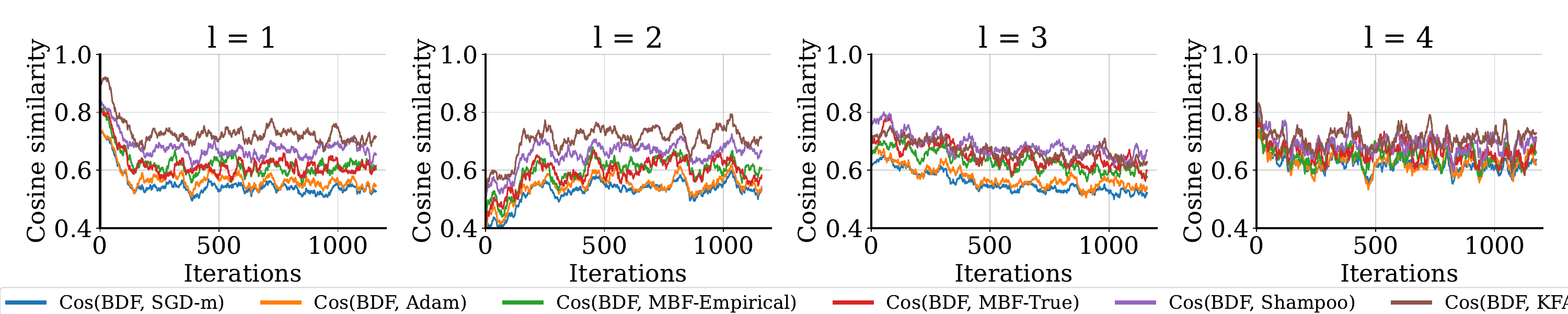}
    \caption{
    Cosine similarity between the directions produced by the methods shown in the legend and that of a block diagonal Fisher method (BDF).
    }
    \label{fig_8}
\end{figure*}
\section{Linear Convergence}\label{linear_convergence}
We follow the framework established in \citep{zhang2019fast} to provide convergence guarantees for the idealized MBF with exact gradients (i.e. full batch case with $m=n$) {and the min-block version of the true Fisher matrix, rather than the EFM, as the underlying preconditioning matrix}. We focus on the single-output case with squared error loss, but analysis of the  multiple-output case is similar. 

We denote by $\bo{u}(\bm{W}) = [f(\bm{W}, x_1), ..., f(\bm{W}, x_n) ]^\top$ the output vector and $y = [\normalfont{y}_1, ..., \normalfont{y}_n ]^\top$ the true labels. We consider the squared error loss  $\mathcal{L}$ on a given data-set $\{x_i,y_i \}_{i=1}^{n}$ with $x_i \in\mathbb{R}^{d}$ and $y_i \in \mathbb{R}$, i.e. the objective is to minimize
$$
\min_{\bm{W} \in \mathbb{R}^{p}} \mathcal{L}(\bm{W})= \frac{1}{2}\Vert \bo{u}(\bm{W})-y\Vert^2.
$$ 
The update rule of MBF with exact gradient becomes
\vspace{-0.2cm}
\begin{equation*}
\scalebox{0.8}{$\bm{W}(k+1) = \bm{W}(k) - \eta \left(\bm{F}_{MB}(\bm{W}(k))  + \lambda \bm{I}\right)^{-1} \bm{J}(k)^\top (\bo{u}(\bm{W}(k))-y),$}
\end{equation*}
where $\bm{F}_{MB}(\bm{W}(k)) := \frac{1}{n}\bm{J}_{MB}(\bm{W}(k))^\top \bm{J}_{MB}(\bm{W}(k))$ 
is the mini-block-Fisher matrix and the mini-block Jacobian is defined as  $\bm{J}_{MB}(k) = \text{Diag}_{l\in[L]} \text{Diag}_{b}\left( J^{\bm{W}_{l, b}}(k)\right)$ and 

{
$$ J^{\bm{W}_{l, b}}(k) := [\frac{\partial f(\bm{W}(k), \textbf{x}_1)}{\partial \bm{W}_{l, b}}, ..., \frac{\partial f(\bm{W}(k), \textbf{x}_n)}{\partial \bm{W}_{l, b}}]^\top$$
}


We use similar assumptions to those used in \citep{zhang2019fast}, where the first assumption, ensures that 
at initialization, the mini-block Gram matrices are all positive-definite, (i.e., the rows of their respective Jacobians are linearly independent), and the second assumption 
ensures the stability of the Jacobians by requiring that the network is close to a linearized network 
at
initialization and therefore MBF's update is close to the gradient descent direction in the output space. These assumptions allow us to control the convergence rate.

\begin{assumption}\label{assump 1} The mini-block Gram matrices $J^{\bm{W}_{l, b}}(0) J^{\bm{W}_{l, b}}(0)^T$ at initialization are positive definite, i.e.  $\min_{l\in[L]} \min_{b}\lambda_{min}(J^{\bm{W}_{l, b}}(0)^T J^{\bm{W}_{l, b}}(0)) =  \lambda_{0} > 0$.
\end{assumption} 
\vspace{-0.2cm}
\begin{assumption}\label{assump 2}
There exists $0<C\leq \frac{1}{2}$  that satisfies $\Vert{\bm{J}(\bm{W}(k))} - \bm{J}(\bm{W}(0))\Vert_2 \leq \frac{C}{3} \sqrt{\lambda_{0}}$ if $\Vert \bm{W}(k) - \bm{W}(0)\Vert_2 \leq\frac{3}{\sqrt{\lambda_{0}}} \Vert \bm{y}-\bm{u}(0)\Vert_2 $.
\end{assumption} 

\begin{thm}\label{thm: Convergence} \label{theorem1}
Suppose Assumptions \ref{assump 1}, \ref{assump 2} hold. Consider the Generic BMF Algorithm 1,  using exact gradients {and the mini-block version of the true Fisher as the underlying preconditioning matrix} for a network with $L$ layers. Then there exists an interval of suitable damping values $\lambda$ in $[\underline{\lambda}, \overline{\lambda}]$ and corresponding small enough learning rates $\eta_{\lambda}$, such that for any learning rate $0 \leq \eta \leq \eta_{\lambda}$ we have
$  {\footnotesize
    \Vert \bo{u}(\bm{W}(k)) - \bo{y}\Vert_2^2 \leq (1 - \eta)^k\Vert \bo{u}(\bm{W}(0)) - \bo{y}\Vert_2^2.
    }$
\end{thm}

\autoref{theorem1} states that an idealized verion of MBF converges to the global optimum with a linear rate under Assumptions \ref{assump 1} and \ref{assump 2}. Our analysis is an adaptation of the proof in \citep{zhang2019fast}, where we first exploit Assumptions \ref{assump 1} and \ref{assump 2} to obtain a positive lower bound on the eigenvalues of mini-block version of the true Fisher matrix $\bm{F}_{MB}(\bm{W}(k))$, which then allows us to characterize the rate of convergence of the method. The proof can be found in the Appendix \ref{apx_proof}.

\section{Implementation Details of MBF and Comparison on Complexity}
{\bf Mini-batch averages, Exponentially decaying averages and Momentum:} Because the size of training data sets is usually large, we use mini-batches to estimate the quantities needed for MBF. We use $\overline{X}$ to denote the average value of $X$ over a mini-batch for any quantity $X$. Moreover, for the EFM  mini-blocks, we use moving averages to both reduce the stochasticity and incorporate more information from the past, more specifically, we use a moving average scheme to get a better estimate of the EFM mini-blocks, i.e.$ \widehat{G_{W_{l, b}}} = \beta \widehat{G_{W_{l, b}}} + (1-\beta) \overline{G_{W_{l, b}}},$ where $\overline{G_{W_{l, b}}}$ is the current approximation to the mini-block EFM defined below. In order to bring MBF closer to a drop-in replacement for adaptive gradient methods such as Adam, we add momentum
to the mini-batch gradient, let: $\widehat{\mathcal{D} W_l} = \mu \widehat{\mathcal{D} W_l} + \overline{\mathcal{D} W_l}$ and then apply the preconditioner to $\widehat{\mathcal{D} W_l}$ to compute the step. 

{\bf Approximating the mini-block Fisher matrices:} 
As mentioned previously, computing the matrices $\overline{G_{W_{l, b}}} := \frac{1}{m} (\bm{J}^{W_{l, b}})^T\bm{J}^{W_{l, b}}$ to update the EFM mini-blocks is inefficient in practice as this requires storing and computing the individual gradients. Hence, we
approximate these mini-block matrices by the outer product of the part of the mini-batch gradient corresponding to the subset of weights $W_{l, b}$, i.e., $\overline{G_{W_{l, b}}} \approx (\overline{\mathcal{D} W_{l, b}}) (\overline{\mathcal{D} W_{l, b}})^\top$.

{\bf Spacial average for large fully-connected layers:}
In some CNN and autoencoder models, using the EFM mini-blocks can still be computationally prohibitive for fc layers,
where both the input and output dimensions are large. Therefore, for such layers we used a Spatial Averaging technique, similar to one used in \citep{yao2020adahessian}, where we maintained a single preconditioning matrix for all the mini-blocks by averaging the approximate mini-block 
EFM matrices whenever we updated the preconditioning matrix. This technique also leads to more stable curvature updates as a side benefit, as observed for the method proposed in \citep{yao2020adahessian}, where the Hessian diagonal was "smoothed" across each  layer. We also explored using spacial averaging for convolutional layers. However since the kernel-wise mini-blocks are small in size, spacial averaging doesn't compare favorably to the full MBF method (see Appendix \ref{apx_cnn_avg}).

{\bf Amortized updates of the preconditioning matrices:} 
The extra work for the above computations, as well as for updating the inverses 
$ F^{-1}_{W_{l, j}}$ compared with first-order methods is amortized by only performing the Fisher matrix
updates every $T_1$ iterations and computing their inverses every $T_2$ iterations. This approach which is also used in KFAC and Shampoo, does not seem to degrade MBF's overall performance, in terms of computational speed.

{\bf Comparison of Memory and Per-iteration Time Complexity.} In Table \ref{table_comp}, we compare the space and computational requirements of the proposed MBF method with KFAC (see Appendix \ref{apx_algos}) and Adam, which are among the predominant 2nd and 
1st-order methods, respectively, used to train DNNs. We focus on one convolutional layer, with $J$ input channels, $I$ output channels, kernel size $|\Delta| = (2R+1)^2$, and $|\mathcal{T}|$ spacial locations. Let $m$ denote the size of the minibatches, and {$T_1$ and $T_2$} denote, respectively, the frequency for updating the preconditioners and inverting them for both KFAC and MBF. As indicated in Table \ref{table_comp}, the amount of memory required by MBF is the same order of magnitude as that required by Adam, (specifically, more by a factor of $|\Delta|$, which is usually small in most CNN architectures; e.g, in VGG16 \citep{simonyan2014very} $|\Delta| = 9$) and less than KFAC, Shampoo and other SOTA Kronecker-factored preconditioners, (specifically, e.g., by a factor of $O\left(J + \frac{I}{|\Delta|}\right)$ for KFAC. 

We can also see  that MBF requires only a small amount more per-iteration time than Adam (i.e., by a factor of $\left(\frac{|\Delta|}{T_1} + \frac{|\Delta|^2}{T_2} + |\Delta|\right)$). Note that in our experiments, $T_1 \approx |\Delta|$ and $T_2 \approx |\Delta|^2$. The computationnal and storage requirements for fully connected layers are discussed in Appendix \ref{apx_other_details}. Our MBF algorithm is described fully as Algorithm \ref{algo_MBF_full} in the Appendix \ref{apx_mbf_full}.

\begin{table*}
\scriptsize
  \caption{Computation and Storage Requirements per iteration for convolutional layer.}
  \vskip 0.15in
  \label{table_comp}
  \centering
  \begin{tabular}{l|cccllllll|c}
    \hline
    Algorithm
    & Additional pass
    & Curvature
    & Step $\Delta W_l$
    & Storage $P_l$
    \\
    \hline
    MBF
    & ---
    & {$O(\frac{ IJ |\Delta|^2}{T_1}   + \frac{IJ |\Delta|^3}{T_2}) $}
    & $O(I J |\Delta|^2 )$
    & $O(I J |\Delta|^2 )$
    \\
    Shampoo
    & ---
    & {$O(\frac{(J^2 + |\Delta|^2 + I^2)}{T_1} + \frac{J^3  + I^3 + |\Delta|^3}{T_2} )$}
    & $O((I+J+|\Delta|) I J |\Delta|)$
    & $O(I^2 + J^2 + |\Delta|^2)$
    \\
    KFAC
    &
    {$O(\frac{m I J |\Delta| |\mathcal{T}|}{T_1} )$}
    &
    {$O(\frac{m (J^2 |\Delta|^2 + I^2) |\mathcal{T}|}{T_1} + \frac{J^3 |\Delta|^3 + I^3}{T_2} )$}
    & $O(I J^2 |\Delta|^2 + I^2 J |\Delta|)$
    & $O(J^2 |\Delta|^2 + I^2)$
    \\
    Adam
    & ---
    & $O(I J |\Delta|)$
    & $O(I J |\Delta|)$
    & $O(I J |\Delta|)$
    \\
    \hline
  \end{tabular}
\end{table*}


\section{Experiments}\label{experiments}
In this section, we compare MBF with some
SOTA
first-order (SGD-m, Adam) and second-order (KFAC, Shampoo) methods. (See Appendix \ref{apx_algos} on how these methods were implemented.) Since MBF uses information about the second-moment of the gradient to construct a preconditioning matrix, 
Adam, KFAC and Shampoo were obvious choices for comparison with MBF.

\begin{figure}[t]
    \centering
    \begin{minipage}{0.49\textwidth}
        \centering
        \subfigure[\footnotesize CIFAR-10, ResNet-32]{\includegraphics[width=\textwidth,height=3cm]{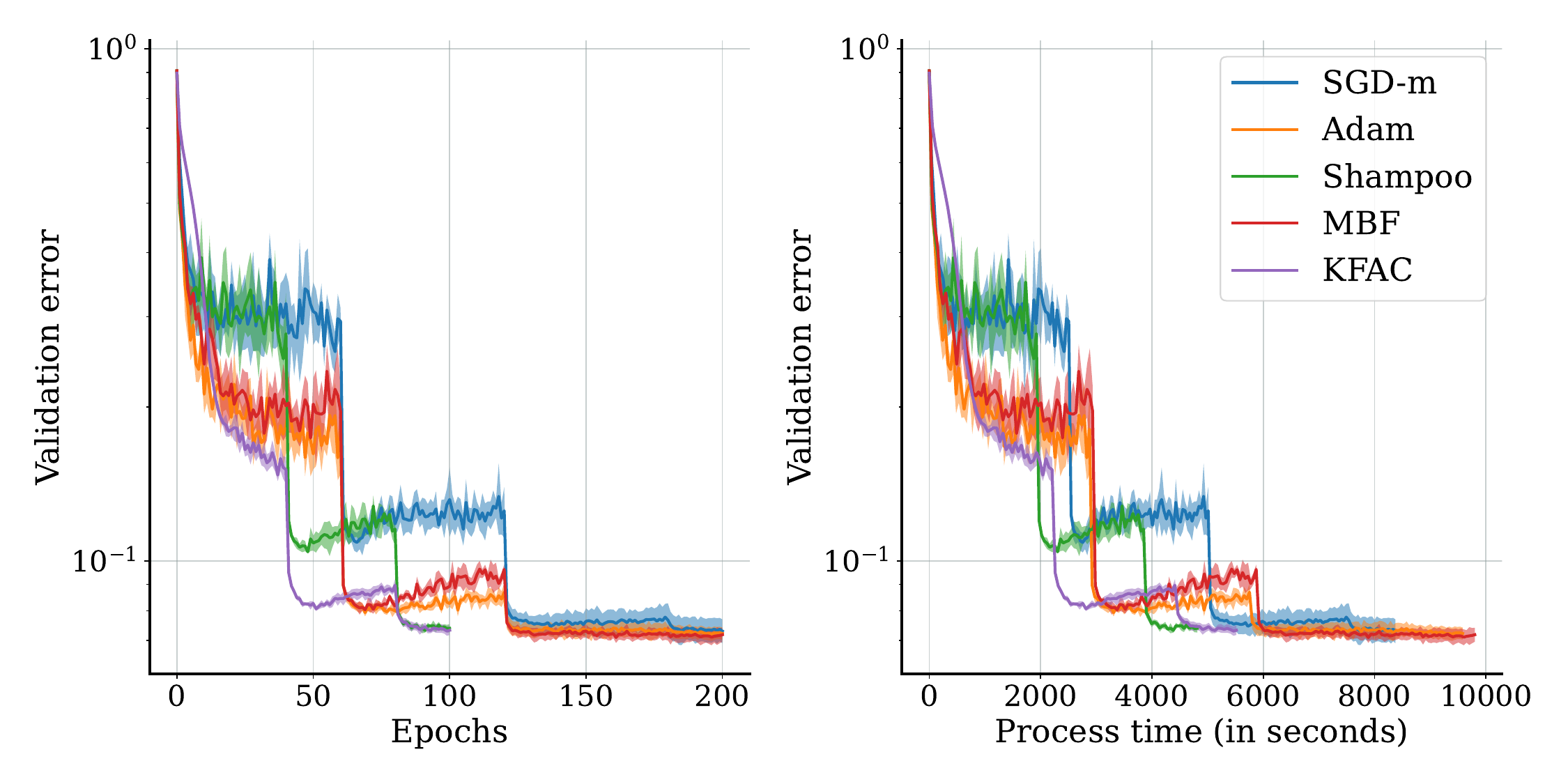}}\\
        \subfigure[\footnotesize CIFAR-100, VGG16]{\includegraphics[width=\textwidth,height=3cm]{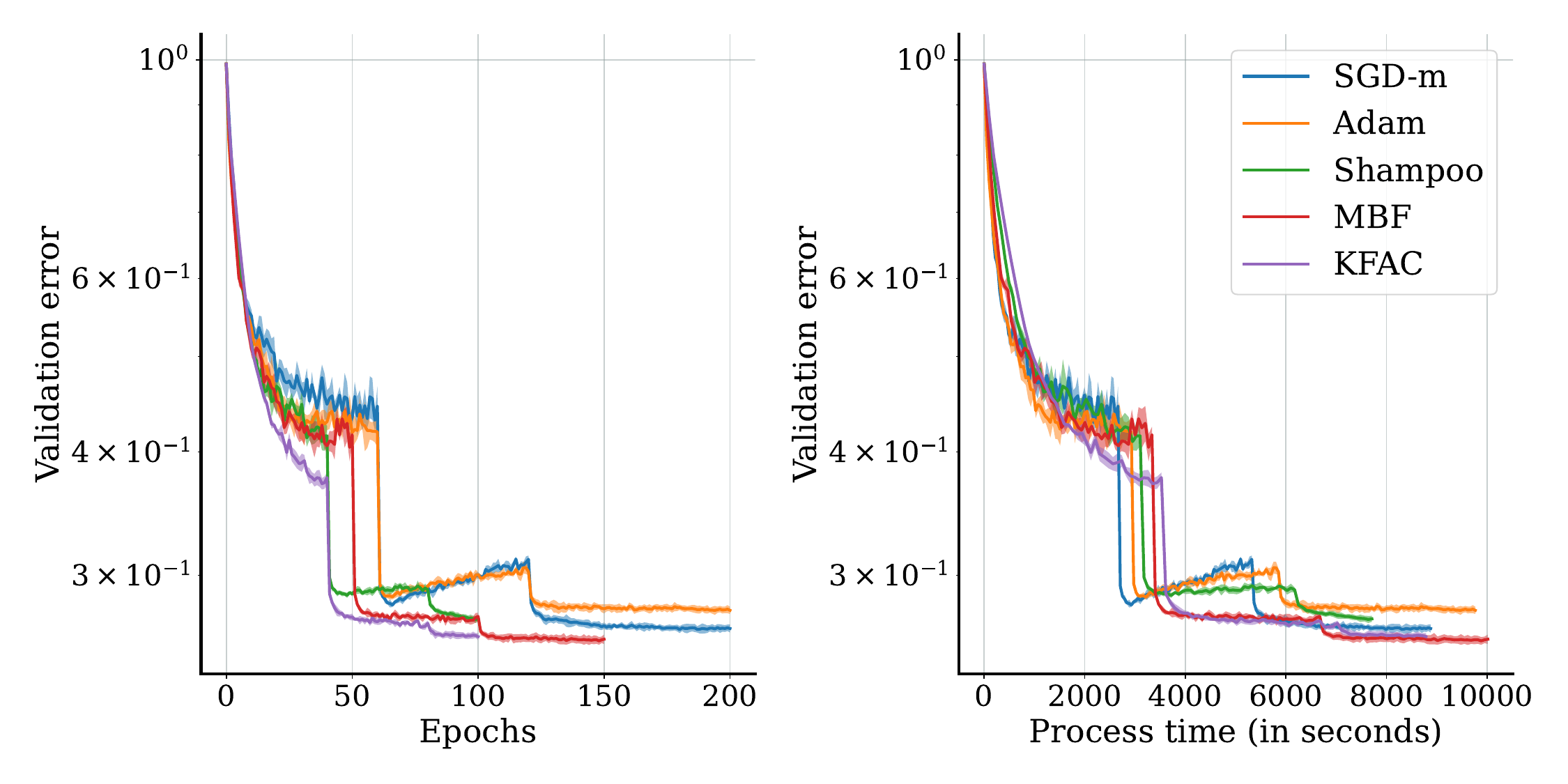}}\\
        \subfigure[\footnotesize SVHN, VGG11]{\includegraphics[width=\textwidth,height=3cm]{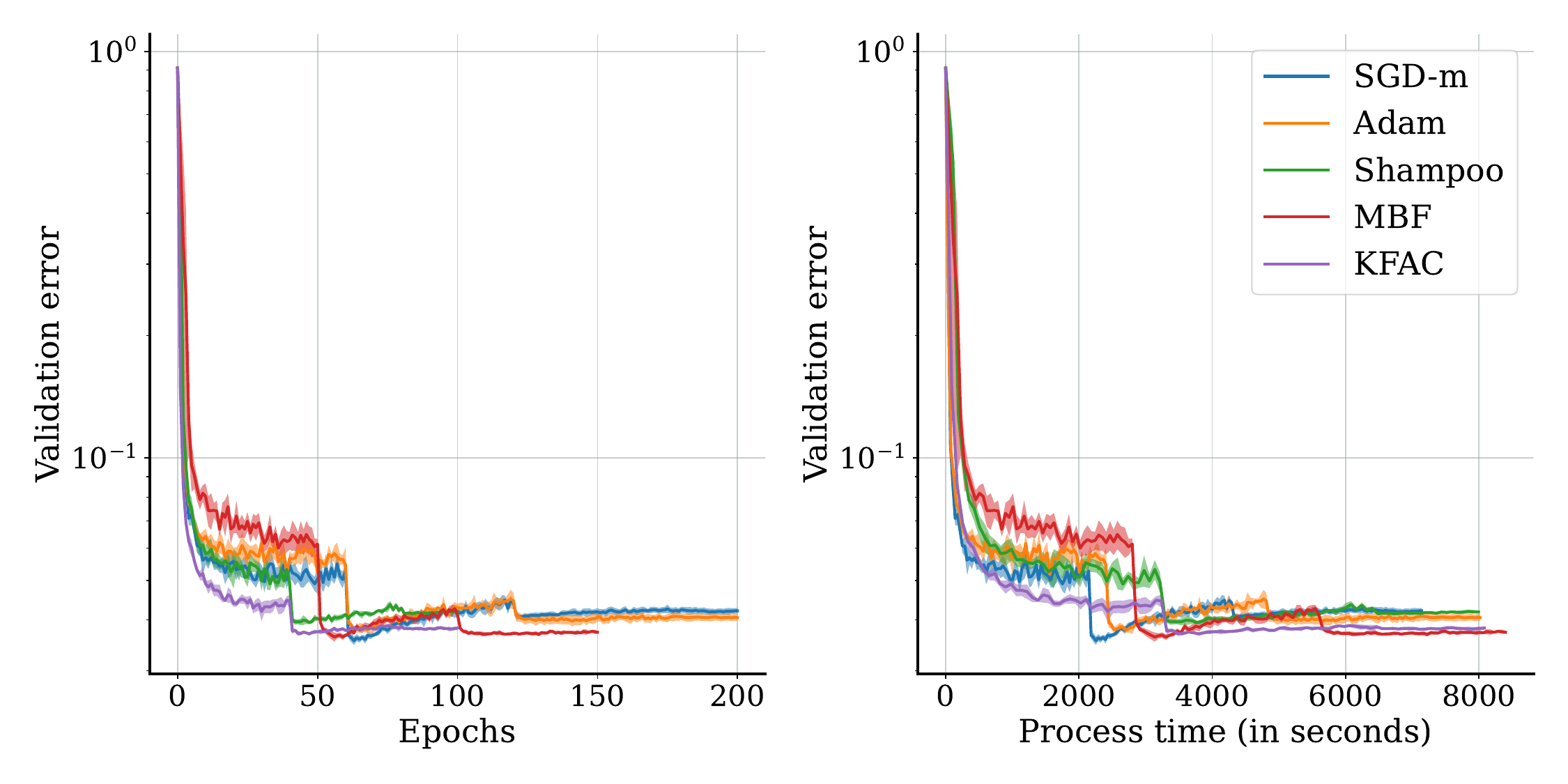}}
        \caption{Generalization ability of MBF, KFAC, Shampoo, Adam, and SGD-m on three CNN problems.}
         \label{fig_cnns}
    \end{minipage}
\end{figure}

\begin{figure}[t]
    \begin{minipage}{0.49\textwidth}
        \centering
        \subfigure[\footnotesize MNIST autoencoder]{\includegraphics[width=\textwidth,height=3cm]{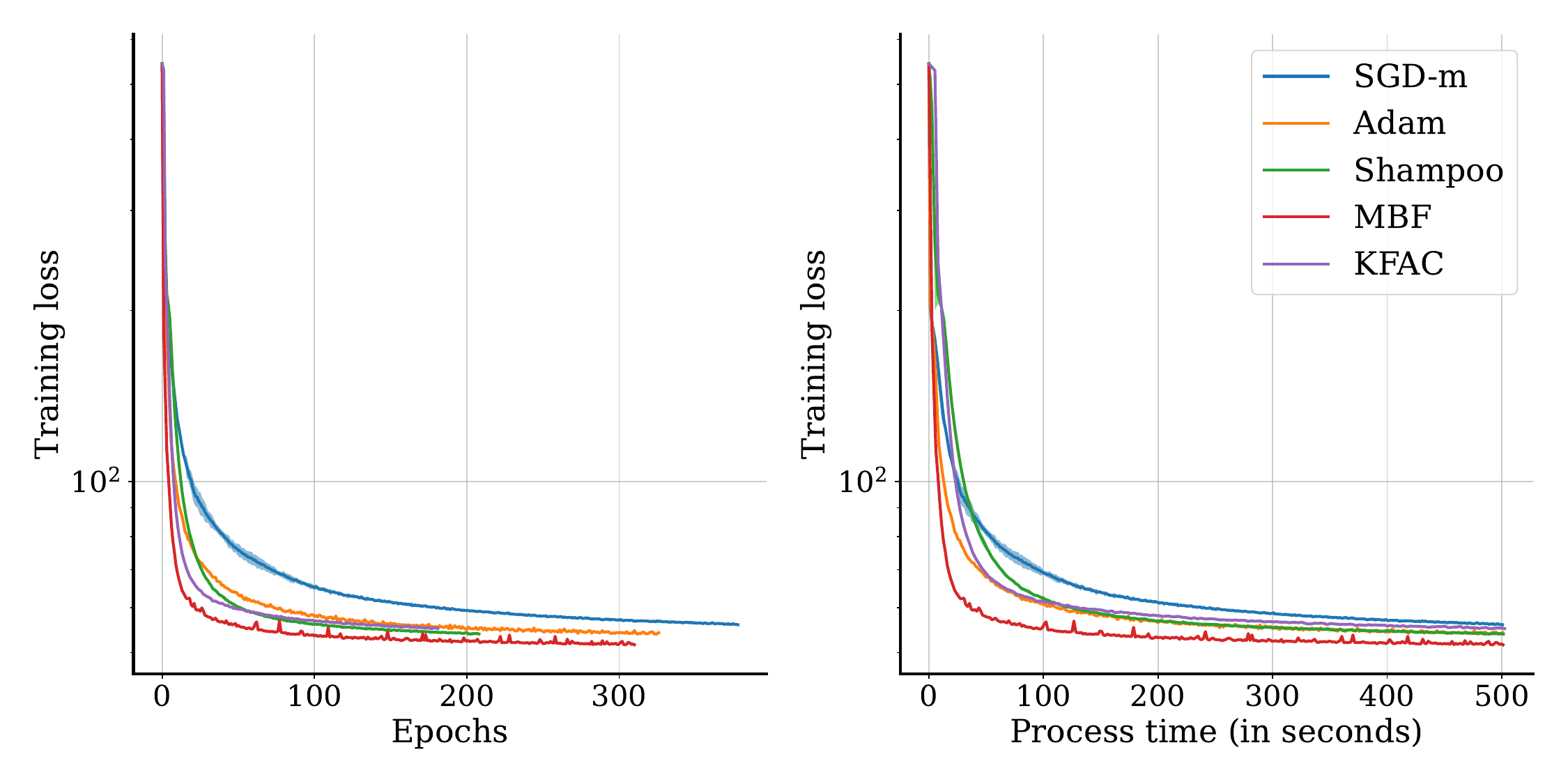}}\\
        \subfigure[\footnotesize FACES autoencoder]{\includegraphics[width=\textwidth,height=3cm]{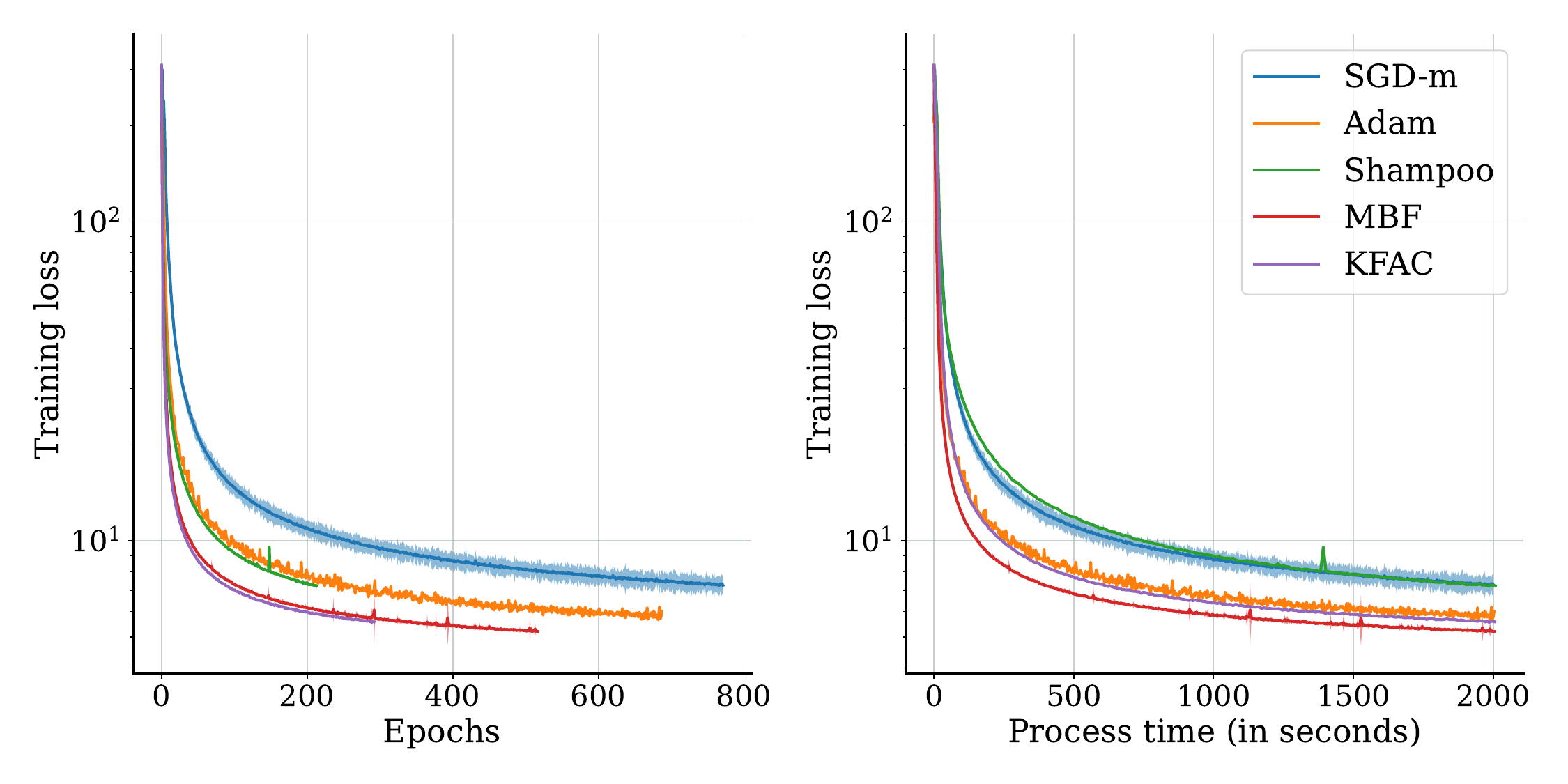}}\\
        \subfigure[\footnotesize CURVES autoencoder]{\includegraphics[width=\textwidth,height=3cm]{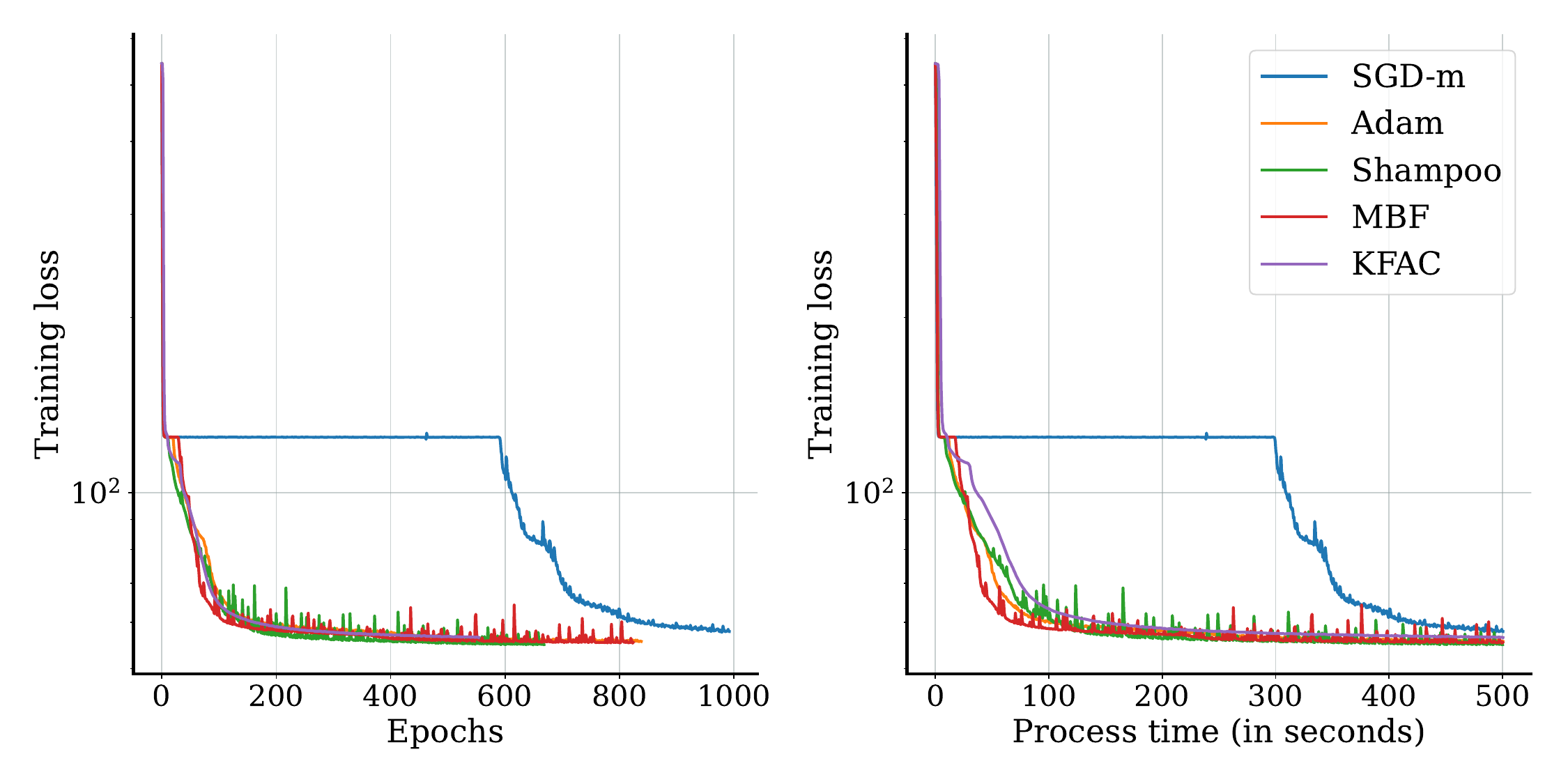}}
        \caption{Optimization performance of MBF, KFAC, Shampoo, Adam, and SGD-m on three autoencoder problems.}
         \label{fig_autoencoders}
    \end{minipage}
\end{figure}

Our experiments were run on a machine with one V100 GPU and eight Xeon Gold 6248 CPUs using PyTorch \cite{NEURIPS2019_9015}. Each algorithm was run using the best hyper-parameters, determined by a grid search (specified in Appendices \ref{apx_exp_ccn} and \ref{apx_exp_fcc}), and 5 different random seeds. The performance of MBF and the comparison algorithms is plotted in Figures \ref{fig_cnns} and  \ref{fig_autoencoders}: the solid curves depict the results averaged over the 5 different runs, and the shaded areas depict the  $\pm$standard deviation range for these runs.

\label{sec_9}





\paragraph{Generalization performance, CNN problems:} We first compared the generalization performance of MBF to SGD-m, Adam, KFAC and Shampoo on three CNN models, namely,
 ResNet32 \cite{he2016deep},
  VGG16 \cite{simonyan2014very} and VGG11 \cite{simonyan2014very}, respectively, on the datasets
 CIFAR-10, CIFAR-100 and SVHN
 \cite{krizhevsky2009learning}. The first two have 50,000 training data and 10,000 testing data (used as the validation set in our experiments), while SVHN
 has 73,257 training data and 26,032 testing data.
For all algorithms, we used a batch size of 128.
In training, we applied data augmentation as described in  \cite{krizhevsky2012imagenet}, including random horizontal flip and random crop, since these setting choices have been used and endorsed in many previous research papers, e.g. \citet{zhang2018three,choi2019empirical,ren2021tensor}. (see Appendix \ref{apx_exp} for more details about the experimental set-up)

On all three model/dataset problems, the first-order methods were run for 200 epochs, and KFAC and Shampoo for 100 epochs, while MBF was run for 150 epochs on VGG16/CIFAR-100 and VGG11/SVHN,
and 200 epochs on ResNet32/CIFAR-10.
The reason that we ran MBF for 200 epochs (i.e., the same number as run for Adam) on ResNet32 was because all of ResNet32's convolutional layers use small ($3\times3$) kernels, and it contains just one fully connected layer of modest size $(I,O) = (64, 10)$. Hence as we expected, MBF and Adam took almost the same time to complete 200 epochs. As can be seen in Figure \ref{fig_cnns}, MBF could have been terminated after 150 epochs, without a significant change in validation error. On the other hand, since VGG16 and VGG11 have two large fully connected-layers (e.g [4096, 4096, 10/100]), MBF's per-iteration computational cost is substantially larger than Adam's due to these layers. Consequently, for both methods to finish roughly in the same amount of time, we ran MBF for only 150 epochs.

All methods employed a learning rate (LR) schedule that decayed LR by a factor of 0.1 every K epochs, where K was set to 40, 50 and 60, for the first-order methods, MBF, and KFAC/Shampoo, respectively, on the VGG16 and VGG11 problems, and set to 40, 60, and 80, respectively, on the ResNet32 problem

Moreover, weight decay, which has been shown to improve generalization across different optimizers  \cite{loshchilov2018decoupled,zhang2018three}, was employed by all of the algorithms, and a grid search on the weight decay factor and the initial learning rate based on the criteria of maximal validation classification accuracy, was performed. Finally, the damping parameter was set to 1e-8 for Adam (following common practice), and 0.03 for KFAC (\url{https://github.com/alecwangcq/KFAC-Pytorch}).
For Shampoo, we set $\epsilon=0.01$. For MBF, we set $\lambda = 0.003$. We set $T_1=10$ and $T_2=100$ for KFAC, Shampoo and MBF. 

From Figure \ref{fig_cnns}, we see that MBF has a similar (and sometimes better) generalization {performance} than the other methods. Moreover, in terms of process time, MBF is roughly as fast as SGD-m and Adam on ResNet32/CIFAR-10 in Figure \ref{fig_cnns}, and is competitive with all of the SOTA first and second-order methods in our experiments. 


\paragraph{Optimization performance, Autoencoder Problems:}


We also compared the optimization performance of the algorithms on three autoencoder problems \cite{hinton2006reducing} with datasets MNIST \cite{lecun2010mnist}, FACES, and CURVES, which were also used for benchmarking algorithms in \cite{martens2010deep,martens2015optimizing,botev2017practical,goldfarb2020practical}.
The details of the layer shapes of the autoencoders are specified in Appendix \ref{apx_exp_fcc}. For all algorithms, we used a batch size of 1,000, and settings that largely mimic the settings in the latter papers. Each algorithm was run for 
500 seconds for MNIST and CURVES, and 2000 seconds for FACES. 

For each algorithm, we conducted a grid search on the LR and damping value based on minimizing the training loss. We set the Fisher matrix update frequency $T_1=1$ and inverse update frequency $T_2=20$ for second-order methods, as in \cite{ren2021tensor}. From Figure \ref{fig_autoencoders}, it is clear that MBF outperformed SGD-m and Adam, both in terms of per-epoch progress and process time. Moreover, MBF performed (at least) as well as KFAC and Shampoo. We postulate that  the performance of MBF is due to its ability to capture important curvature information from the mini-block Fisher matrix, while keeping the computational cost per iteration low and close to
that of Adam.

\section{Conclusion and Future Research}
We proposed a new EMF-based method, MBF, for training DNNs, by approximating the EMF by a mini-block diagonal matrix that arises naturally from the structure of convolutional and ff-fc layers. MBF  requires very mild memory and computational overheads, compared with first-order methods, and is easy to implement. Our experiments on various DNNs and datasets, demonstrate conclusively that MBF provides comparable and sometimes better results than SOTA methods, {both} from an optimization and generalization perspective. Future research will investigate extending MBF to other deep learning architectures such as Recurrent neural networks.

\newpage
\medskip
\bibliographystyle{agsm}
\bibliography{references}

\newpage 
\onecolumn
\aistatstitleapx{Supplementary Materials for "A Mini-Block Fisher Method for Deep Neural Networks"}
\section{MBF Full implementation}\label{apx_mbf_full}

We present below pseudo-code for the full detailed implementation of our MBF algorithm thatwe us ed in producing the results in the main text.

\begin{algorithm}
\caption{Mini-Block Fisher method (MBF)}
    \label{algo_MBF_full}
    \begin{algorithmic}[1]
    \REQUIRE
    Given batch size $m$, learning rate $\{ \eta_k \}_{k \ge 1}$, {weight decay factor $\gamma$,} damping value $\lambda$, statistics update frequency $T_1$, inverse update frequency $T_2$
    \STATE $\mu = 0.9$, $\beta = 0.9$
    \STATE Initialize $\widehat{G_{l, b}} = \E [G_{l, b}]$ ($l = 1,..,k$, mini-blocks $b$) by iterating through the whole dataset, $\widehat{\mathcal{D} W_{l, b}}=0$ ($l = 1,..,k$, mini-blocks $b$)
    \FOR {$k=1,2,\ldots$}
        \STATE Sample mini-batch $M_t$ of size $m$ 
        \STATE Perform a forward-backward pass over $M_t$ to compute the mini-batch gradient $\overline{\mathcal{D} W_{l, b}}$
        \FOR {$l = 1, ... L$}
        \FOR{mini-block $b$ in layer $l$, \textbf{in parallel}}
        \STATE $\widehat{\mathcal{D} W_{l, b}} = \mu \widehat{\mathcal{D} W_{l, b}} + \overline{\mathcal{D} W_{l, b}}$
        \IF {$k \equiv 0 \pmod{T_1}$}
            \STATE If Layer $l$ is convolutional: $\widehat{G_{l, j,i}} = \beta \widehat{G_{l, j,i}} + (1-\beta)\overline{\mathcal{D} W_{l, j,i}} \left(\overline{\mathcal{D} W_{l, j,i}}\right)^\top$ 
            \STATE If Layer $l$ is fully-connected: $\widehat{G_{l}} = \beta \widehat{G_{l}} +  \frac{1-\beta}{O} \sum_{j = 1}^{O}\overline{\mathcal{D} W_{l, j}} \left(\overline{\mathcal{D} W_{l, j}}\right)^\top$
        \ENDIF
        \IF {$k \equiv 0 \pmod{T_2}$}
            \STATE
            Recompute and store $(\widehat{G_{l, b}} + \lambda I)^{-1}$
        \ENDIF
        \STATE $p_{l, b} = (\widehat{G_{l, b}} + \lambda I)^{-1} \widehat{\mathcal{D} W_{l, b}} + \gamma W_{l, b}$ 
        \STATE $W_{l, b} = W_{l, b} - \eta_k p_{l, b}$
        \ENDFOR
        \ENDFOR
        \ENDFOR  
    \end{algorithmic}
\end{algorithm}

\section{Proof of Convergence of Algorithm MBF and Associated Lemmas}
\label{apx_proof}
We follow the framework used in \cite{zhang2019fast} to prove linear convergence of NG descent, to provide similar convergence guarantees for our idealized MBF Algorithm, that uses exact gradients (i.e. full batch case with $m=n$) \footnote{in \cite{tengrad}, a similar extension of the proof in \cite{zhang2019fast} is used to analyse the convergence of a layer-wise block Fisher method.} {and the mini-block version of the true Fisher as the underlying preconditioning matrix}.

\textbf{Proof of Theorem 1}. 
If Assumption 6.2 holds, then one can obtain a lower bound on the minimum eigenvalue of the mini-block Fisher matrix $\bm{F}_{MB}(\bm{W}(k)) = \frac{1}{n}\bm{J}_{MB}(k)^\top \bm{J}_{MB}(k)$ at each iteration.

In fact, if $\Vert \bm{W}(k) - \bm{W}(0)\Vert_2 \leq\frac{3}{\sqrt{\lambda_{0}}} \Vert \bm{y}-\bm{u}(0)\Vert_2 $, then, by Assumption 6.2, there exists $0<C\leq \frac{1}{2}$  that satisfies $\Vert{\bm{J}(\bm{W}(k))} - \bm{J}(\bm{W}(0))\Vert_2 \leq \frac{C}{3} \sqrt{\lambda_{0}}$, and therefore, we have that 
\begin{align*}
\|\bm{J}_{MB}(k)-\bm{J}_{MB}(0)\|_{2} \leq \frac{C \sqrt{\lambda_{0}}}{3} \leq \frac{ \sqrt{\lambda_{0}}}{3}.
\end{align*}
On the other hand, based on the inequality $\sigma_{\min }(\mathbf{A}+\mathbf{B}) \geq \sigma_{\min }(\mathbf{A})-\sigma_{\max }(\mathbf{B})$, where $\sigma$ denotes singular value, we have
\begin{align*}
\sigma_{\min }(\bm{J}_{MB}(k)) &\geq \sigma_{\min }(\bm{J}_{MB}(0))- \sigma_{\min }(\bm{J}_{MB}(k)-(\bm{J}_{MB}(k))) \\
&\geq \sigma_{\min }(\bm{J}_{MB}(0))- \|\bm{J}_{MB}(k)-\bm{J}_{MB}(0)\|_{2} \geq \sqrt{\lambda_0} -\frac{ \sqrt{\lambda_{0}}}{3} = \frac{2\sqrt{\lambda_{0}}}{3}.
\end{align*}
Therefore
\begin{align*}
\lambda_{min}(\bm{G}_{MB}(\bm{W}(k))) \geq \frac{4\sqrt{\lambda_{0}}}{9}.
\end{align*}

where $\bm{G}_{MB}(\bm{W}(k)) := \bm{J}_{MB}(\bm{W}(k)) \bm{J}_{MB}(\bm{W}(k))^\top$ is the mini-block Gram matrix. We prove Theorem 1 by induction. Assume $|| \bo{u}(\bm{W}(k)) - \bm{y}||_2^2 \leq (1 - \eta)^k ||  \bo{u}(\bm{W}(0)) - \bo{y}||_2^2$. One can see that the relationship between the Jacobian $\bm{J}(\bm{W}(k))$ and the mini-Block Jacobian $\bm{J}_{MB}(\bm{W}(k))$ is:
$$\bm{J}^\top(\bm{W}(k)) = \bm{J}_{MB}(\bm{W}(k))^\top \bm{K},$$
where the matrix $\bm{K} = \underbrace{[I_n,\dots,I_n] ^\top}_{K} \in \mathbb{R}^{Kn \times n}$, $I_n$ is the identity matrix of dimension $n$, the number of samples, and $K$ is the total number of mini-blocks. We define 
\begin{align*}
    \bm{W}_k(s) &= s \bm{W}(k+1)+(1-s) \bm{W}(k) \\
    &= \bm{W}(k) -s \frac{\eta}{n} \left( \bm{F}_{MB}(\bm{W}(k)) + \lambda I \right) ^{-1}  \bm{J}(\bm{W}(k))^\top (\bo{u}(\bm{W}(k)) - \bm{y}))    - \bo{u}(\bm{W}(k)),
\end{align*}
we have:
\begin{align*}
&\bo{u}(\bm{W}(k+1)) -  \bo{u}(\bm{W}(k)) \\
&= \bm{u}(\bm{W}(k) - \frac{\eta}{n} \left( \bm{F}_{MB}(\bm{W}(k)) + \lambda I \right) ^{-1}  \bm{J}(\bm{W}(k))^\top (\bo{u}(\bm{W}(k)) - \bm{y}))    - \bo{u}(\bm{W}(k))\\
&= -\int_{s=0}^1  \inp[\Big]{\frac{\partial \bm{u}(\bm{W}_k(s))   }{\partial \bm{W} ^\top }}{\frac{\eta}{n}  \left( \bm{F}_{MB}(\bm{W}(k)) + \lambda I \right) ^{-1}  \bm{J}(\bm{W}(k))^\top (\bm{u}(\bm{W}(k)) - \bm{y}))} ds \\
&= -\underbrace{\int_{s=0}^1  \inp[\Big]{\frac{\partial \bm{u}(\bm{W}(k))   }{\partial \bm{W} ^\top }}{\frac{\eta}{n}  \left( \bm{F}_{MB}(\bm{W}(k)) + \lambda I \right) ^{-1}  \bm{J}(\bm{W}(k))^\top (\bm{u}(\bm{W}(k)) - \bm{y}))} ds}_{\circled{A}}\\
&+ \underbrace{\int_{s=0}^1  \inp[\Big]{\frac{\partial \bm{u}(\bm{W}(k))   }{\partial \bm{W} ^\top } - \frac{\partial \bm{u}(\bm{W}_k(s))   }{\partial \bm{W} ^\top }}{\frac{\eta}{n}  \left( \bm{F}_{MB}(\bm{W}(k)) + \lambda I \right) ^{-1}  \bm{J}(\bm{W}(k))^\top (\bm{u}(\bm{W}(k)) - \bm{y}))} ds}_{\circled{B}}
\end{align*}
In what follows, to simplify the notation, we drop $\bm{W}(k)$ whenever the context is clear. Thus, we have
\begin{align}
\circled{A} &= \frac{\eta}{n}  \bm{J} \left( \bm{F}_{MB} + \lambda I \right) ^{-1}  \bm{J}^\top (\bm{y}-\bm{u}(k)). \label{eq:proof_1}    
\end{align}
Now, we bound the norm of $\circled{B}$:
\begin{align}
||\circled{B}||_2 &\leq \frac{\eta}{n} \norm{ \int_{s=0}^1 \bm{J}(\bm{W}_k(s)) - \bm{J}(\bm{W}(k)) ds}_2 \norm{\left( \bm{F}_{MB} + \lambda I \right) ^{-1}  \bm{J}^\top (\bm{u}(k)-\bm{y})}_2\nonumber\\
&\overset{(1)}{\leq} \frac{\eta 2C}{3n} \lambda_0^{\frac{1}{2}} \norm{\left( \frac{1}{n} \bm{J}_{MB} ^\top \bm{F}_{MB}  + \lambda I \right) ^{-1}  \bm{F}_{MB} ^\top \bm{K} (\bm{u}(k) - \bm{y}))}_2  \nonumber\\
&\leq \frac{\eta 2C}{3n} \lambda_0^{\frac{1}{2}} \norm{\left( \frac{1}{n} \bm{J}_{MB}^\top \bm{J}_{MB}  + \lambda I \right) ^{-1}  \bm{J}_{MB}^\top}_2 \norm{\bm{K} (\bm{u}(k) - \bm{y}))}_2 \nonumber\\
& \overset{(2)}{\leq} \frac{\eta C}{3 \sqrt{\lambda n} } \sqrt{\lambda_0} \norm{\bm{K} (\bm{u}(k) - \bm{y}))}_2 \overset{(3)}{=}\frac{\eta C \sqrt{\lambda_0 K}}{3 \sqrt{\lambda n}}  \norm{(\bm{u}(k) - \bm{y}))}_2, \label{eq:proof_2}
\end{align}
where in (1) we used Assumption 6.2, which implies
\begin{align*}
    \norm{ \int_{s=0}^1 \bm{J}(\bm{W}_k(s)) - \bm{J}(\bm{W}(k)) ds}_2 &\leq \norm{ \bm{J}(\bm{W}(k)) - \bm{J}(\bm{W}(0))}_2 + \norm{ \bm{J}(\bm{W}(k+1)) - \bm{J}(\bm{W}(0)) }_2\nonumber\\
    &\leq \frac{2C}{3} \sqrt{\lambda_0}.
\end{align*}
The inequality (2) follows from the fact that 
\begin{align*}
    \norm{\left( \frac{1}{n} \bm{J}_{MB}^\top \bm{J}_{MB} + \lambda I \right) ^{-1}  \bm{J}_{MB}^\top }_2 &= \sigma_{max} \left(\left( \frac{1}{n} \bm{J}_{MB}^\top \bm{J}_{MB} + \lambda I \right) ^{-1}  \bm{J}_{MB}^\top \right) \\
    &= \sqrt{\lambda_{max}\left( \bm{J}_{MB} \left( \frac{1}{n} \bm{J}_{MB}^\top \bm{J}_{MB} + \lambda I \right) ^{-2}  \bm{J}_{MB}^\top \right)},
\end{align*}
and that
\begin{align*}
    \lambda_{max}\left( \bm{J}_{MB} \left( \frac{1}{n} \bm{J}_{MB}^\top \bm{J}_{MB} + \lambda I \right) ^{-2}  \bm{J}_{MB}^\top \right) = \max_{\mu \text{ eigenvalue of } \bm{G}_{MB}} \frac{\mu}{(\frac{\mu}{n} + \lambda)^2} \leq \frac{n \lambda}{(\frac{n \lambda}{n} + \lambda)^2} = \frac{n}{4 \lambda}.
\end{align*}
and in the 
equality (3), we have used the fact that $\norm{\bm{K} (\bm{u}(k) - \bm{y}))}_2 = \sqrt{K} \norm{ (\bm{u}(k) - \bm{y}))}_2$.
Finally, we have:
\begin{align*}
|| \bm{u}(k+1) - \bm{y}||_2^2 &= || \bm{u}(k) - \bm{y} + \bm{u}(k+1)-\bm{u}(k)||_2^2  \\
&= || \bm{u}(k) - \bm{y}||_2^2 -2 \left( \bm{y} - \bm{u}(k) \right) ^\top  \left( \bm{u}(k+1)- \bm{u}(k) \right) + ||\bm{u}(k+1)- \bm{u}(k) ||_2^2\\
&\leq || \bm{u}(k) - \bm{y}||_2^2 -\frac{2\eta}{n} \underbrace{\left( \bm{y} - \bm{u}(k) \right) ^\top  \bm{J}(k) \left( \bm{F}_{MB} + \lambda I \right) ^{-1}  \bm{J}(k)^\top \left( \bm{y} - \bm{u}(k) \right) }_{\circled{1}}\\
&\quad\quad\quad+ \frac{2\eta C \sqrt{\lambda_0 K}}{3\sqrt{\lambda n}}  \norm{(\bm{u}(k) - \bm{y}))}_2^2 + \underbrace{||\bm{u}(k+1)- \bm{u}(k) ||_2^2}_{\circled{2}}\\
&\leq || \bm{u}(k) - \bm{y}||_2^2 - \frac{2\eta K \lambda_0}{\lambda_0 + \frac{9}{4}n\lambda}  || \bm{u}(k) - \bm{y}||_2^2 \\
&+ \frac{2\eta C \sqrt{\lambda_0 K}}{3 \sqrt{\lambda n}}  \norm{(\bm{u}(k) - \bm{y}))}_2^2 + \eta^2 \left(K + \frac{C \sqrt{\lambda_0 K}}{3 \sqrt{\lambda n}}\right)^2\norm{(\bm{u}(k) - \bm{y}))}_2^2 \\
&\leq (1-\eta) \norm{(\bm{u}(k) - \bm{y}))}_2^2 \\ 
&+ \eta \norm{(\bm{u}(k) - \bm{y}))}_2^2 \left( \eta \left(K + \frac{C \sqrt{\lambda_0 K}}{3 \sqrt{\lambda n}}\right)^2 - \left(\frac{2 K \lambda_0}{\lambda_0 + \frac{9}{4}n\lambda} - \frac{2 C \sqrt{\lambda_0 K}}{3 \sqrt{\lambda n}} -1\right) \right).\\
\end{align*}
Part $\circled{1}$ is lower bounded as follows:
\begin{align*}
    \begin{split}
        \circled{1} &\geq \lambda_{\min} \left(\bm{J}_{MB} \left( \frac{1}{n} \bm{J}_{MB}^\top \bm{J}_{MB} + \lambda I \right) ^{-1}  \bm{J}_{MB}^\top  \right) 
        \Vert \bm{K} (\bm{u}(k) - \bm{y}) \Vert_2^2 \\
        & = K \lambda_{\min} \left(\bm{J}_{MB} \left( \frac{1}{n} \bm{J}_{MB}^\top \bm{J}_{MB} + \lambda I \right) ^{-1}  \bm{J}_{MB}^\top  \right) 
        \Vert \bm{u}(k) - \bm{y} \Vert_2^2 \\
        &= n K  \Vert \bm{u}(k) - \bm{y} \Vert_2^2 \frac{\lambda_{min}(\bm{G}_{MB}(k))}{\lambda_{min}(\bm{G}_{MB}(k)) + n\lambda} \\
        &\geq \frac{n K\lambda_0}{\lambda_0 + \frac{9}{4}n\lambda} \Vert \bm{u}(k) - \bm{y} \Vert_2^2.
    \end{split}
\end{align*}
Part $\circled{2}$ is upper bounded, on the other hand, using  equality \eqref{eq:proof_1}
and inequality \eqref{eq:proof_2}. More specifically, we have:
\begin{align*}
    &||\bm{u}(k+1)- \bm{u}(k) ||_2 \leq \frac{\eta}{n}  \norm{\bm{J}(k) \left( \bm{F}_{MB} + \lambda I \right) ^{-1}  \bm{J}(k)^\top (\bm{y} - \bm{u}(k)))} + ||\circled{B}||_2 \\
    &\leq \frac{\eta K}{n}  \norm{\bm{J}_{MB}(k) \left( \bm{F}_{MB} + \lambda I \right) ^{-1}  \bm{J}_{MB}(k)^\top} \norm{(\bm{u}(k) - \bm{y}))}_2 + \frac{\eta C \sqrt{\lambda_0 K}}{3 \sqrt{\lambda n}}  \norm{(\bm{u}(k) - \bm{y}))}_2 \\
    &\leq \eta \left(K + \frac{ C \sqrt{\lambda_0 K}}{3 \sqrt{\lambda n}} \right) \norm{(\bm{u}(k) - \bm{y}))}_2.
\end{align*}

The last inequality follows from the fact that if $(\mu, v)$ is an (eigenvalue, eigenvector) pair for $\bm{G}_{MB} = \bm{J}_{MB} \bm{J}_{MB}^\top$, then $(\mu, \bm{J}_{MB}^\top v)$ and $(\frac{1}{\frac{\mu}{n} + \lambda}, \bm{J}_{MB}^\top v)$ are such pairs for $\bm{F}_{MB}$ and $(\frac{1}{n} \bm{F}_{MB} + \lambda I )^{-1}$, respectively, and it follows that

\begin{align*}
    \norm{\bm{J}_{MB}(k) \left( \bm{F}_{MB} + \lambda I \right) ^{-1}  \bm{J}_{MB}(k)^\top}_2 &= \lambda_{max} \left(\bm{J}_{MB}(k) \left(\bm{F}_{MB} + \lambda I \right) ^{-1}  \bm{J}_{MB}(k)^\top\right)\\
    &= \max_{\mu \text{ eigenvalue of } \bm{G}_{MB}(k)} \frac{n \mu}{\mu + n \lambda} \leq n.
    \end{align*}

Let us consider the function $\lambda \overset{f}{\to} f(\lambda) := \left(\frac{2 K \lambda_0}{\lambda_0 + \frac{9}{4}n\lambda} - \frac{2 C \sqrt{\lambda_0 K}}{3 \sqrt{\lambda n}} -1\right)$. We have that 

\begin{align*}
  f(\frac{4\lambda_0}{9n})  = K - C \sqrt{K} - 1 \geq K-\frac{1}{2} \sqrt{K} -1 >0\quad \text{for }K \geq 3.
\end{align*}

Thereforem by continuity of the function $f(.)$, there exists an interval $[\underline{\lambda}, \overline{\lambda}]$, such as $\frac{4\lambda_0}{9n}\in [\underline{\lambda}, \overline{\lambda}]$, and for all damping values $\lambda$ in $[\underline{\lambda}, \overline{\lambda}]$, the function $f(.)$ is positive. For such choice of damping value $\lambda$ (for example $\lambda = \frac{4\lambda_0}{9n}$), and for a small enough learning rate, i.e:
\begin{align*}
    \eta \leq \frac{\frac{2 K \lambda_0}{\lambda_0 + \frac{9}{4}n\lambda} - \frac{2 C \sqrt{\lambda_0 K}}{3 \sqrt{\lambda n}} -1}{\left(K + \frac{C \sqrt{\lambda_0 K}}{3 \sqrt{\lambda n}}\right)^2} := \eta_{\lambda}.
\end{align*}
We
Hence, we get that
\begin{align*}
    || \bm{u}(k+1) - \bm{y}||_2^2 \leq (1-\eta) \norm{(\bm{u}(k) - \bm{y}))}_2^2,
\end{align*}
which concludes the proof.

\section{Motivation for kernel-wise mini-blocks choice in convolutional layers} \label{apx_cnn_motivation}

We recall from the main manuscript the following assumptions and notation for a single convolutional layer from the CNN with trainable parameters (i.e. weights $W$ and biases $b$) :
\begin{enumerate}[topsep=0pt,itemsep=-1ex,partopsep=1ex,parsep=1ex]
\item
the convolutional layer is 2-dimensional;
\item 
the layer has $J$ input channels indexed by $j = 1, ..., J$, $I$ output channels indexed by $i = 1, ..., I$;
\item 
there are $I \times J$ filters, each of size $(2R+1) \times (2R+1)$, with spatial offsets from the centers of each filter indexed by $\delta \in {\Delta} := \{ -R, ..., R \} \times \{ -R, ..., R \}$;

\item
the stride is of length 1, and the padding is equal to $R$, so that the sets of input and output spatial locations ($t \in \mathcal{T} \subset \mathbf{R}^2$) are the same.\footnote{The derivations in this paper can also be extended to the case where stride is greater than 1.};
\end{enumerate}

{The weights $W$, corresponding to the elements of all of the filters in this layer, can be viewed as a 3-dimensional tensor of size $I \times J \times \Delta$, where $\Delta = (2R+1)^2$. We shall use $I$, $J$ and $\Delta$ to denote both sets of indices and the cardinalities of these sets. Each element of $W$ is denoted by $W_{i,j,\delta}$, where the first two indices $i,j$ are the output/input channels, and the third index $\delta$ specifies the spatial offset within a filter as indicated in item 3 above. The bias $b$ is a vector of length $I$.} 

For the weights and biases, we define the vectors
\begin{align*}
    \vw_{i}
    & : = \left( w_{i,1,\delta_1}, ..., w_{i,J,\delta_{|\Delta|}}, b_i \right)^\top \in \mathbb{R}^{J|\Delta|+1},
\end{align*}
for $i = 1, ..., I$, 
and from them the matrix

{
$$
{W} := ({\vw}_1, ..., {\vw}_I)^\top \in \mathbb{R}^{I \times (J|\Delta|+1)}. \quad (1)
$$
}
We shall also express the vectors $\vw_i$ as
\begin{align*}
    & \vw_{i} := \left( \hat \vw_{i,1}^\top, ..., \hat \vw_{i,J}^\top, b_i  \right)^\top \in \mathbb{R}^{J\Delta+1}, \; \;  \forall \; i \in I,\\
    \text{where} &\\
    & \hat{\vw}_{i,j} := (\vw_{i,1,j}, \ldots, \vw_{i,\Delta,j})^\top
    \in \mathbb{R}^{\Delta}, \;  \; \forall \;  i \in I,
    \; j \in J.
\end{align*}
Let the vector
$\va := \{ a_{1, t}, \ldots, a_{J, t}\}$,
where $a_{j, t}$,
denotes the input from channel $j$ of the previous layer to the current layer after padding is added, where $t$ denotes the spatial location of the padded input. Note that the index pairs $t \in \mathcal{T} \subset \mathbf{R}^2 $ can be ordered, for example, lexicographically, into a one dimensional set of $\Delta$ indices. 

 It is useful to expand each component $a_{j,t}$ of $\va$  to a $\Delta$-dimensional vector
 $\hat{\va}_{j,t}$, that includes all components in the input $\va$ covered by the filter centered at $t$,
 yielding the following vectors defined for all locations $t \in \mathcal{T}$:
\begin{align*}
    & \va_{t} := \left( \hat \va_{1,t}^\top, ..., \hat \va_{J,t}^\top, 1 \right)^\top \in \mathbb{R}^{J\Delta+1},\\ 
    \text{where} &\\
    & \hat{\va}_{j,t} := (\va_{j,1,t}, \ldots, \va_{j,\Delta,t})^\top
    \in \mathbb{R}^{\Delta}, \;  \; \forall \; j \in J\\
    \text{hence} &\\
     & \va_{t} := \left( a_{1,t+\delta_1}, ..., a_{J,t+\delta_{|\Delta|}}, 1 \right)^\top \in \mathbb{R}^{J|\Delta|+1}.
\end{align*}
Note that a single homogeneous coordinate is concatenated at the end of $\va_t$. Expressing the pre-activation output for the layer at spatial location $ t \in \mathcal{T}$ as a vector of length equal to the number of output channels, i.e., 
$$\vh_{t} := \left( h_{1,t}, ..., h_{I,t} \right)^\top \in \mathbb{R}^I,$$
for all spatial locations $ t \in \mathcal{T}$. We note that, given inputs $\va$ and $W$, the pre-activation outputs $\vh$ can be computed, for all locations $t \in \mathcal{T}$, as
\begin{align}
    h_{i,t} = \sum_{j=1}^J \sum_{\delta \in \Delta} w_{i,j,\delta} a_{j,t+\delta} + b_i,
    \quad t \in \mathcal{T}, \, i = 1, ..., I.
    \label{eq_12}
\end{align}
or equivalently, $\vh_t = W \va_t$, whose $i$-th component $h_{i,t}$ we can write as 

$$
h_{i,t} = \sum_{j \in J}\hat \vw_{i,j}^\top \hat \va_{j,t} + b_i.
\quad (2)
$$

Expressing the input-output relationship in a CNN this way, we see that it is analogous to the input-output relationship in a fully connected feed-forward NN, except that the role of input and output node sets $J$ and $I$ are taken on by the input and output channels
and the affine mapping of of the vector of inputs $\va$ to the vector of outputs $\vh$, 
$$h_i = \sum_{j\in J} w_{i,j} a_j + b_i, \quad \forall \; i \in I,$$
where the  the terms $w_{i,j}a_j$ are the products of two
scalars becomes in (2) the inner product of two $\Delta$-dimensional vectors, and this mapping is performed for all locations $t$.

Hence, MBF is analaous to using the squares of the components of the gradient in a ff-cc network, and hence is analagous to a "squared" version of an adaptive first-order method.

\section{Experiment Details}
\label{apx_exp}
\subsection{Competing Algorithms}
\label{apx_algos}

\subsubsection{SGD-m}
In SGD with momentum, we updated the momentum $m_t$ of the gradient using the recurrence $$m_t = \mu \cdot m_{t-1} + g_t$$ at every iteration, where $g_t$ denotes the mini-batch gradient at current iteration and $\mu = 0.9$. The gradient momentum is also used in the second-order methods, in our implementations. 
For the CNN problems, we used weight decay with SGD-m, as it is used in SGDW in \cite{loshchilov2018decoupled}.

\subsubsection{Adam}

For Adam, we followed exactly the algorithm in \cite{kingma2014adam} with $\beta_1 = 0.9$ and $\beta_2 = 0.999$, updating the momentum of the gradient at every iteration by the recurrence
$$m_t = \beta_1 \cdot m{t-1} + (1-\beta_1) \cdot g_t.$$
The role of $\beta_1$ and $\beta_2$ is similar to that of $\mu$ and $\beta$ in Algorithms \ref{algo_MBF_full} and  \ref{shampoo}, as we will describe below. 
For the CNN problems, we used weight decay with Adam, as it is used in AdamW in \cite{loshchilov2018decoupled}.

\subsubsection{Shampoo}
\label{sec_15}

We implemented Shampoo as described below in Algorithm \ref{shampoo}
following 
the description given in \cite{gupta2018shampoo}, and  includes major improvements, following the suggestions in \cite{anil2021scalable}. These improvements are (i) using a moving average to update the estimates $\widehat{G_l^{(i)}}$ and (ii) using a coupled Newton method to compute inverse roots of the preconditioning matrices,

\begin{algorithm}[H]
    \caption{Shampoo}
    \label{shampoo}
    \begin{algorithmic}[1]
    
    \REQUIRE
    Given batch size $m$, learning rate $\{ \eta_k \}_{k \ge 1}$, {weight decay factor $\gamma$,} damping value $\epsilon$, {statistics update frequency $T_1$, inverse update frequency $T_2$}
    
    \STATE $\mu = 0.9$, $\beta = 0.9$
    
    \STATE Initialize $\widehat{G_l^{(i)}} = \E [G^{(i)}_l]$ ($l = 1,..,k$, $i = 1,...,k_l$) by iterating through the whole dataset, $\widehat{\nabla_{W_l} \mathcal{L}}=0$ ($l=1,...,L$)
    
    \FOR {$k=1,2,\ldots$}
        \STATE Sample mini-batch $M_k$ of size $m$ 
        \STATE Perform a forward-backward pass over the current mini-batch $M_k$ to compute the minibatch gradient $\overline{\nabla \mathcal{L}}$
        \FOR {$l = 1, ... L$
        }
        
        \STATE $\widehat{\nabla_{W_l} \mathcal{L}} = \mu \widehat{\nabla_{W_l} \mathcal{L}} + \overline{\nabla_{W_l} \mathcal{L}}$
        
        \IF {$k \equiv 0 \pmod{T_1}$}
        
        \STATE Update $\widehat{G_l^{(i)}} = \beta \widehat{G_l^{(i)}} + (1-\beta) \overline{G_l}^{(i)}$ for $i = 1, ..., k_l$ where $\overline{G_l} = \overline{\nabla_{W_l} \mathcal{L}}$
        \label{line_1}
        
        \ENDIF

        \IF {$k \equiv 0 \pmod{T_2}$}
            
            \STATE 
            Recompute $\left( \widehat{G_l^{(1)}} + \epsilon I \right)^{-1/2 k_l}, ..., \left( \widehat{G_l^{(k_l)}} + \epsilon I \right)^{-1/2 k_l}$ with the coupled Newton method
            \label{line_5}
        
        \ENDIF

            \STATE $p_l = \widehat{\nabla_{W_l} \mathcal{L}} \times_1 \left( \widehat{G_l^{(1)}} + \epsilon I \right)^{-1/2 k_l} \times_2 \cdots \times_k \left( \widehat{G_l^{(k_l)}} + \epsilon I \right)^{-1/2 k_l}$
            
            \STATE {$p_l = p_l + \gamma W_l$}
            
            \STATE 
            $W_l = W_l - \eta_k \cdot p_{l}$
        \ENDFOR
        
        \ENDFOR  
    
    \end{algorithmic}
\end{algorithm}

\subsubsection{KFAC}
\label{sec_14}

In our implementation of KFAC, the preconditioning matrices that we used for linear layers and convolutional layers are precisely those described in \cite{martens2015optimizing} and \cite{grosse2016kronecker}, respectively. For the parameters in the BN layers, we used the gradient direction, exactly as in \url{https://github.com/alecwangcq/KFAC-Pytorch}. We did a warm start to estimate the pre-conditioning KFAC matrices in an initialization step that iterated through the whole data set, and adopted a moving average scheme to update them with $\beta=0.9$ afterwards. As in the implementation described in \cite{ren2021kronecker}, for autoencoder experiments, we inverted the damped KFAC matrices and used them to compute the updating direction, where the damping factors for both $A$ and $G$ were set to be $\sqrt{\lambda}$, where $\lambda$ is the overall damping value; and for the CNN experiments, we employed the SVD (i.e. eigenvalue decomposition) implementation suggested in \url{https://github.com/alecwangcq/KFAC-Pytorch}, which, as we verified, performs better than splitting the damping value and inverting the damped KFAC matrices (as suggested in \cite{martens2015optimizing,grosse2016kronecker}). Further, for the CNN problems, we implemented  weight decay exactly as in MBF (Algorithm \ref{algo_MBF_full}) and Shampoo (Algorithm \ref{shampoo}). 

\subsubsection{MBF, other details}\label{apx_other_details}

In Tables \ref{table_storage_fcc} and \ref{table_comp_fcc}, we compared the space and computational requirements of the proposed MBF method with KFAC and Adam for a fully connected layer, with $d_i$ inputs and $d_o$ outputs. Note that these tables are the fully-connected analogs to Table \ref{table_comp} in Section 7, which compare the storage and computational requirements for MBF, KFAC and Adam for a convolutional layer.
Here, $m$ denotes the size of the minibatches, and {$T_1$ and $T_2$} denote, respectively, the frequency for updating the preconditioners and inverting them for both KFAC and MBF.

\begin{table}[t]
  \caption{Storage Requirements for fully connected layer}
  \vskip 0.15in
  \label{table_storage_fcc}
  \centering
  \begin{tabular}{l|ccclll}
    \hline              
    Algorithm
    & $\mathcal{D} W$ & $P_l$ 
    \\
    \hline
    MBF
    & $O(d_i d_o)$
    & $O(d_i^2)$
    \\
    KFAC
    & $O(d_i d_o)$
    & $O(O(d_i^2 + d_o^2 + d_i d_o))$
    \\
    Shampoo
    & $O(d_i d_o)$
    & $O(O(d_i^2 + d_o^2))$
    \\
    Adam
    & $O(d_i d_o)$
    & $O(d_i d_o)$
    \\
    \hline
  \end{tabular}
\end{table}

\begin{table*}[t]
  \caption{Computation per iteration beyond that required for the minibatch stochastic gradient for fully connected layer}
  \vskip 0.15in
  \label{table_comp_fcc}
  \centering
  \begin{tabular}{l|cccllllll}
    \hline
    Algorithm
    & Additional pass
    & Curvature
    & Step $\Delta W_l$
    \\
    \hline
    MBF
    & ---
    & {$O(\frac{ d_o d_i^2}{T_1}   + \frac{d_i^3}{T_2}) $}
    & $O(d_o d_i^2 )$
    \\
    KFAC
    &
    {$O(\frac{m d_i d_o}{T_1} )$}
    &
    {$O(\frac{m d_i^2 + m d_o^2}{T_1} + \frac{d_i^3 + d_o^3}{T_2}  )$}
    & $O(d_i^2 d_o + d_o^2 d_i)$
    \\
    Shampoo
    &
    ---
    &
    {$O(\frac{ d_i^2 + d_o^2}{T_1} + \frac{d_i^3 + d_o^3}{T_2}  )$}
    & $O((d_i + d_o) d_i d_o)$
    \\
    Adam
    & ---
    & $O(d_i d_o)$
    & $O(d_i d_o)$
    \\
    \hline
  \end{tabular}
\end{table*}

For the parameters in the BN layers, we used the  direction used in Adam, which is equivalent to using mini-blocks of size 1, dividing each stochastic gradient component by that blocks square root. We did a warm start to estimate the pre-conditioning mini-block matrices in an initialization step that iterated through the whole data set, and adopted a moving average scheme to update them with $\beta=0.9$ afterwards as described in Algorithm \ref{algo_MBF_full}). 

\subsection{Experiment Settings for the Autoencoder Problems}
\label{apx_exp_fcc}

Table \ref{table_3} describes the model architectures of the autoencoder problems. The activation functions of the hidden layers are always ReLU, except that there is no activation for the very middle layer. 

\begin{table}[!ht]
  \caption{
  DNN architectures for the MLP autoencoder problems
  }
  \label{table_best_h_fcc}
  \vskip 0.15in
  \centering
  \begin{tabular}{cccccccccc}
    \toprule
    & Layer width
    \\
    \midrule
    MNIST
    &  [784, 1000, 500, 250, 30, 250, 500, 1000, 784]
    \\
    FACES
    & [625, 2000, 1000, 500, 30, 500, 1000, 2000, 625]
    \\
    CURVES
    &  [784, 400, 200, 100, 50, 25, 6, 25, 50, 100, 200, 400, 784]
    \\
    \bottomrule
  \end{tabular}
\end{table}

MNIST\footnote{\url{http://yann.lecun.com/exdb/mnist/}}, FACES\footnote{\url{http://www.cs.toronto.edu/~jmartens/newfaces_rot_single.mat}}, and CURVES\footnote{\url{http://www.cs.toronto.edu/~jmartens/digs3pts_1.mat}} contain 60,000, 103,500, and 20,000 training samples, respectively, which we used in our experiment to train the models and compute the training losses. 


We used binary entropy loss (with sigmoid) for MNIST and CURVES, and squared error loss for FACES.
The above settings largely mimic the settings in \cite{martens2010deep,martens2015optimizing,botev2017practical, ren2021tensor}. Since we primarily focused on optimization rather than generalization in these tasks, we also did not include $L_2$ regularization or weight decay. 

In order to obtain Figure \ref{fig_autoencoders}, we first conducted a grid search on the learning rate (lr) and damping value based on the criteria of minimizing the training loss. The ranges of the grid searches used for the algorithms in our tests are specified in Table \ref{table_hyper_fcc}.

\begin{table}[t]
  \caption{Grid of hyper-parameters for autoencoder problems}
  \vskip 0.15in
  \label{table_hyper_fcc}
  \centering
  \begin{tabular}{l|ccclll}
    \hline              
    Algorithm
    & learning rate & damping $\lambda$ 
    \\
    \hline
    SGD-m
    & 1e-4, 3e-4, 1e-3, 3e-3, 1e-2, 3e-2
    & damping: not applicable
    \\
    Adam
    & 1e-5, 3e-5, 1e-4, 3e-4, 1e-3, 3e-3, 1e-2
    & 1e-8, 1e-4, 1e-2
    \\
    Shampoo
    & 1e-5, 3e-5, 1e-4, 3e-4, 1e-3, 3e-3
    & 1e-4, 3e-4, 1e-3, 3e-3, 1e-2
    \\
    MBF
    & 1e-7, 3e-7, 1e-6, 3e-6, 1e-5, 3e-5, 1e-4
    & 1e-5, 3e-5, 1e-4, 3e-4, 1e-3, 3e-3, 1e-2
    \\
    KFAC
    & 1e-4, 3e-4, 1e-3, 3e-3, 1e-2, 3e-2, 1e-2, 3e-2
    & 1e-2, 3e-2, 1e-1, 3e-1, 1e0, 3e0, 1e1
    \\
    \hline
  \end{tabular}
\end{table}






The best hyper-parameter values determined by our grid searches are listed in Table \ref{table_best_h_fcc}. 

\begin{table}[ht]
  \caption{Hyper-parameters (learning rate, damping) used to produce Figure \ref{fig_autoencoders}}
  \label{table_3}
  \centering
  \begin{tabular}{lllll}
    \toprule
    Name & MNIST & FACES & CURVES & \\
    \midrule
    MBF & (1e-5, 3e-4) $\to$ 51.49 & (1e-6, 3e-3) $\to$ 5.17 & (1e-5, 3e-4) $\to$ 55.14\\
    KFAC & (3e-3, 3e-1) $\to$ 53.56 & (1e-1, 1e1) $\to$ 5.55 & (1e-2, 1e0) $\to$ 56.47\\
    Shampoo & (3e-4, 3e-4) $\to$ 53.80 & (3e-4, 3e-4) $\to$ 7.21 & (1e-3, 3e-3) $\to$ 54.86\\
    Adam & (3e-4, 1e-4) $\to$ 53.67 & (1e-4, 1e-4) $\to$ 5.55 & (3e-4, 1e-4) $\to$ 55.23\\
    SGD-m & (3e-3, -) $\to$ 55.63 & (1e-3, -) $\to$ 7.08  & (1e-2, -) $\to$ 55.49 \\
    \bottomrule
  \end{tabular}
\end{table}

\subsection{Experiment Settings for the CNN Problems}
\label{apx_exp_ccn}



The ResNet32 model refers to the one in Table 6 of \cite{he2016deep}, whereas the VGG16 model refers to model D of \cite{simonyan2014very}, with the modification that batch normalization layers were added after all of the convolutional layers in the model. For all algorithms, we used a batch size of 128 at every iteration.


We used weight decay for all the algorithms that we tested, which is related to, but not the same as $L_2$ regularization added to the loss function, and has been shown to help improve generalization performance across different optimizers \cite{loshchilov2018decoupled,zhang2018three}. 
The use of weight decay for MBF and Shampoo is implemented in lines 16 and 17 in Algorithm \ref{algo_MBF_full} and in lines 15 and 16 in  Algorithm \ref{shampoo}, respectively, and is similarly applied to SGD-m , Adam, and KFAC.


For MBF, we set $\lambda = 0.003$. We also tried values around 0.003 and the results were not sensitive to the value of $\lambda$. Hence, $\lambda$ can be set to $0.003$ as a default value.
For KFAC, we set the overall damping value to be 0.03, as suggested in the implementation in \url{https://github.com/alecwangcq/KFAC-Pytorch}. We also tried values around 0.03 for KFAC and confirmed that 0.03 is a good default value. 

In order to obtain Figure \ref{fig_cnns}, we first conducted a grid search on the initial learning rate (lr) and weight decay (wd) factor based on the criteria of maximizing the classification accuracy on the validation set. The range of the grid searches for the algorithms in our tests are specified in Table \ref{table_hyper_cnn}.

\begin{table}[!h]
  \caption{Grid of hyper-parameters for CNN problems}
  \vskip 0.15in
  \label{table_hyper_cnn}
  \centering
  \begin{tabular}{l|ccclll}
    \hline              
    Algorithm
    & learning rate & weight decay $\gamma$ 
    \\
    \hline
    SGD-m
    & 3e-5, 1e-4, 3e-4, 1e-3, 3e-3, 1e-2, 3e-2, 1e-1, 3e-1, 1e0
    & 1e-2, 3e-2, 1e-1, 3e-1, 1e0, 3e0, 1e1
    \\
    Adam
    & 1e-6, 3e-6, 1e-5, 3e-5, 1e-4, 3e-4, 1e-3, 3e-3, 1e-2, 3e-2
    & 1e-2, 3e-2, 1e-1, 3e-1, 1e0, 3e0, 1e1
    \\
    Shampoo
    & 3e-5, 1e-4, 3e-4, 1e-3, 3e-3, 1e-2, 3e-2, 1e-1
    & 1e-2, 3e-2, 1e-1, 3e-1, 1e0, 3e0, 1e1
    \\
    MBF
    &  1e-6, 3e-6, 1e-5, 3e-5, 1e-4, 3e-4, 1e-3, 3e-3
    & 1e-2, 3e-2, 1e-1, 3e-1, 1e0, 3e0, 1e1
    \\
    KFAC
    & 3e-6, 1e-5, 3e-5, 1e-4, 3e-4, 1e-3, 3e-3, 1e-2, 3e-2
    & 1e-2, 3e-2, 1e-1, 3e-1, 1e0, 3e0, 1e1
    \\
    \hline
  \end{tabular}
\end{table}







The best hyper-parameter values, and the validation classification accuracy obtained using them, are listed in Table \ref{table_conv_best}.

\begin{table}[!h]
  \caption{
  Hyper-parameters ({initial} learning rate, weight decay factor) used to produce Figure \ref{fig_cnns} and the average validation accuracy across 5 runs with different random seeds shown in Figure \ref{fig_cnns}
  }
  \label{table_conv_best}
  \centering
  \begin{tabular}{lllll}
    \toprule
    Name & CIFAR-10 + ResNet32 & CIFAR-100 + VGG16 & SVHN + VGG11 & \\
    \midrule
    MBF & (1e-4, 3e0) $\to$ 93.42\% & (3e-5, 1e1) $\to$ 74.80\% & (1e-3, 3e-1) $\to$ 96.59\%
    \\
    KFAC & (3e-3, 1e-1) $\to$ 93.02\% & (1e-3, 3e-1) $\to$ 74.38\% & (3e-3, 1e-1) $\to$ 96.37\%
    \\
    Shampoo & (1e-2, 1e-1) $\to$ 92.97\% & (1e-3, 3e-1) $\to$ 73.37\% & (3e-3, 1e-1) $\to$ 96.15\%
    \\
    Adam & (3e-3, 1e-1) $\to$ 93.34\% & (3e-5, 1e1) $\to$ 72.95\% & (3e-4, 1e0) $\to$ 96.34\%
    \\
    SGD-m & (1e-1, 1e-2) $\to$ 93.23\% & (3e-2, 1e-2) $\to$ 73.99\% & (3e-2, 1e-2) $\to$ 96.63\%
    \\
    \bottomrule
  \end{tabular}
\end{table}

\subsection{More on MBF Implementation Motivations}

\subsubsection{Details on the Cosine similarity experiment}\label{apx_cosine}

We provide in Algorithm \ref{algo_MBF_true} the full implementation of MBF-True for completeness. Note that, in MBF-True, the only difference with MBF is that we are using the mini-batch gradient $\overline{\mathcal{D}_2 W_{l, b}}$ (denoted by $\mathcal{D}_2$ )of the model on sampled labels $y_t$ from the model's distribution (see lines 10-13 in Algorithm \ref{algo_MBF_true}) to update the estimate of mini-block preconditioners, using a moving average (lines 12, 13), with a rank one outer-product, which is different from computing the true Fisher for that mini-block. 

\begin{algorithm}[!h]
\caption{MBF-True}
    \label{algo_MBF_true}
    \begin{algorithmic}[1]
    \REQUIRE
    Given batch size $m$, learning rate $\{ \eta_k \}_{k \ge 1}$, {weight decay factor $\gamma$,} damping value $\lambda$, statistics update frequency $T_1$, inverse update frequency $T_2$
    \STATE $\mu = 0.9$, $\beta = 0.9$
    \STATE Initialize $\widehat{G_{l, b}} = \E [G_{l, b}]$ ($l = 1,..,k$, mini-blocks $b$) by iterating through the whole dataset, $\widehat{\mathcal{D} W_{l, b}}=0$ ($l = 1,..,k$, mini-blocks $b$)
    \FOR {$k=1,2,\ldots$}
        \STATE Sample mini-batch $M_t$ of size $m$ 
        \STATE Perform a forward-backward pass over $M_t$ to compute the mini-batch gradient $\overline{\mathcal{D} W_{l, b}}$
        \FOR {$l = 1, ... L$}
        \FOR{mini-block $b$ in layer $l$, \textbf{in parallel}}
        \STATE $\widehat{\mathcal{D} W_{l, b}} = \mu \widehat{\mathcal{D} W_{l, b}} + \overline{\mathcal{D} W_{l, b}}$
        \IF {$k \equiv 0 \pmod{T_1}$}
            \STATE Sample the labels $y_t$ from the model's distribution
            \STATE Perform a backward pass over $y_t$ to compute the mini-batch gradients $\overline{\mathcal{D}_2 W_{l, b}}$
            \STATE If Layer $l$ is convolutional: $\widehat{G_{l, j,i}} = \beta \widehat{G_{l, j,i}} + (1-\beta)\overline{\mathcal{D}_2 W_{l, j,i}} \left(\overline{\mathcal{D}_2 W_{l, j,i}}\right)^\top$ 
            \STATE If Layer $l$ is fully-connected: $\widehat{G_{l}} = \beta \widehat{G_{l}} +  \frac{1-\beta}{O} \sum_{j = 1}^{O}\overline{\mathcal{D}_2 W_{l, j}} \left(\overline{\mathcal{D}_2 W_{l, j}}\right)^\top$
        \ENDIF
        \IF {$k \equiv 0 \pmod{T_2}$}
            \STATE
            Recompute and store $(\widehat{G_{l, b}} + \lambda I)^{-1}$
        \ENDIF
        \STATE $p_{l, b} = (\widehat{G_{l, b}} + \lambda I)^{-1} \widehat{\mathcal{D} W_{l, b}} + \gamma W_{l, b}$ 
        \STATE $W_{l, b} = W_{l, b} - \eta_k p_{l, b}$
        \ENDFOR
        \ENDFOR
        \ENDFOR  
    \end{algorithmic}
\end{algorithm}

As mentioned in the main manuscript, we explored how close MBF's direction is to the one obtained by a block-diagonal full EFM method (that we call BDF). We provide here a detailed implementation of the procedure that we used for completeness. More specifically, for any algorithm X, we reported the cosine similarity between the direction given by X and that obtained by BDF in the procedure described in Algorithm \ref{algo_cosine}.

\begin{algorithm}[!h]
\caption{Cosine(BDF, Algorithm X)}
    \label{algo_cosine}
    \begin{algorithmic}[1]
    \REQUIRE All required parameters for Algorithm X
    \STATE $m = 1000, \eta = 0.01, \mu = 0.9$, $\beta = 0.9, \lambda = 0.01$
    \STATE Initialize the block EFM matrices $\widehat{F_{l}} = \E [F_{l}]$ ($l = 1,..,L$) by iterating through the whole dataset
    \STATE $\widehat{\mathcal{D} W_{l}}=0$ ($l = 1,..,L$)
    \FOR {$k=1,2,\ldots$}
        \STATE Sample mini-batch $M_t$ of size $m$ 
        \STATE Perform a forward-backward pass over $M_t$ to compute the mini-batch gradient $\overline{\mathcal{D} W_{l}}$
        \FOR {$l = 1, ... L$}
        \STATE $\widehat{\mathcal{D} W_{l}} = \mu \widehat{\mathcal{D} W_{l}} + \overline{\mathcal{D} W_{l}}$
        \STATE $\widehat{F_{l}} = \beta \widehat{F_{l}} + (1-\beta)  \E [F_{l}]$
        \STATE $p_{l} = (\widehat{F_{l}} + \lambda I)^{-1} \widehat{\mathcal{D} W_{l, b}}$
        \STATE Compute the direction $d_{l}$ given by algorithm X at the current iterate $W_{l}$
        \STATE Compute and store the cosine $\frac{|p_{l}^T d_{l}|}{\norm{p_{l}} \norm{d_{l}}}$
        \STATE $W_{l} = W_{l} - \eta p_{l}$
        \ENDFOR
        \ENDFOR  
    \end{algorithmic}
\end{algorithm}

The algorithms were run on a $16 \times 16$ down-scaled MNIST \cite{lecun2010mnist} dataset and a small feed-forward NN with layer widths 256-20-20-20-20-20-10 described in \cite{martens2015optimizing}.  For all methods, we followed the trajectory obtained using the BDF method as described in Algorithm \ref{algo_cosine}.

\subsubsection{Comparison between MBF and MBF-True on Autoencoder and CNN problems}\label{apx_mbftrue_vs_mbf}

\begin{figure}[!h]
\centering
\subfigure[\footnotesize MNIST autoencoder]{\includegraphics[width=0.22\textwidth]{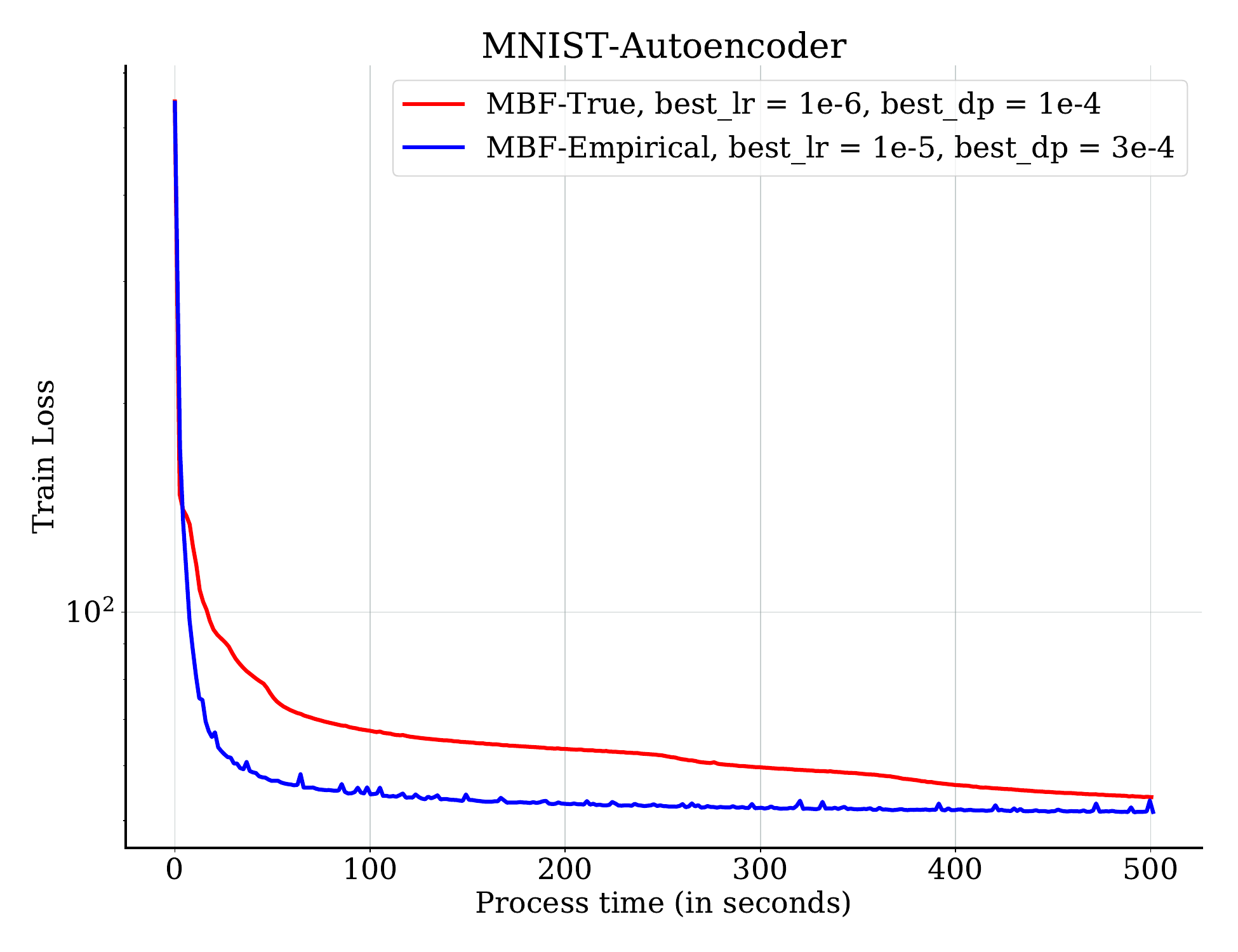} \quad \includegraphics[width=0.22\textwidth]{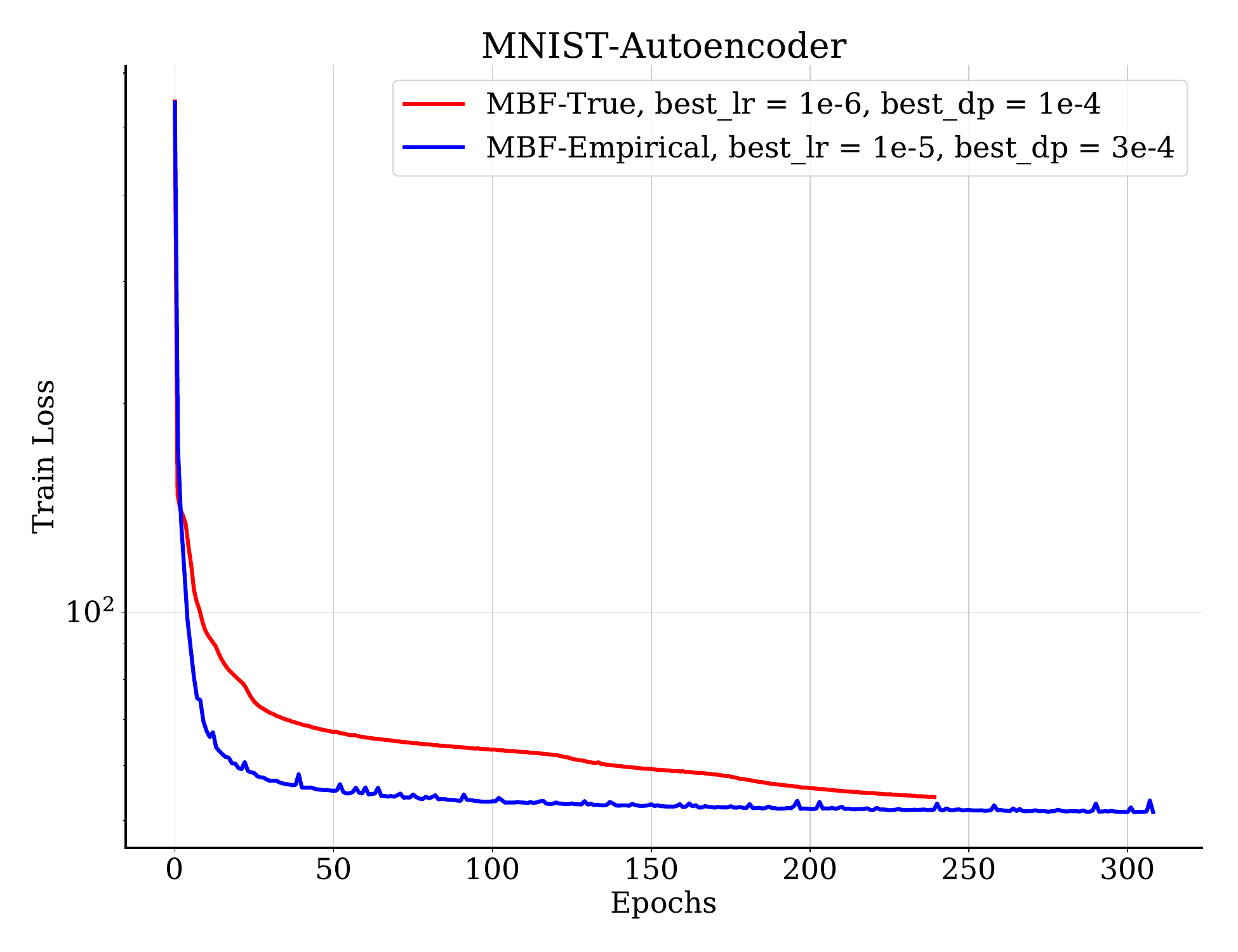}}\quad
  \subfigure[\footnotesize FACES autoencoder]{\includegraphics[width=0.22\textwidth]{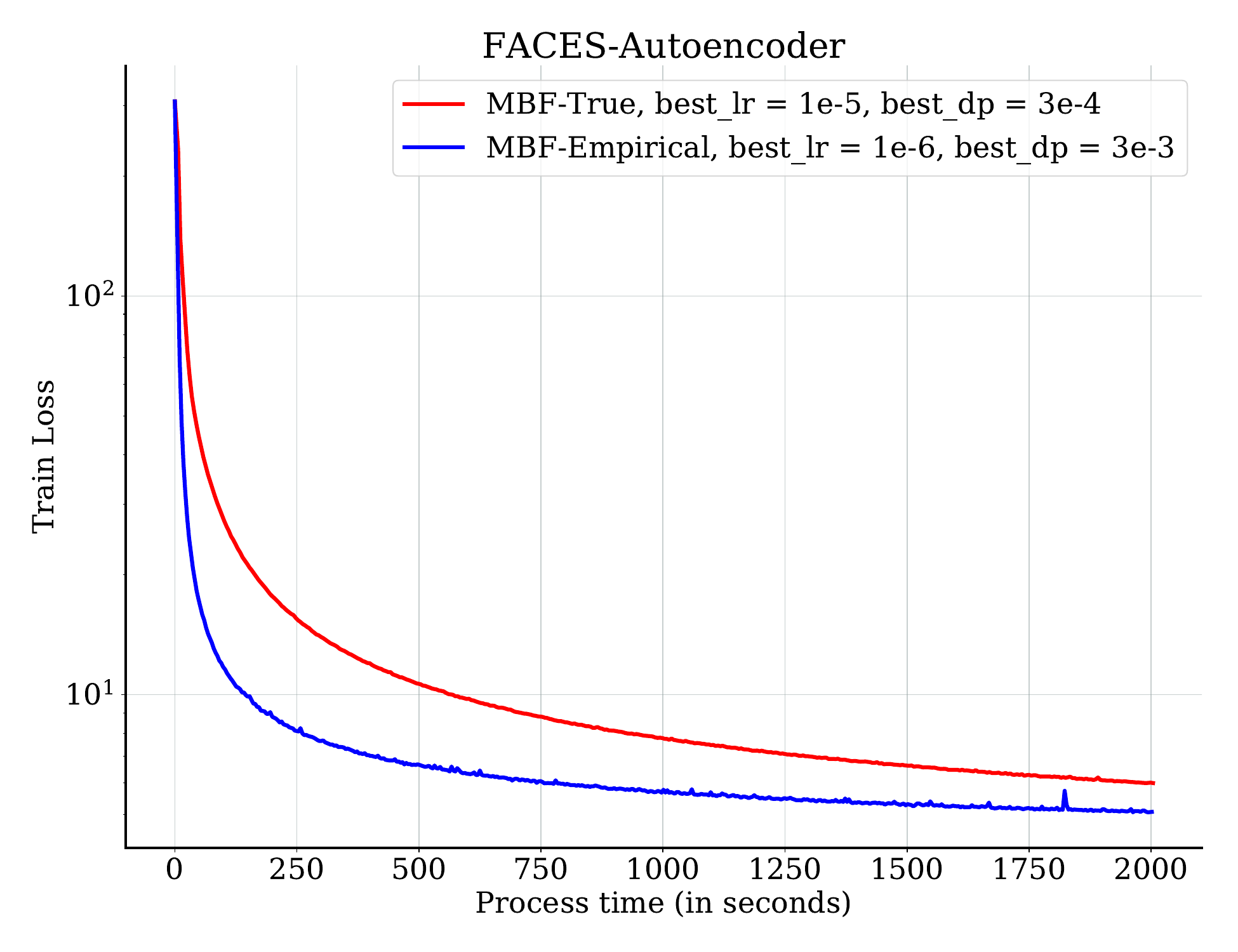} \quad \includegraphics[width=0.22\textwidth]{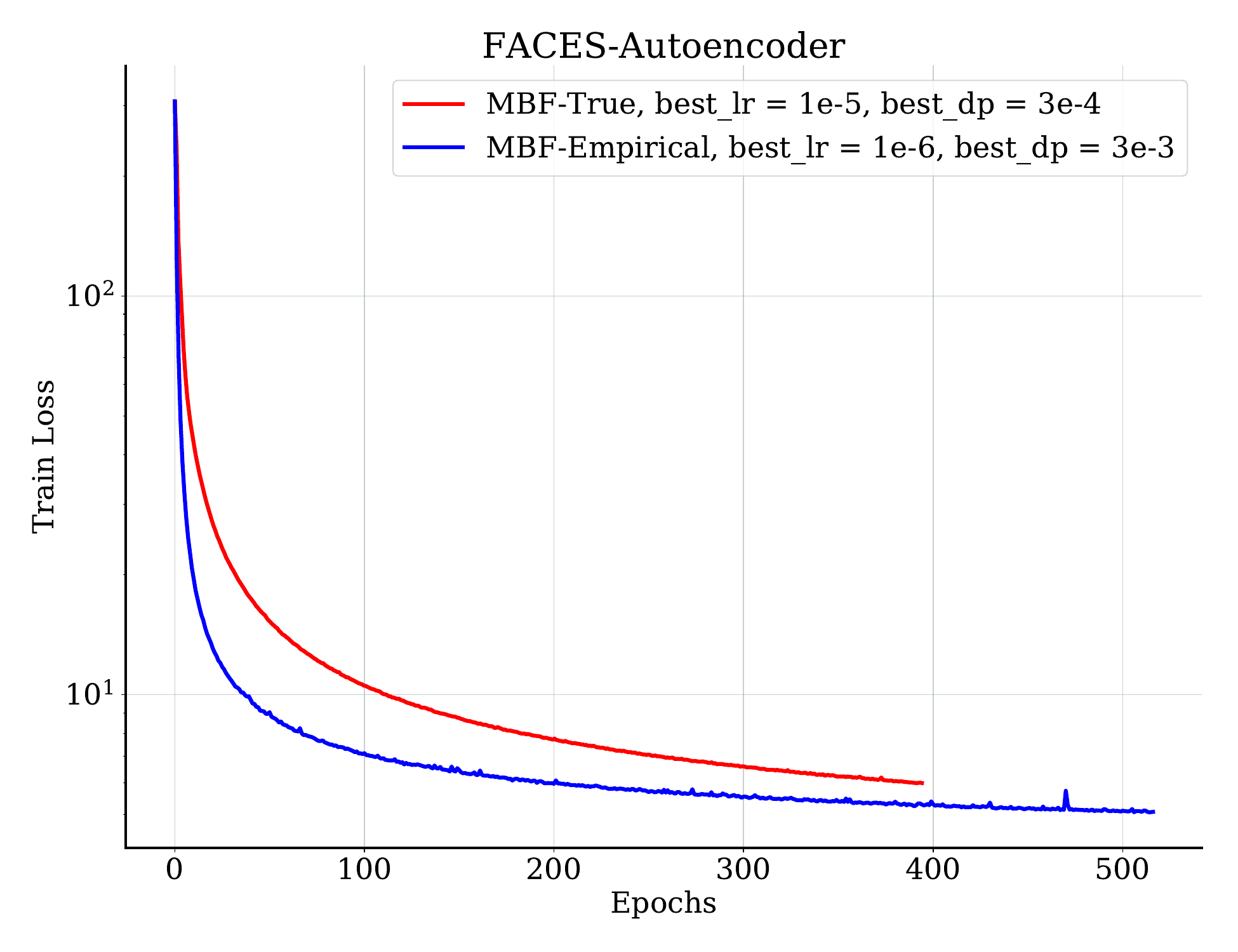}}\\
  \subfigure[\footnotesize CURVES autoencoder]{\includegraphics[width=0.22\textwidth]{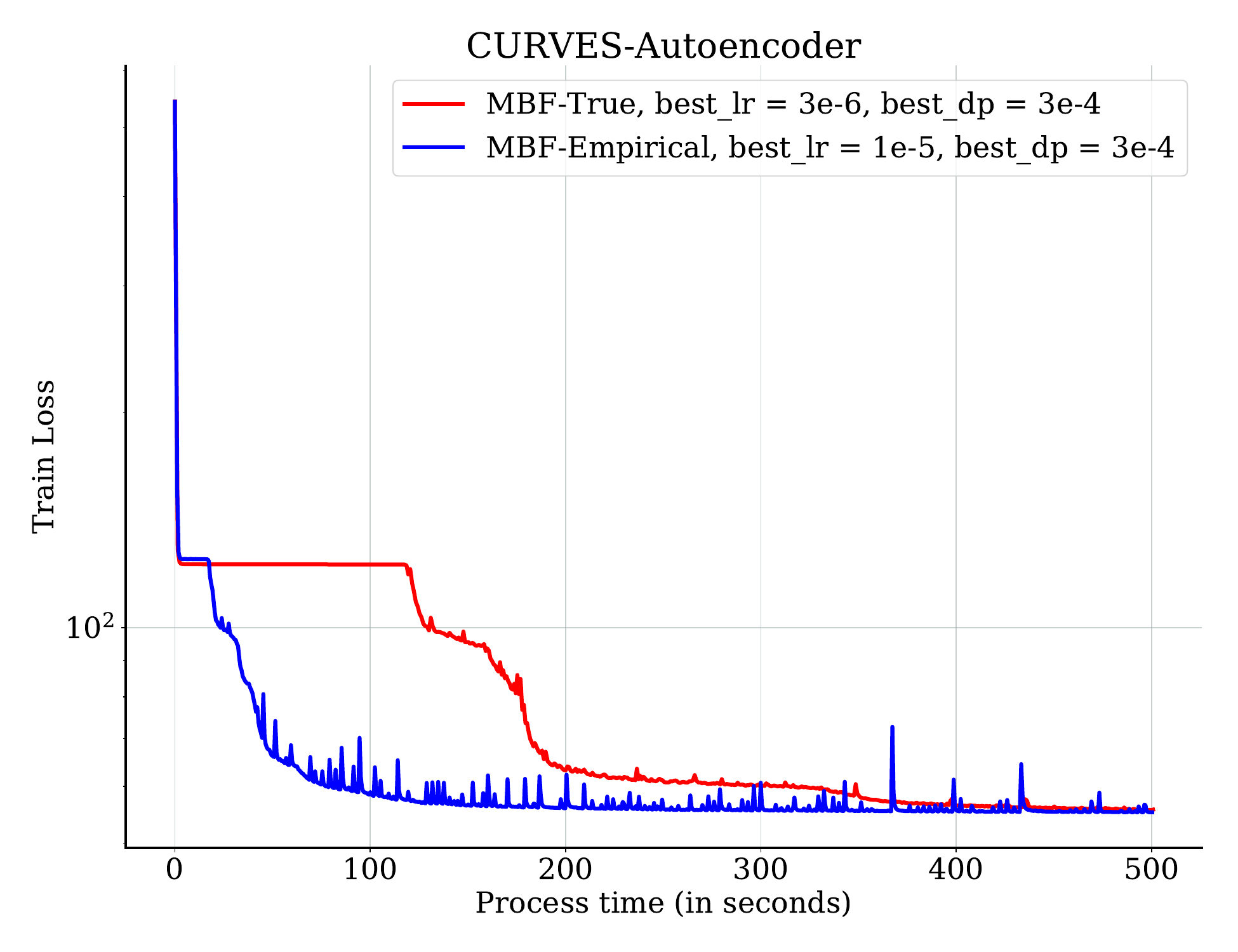} \quad \includegraphics[width=0.22\textwidth]{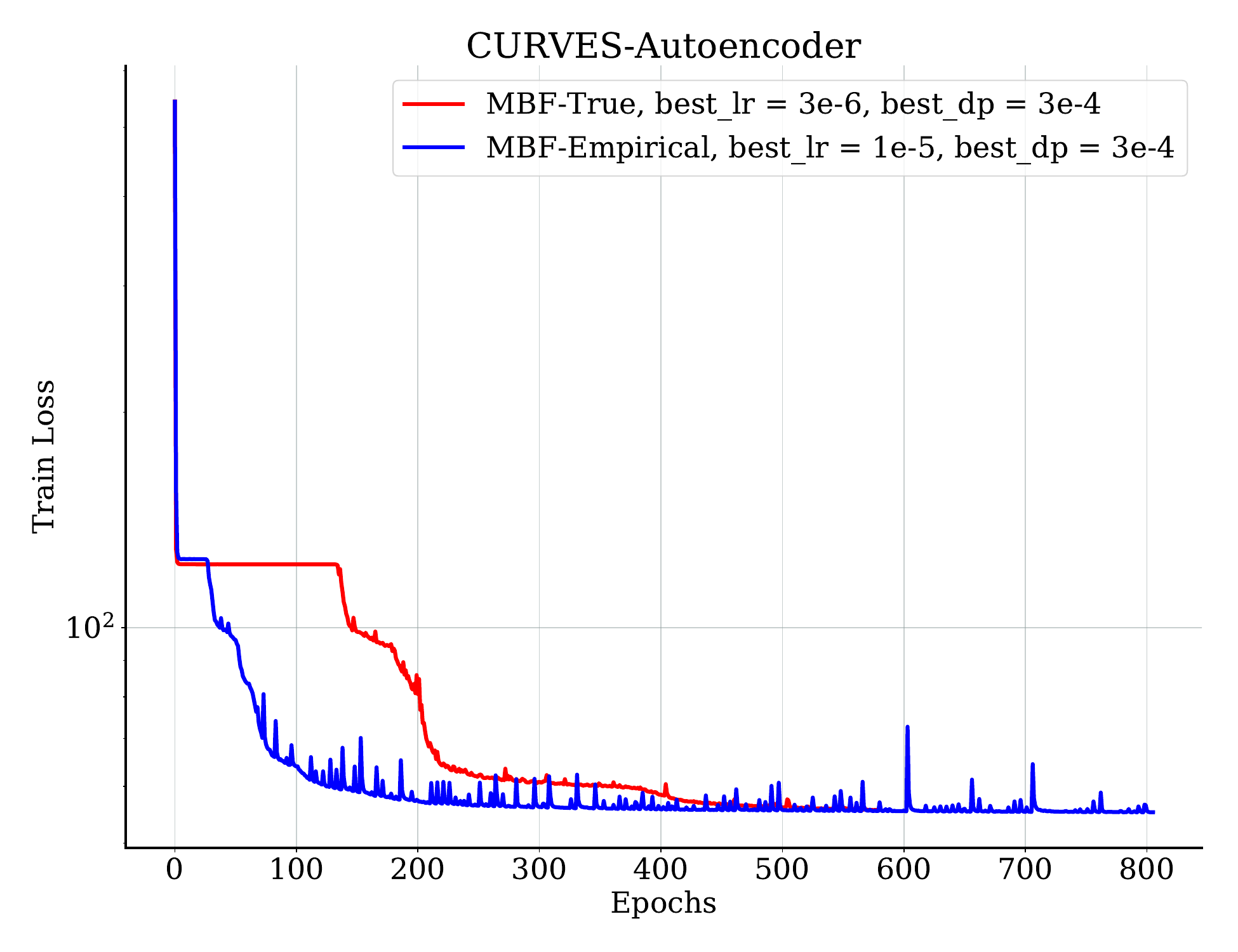}}
    \caption{
    Training performance of MBF-True and MBF on three autoencoder problems.
    }
    \label{fig_autoencoders_true}
\end{figure}

\begin{figure}[!h]
\centering
\subfigure[\footnotesize CIFAR-10, ResNet-32]{\includegraphics[width=0.22\textwidth]{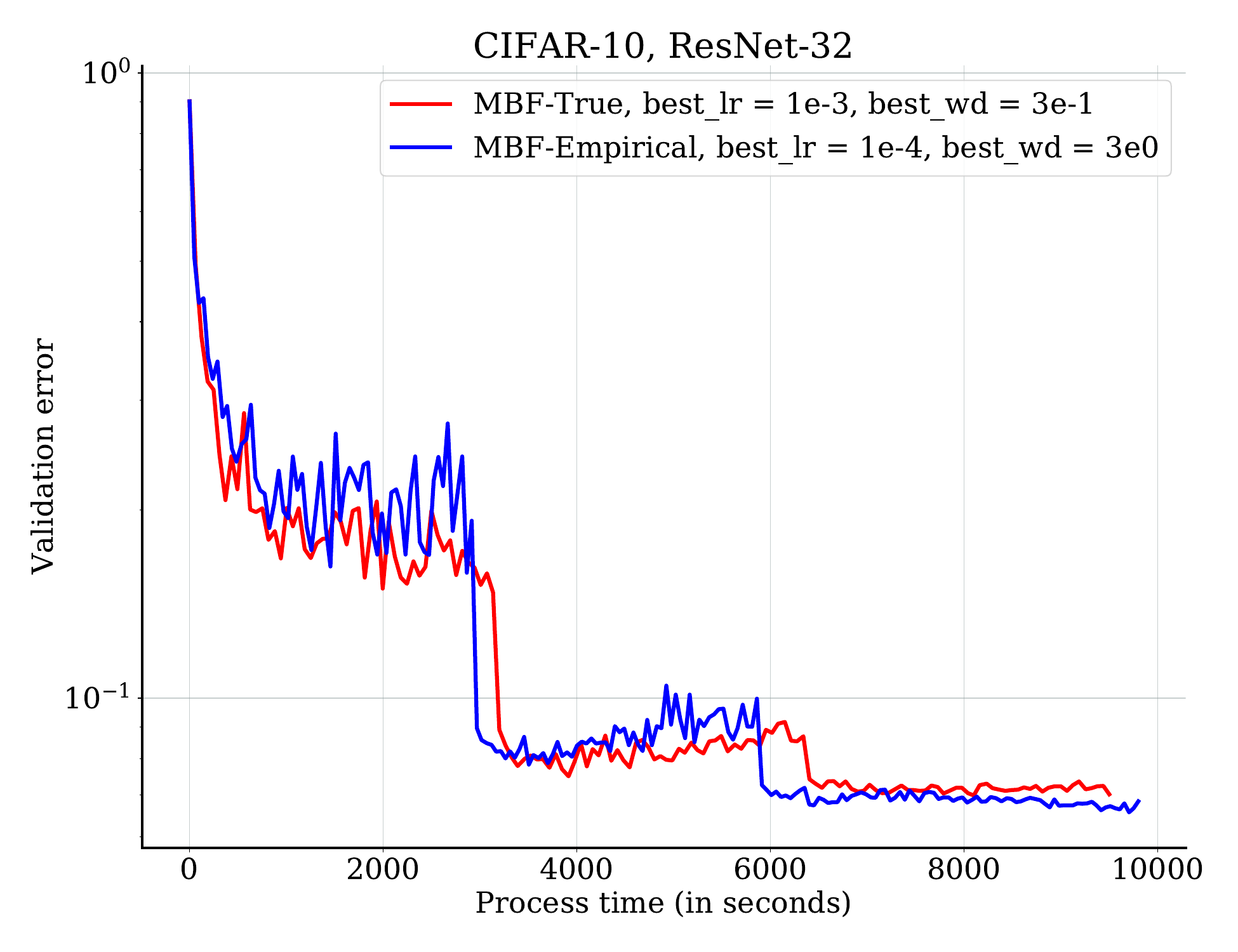} \quad \includegraphics[width=0.22\textwidth]{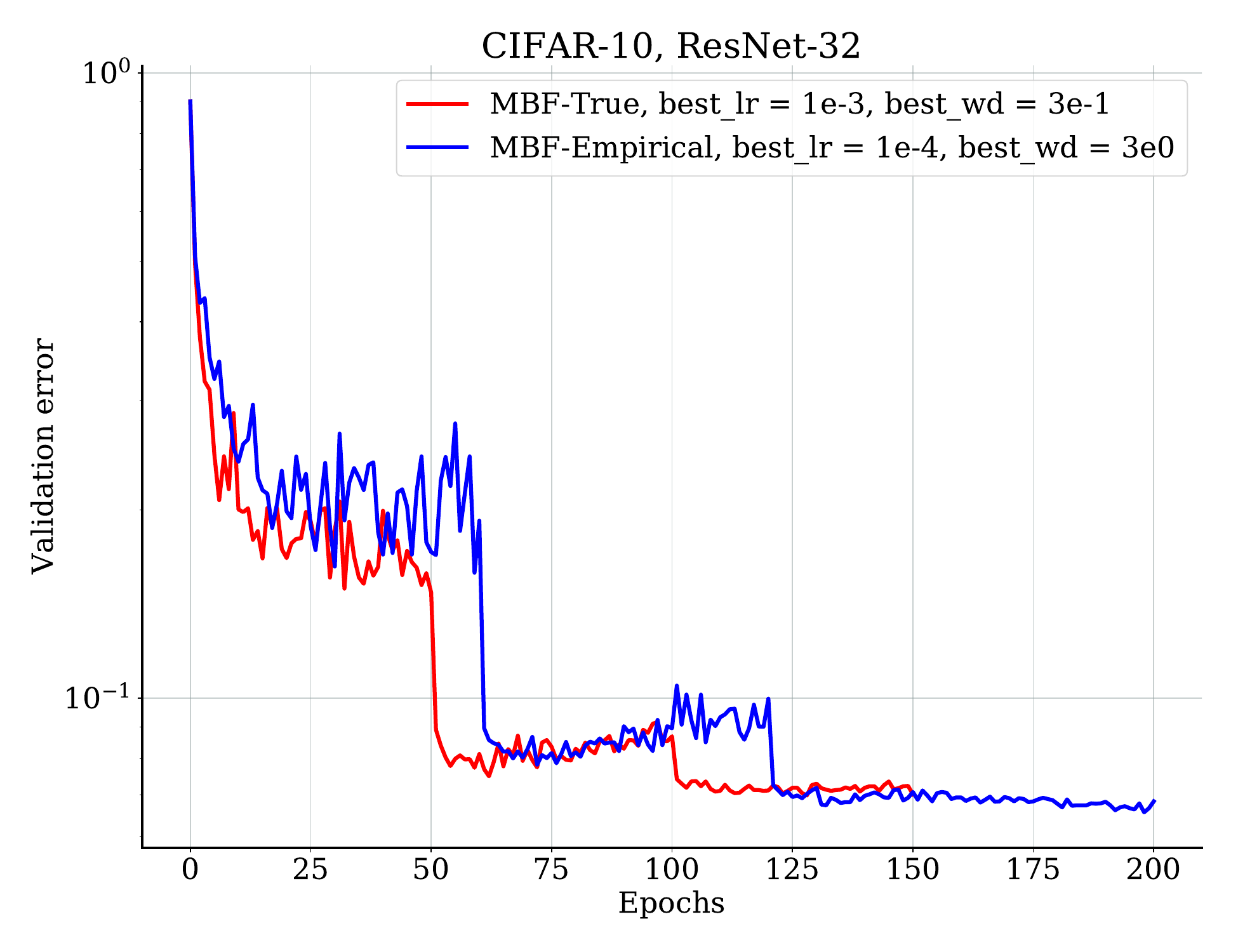}}\quad
  \subfigure[\footnotesize CIFAR-100, VGG-16]{\includegraphics[width=0.22\textwidth]{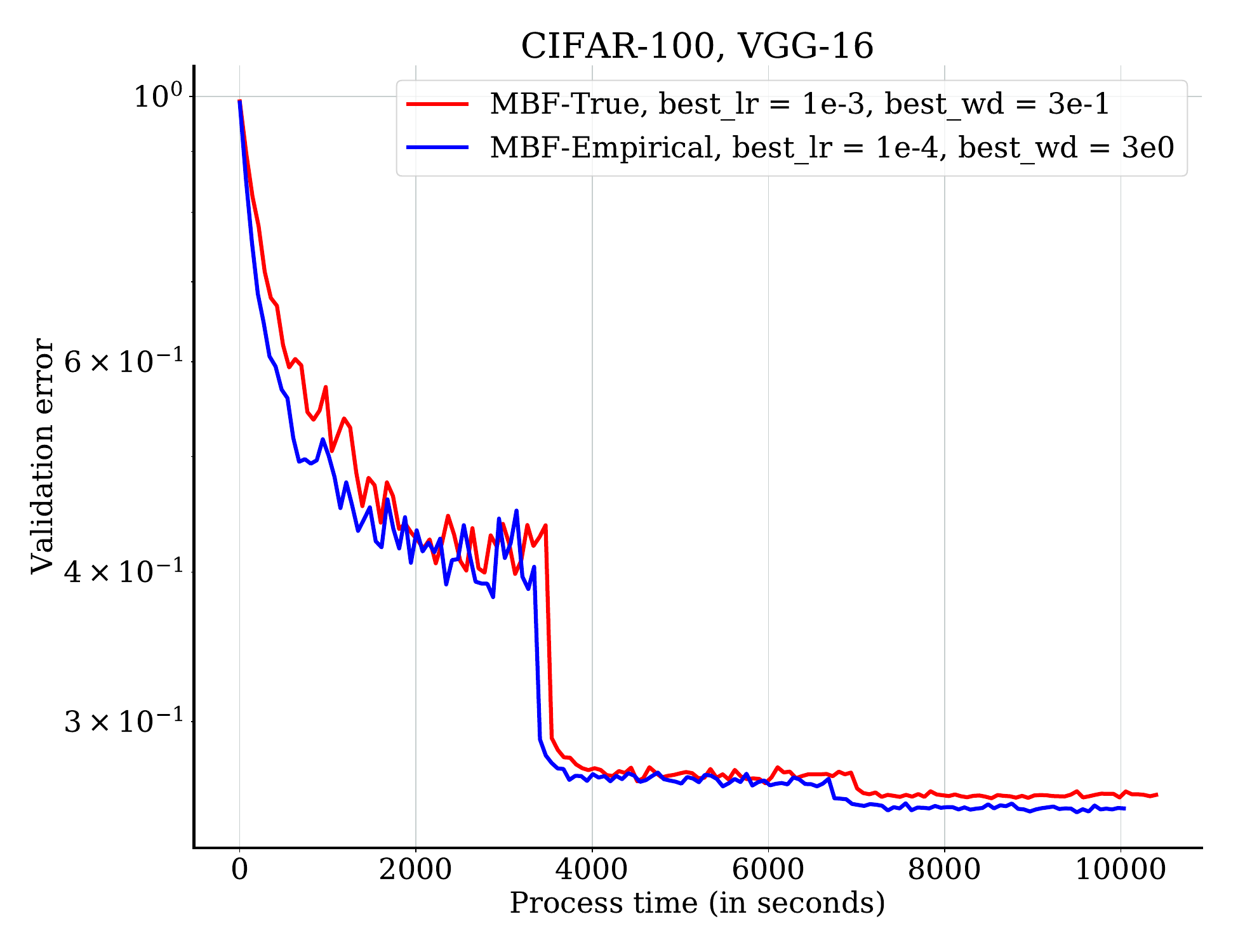} \quad \includegraphics[width=0.22\textwidth]{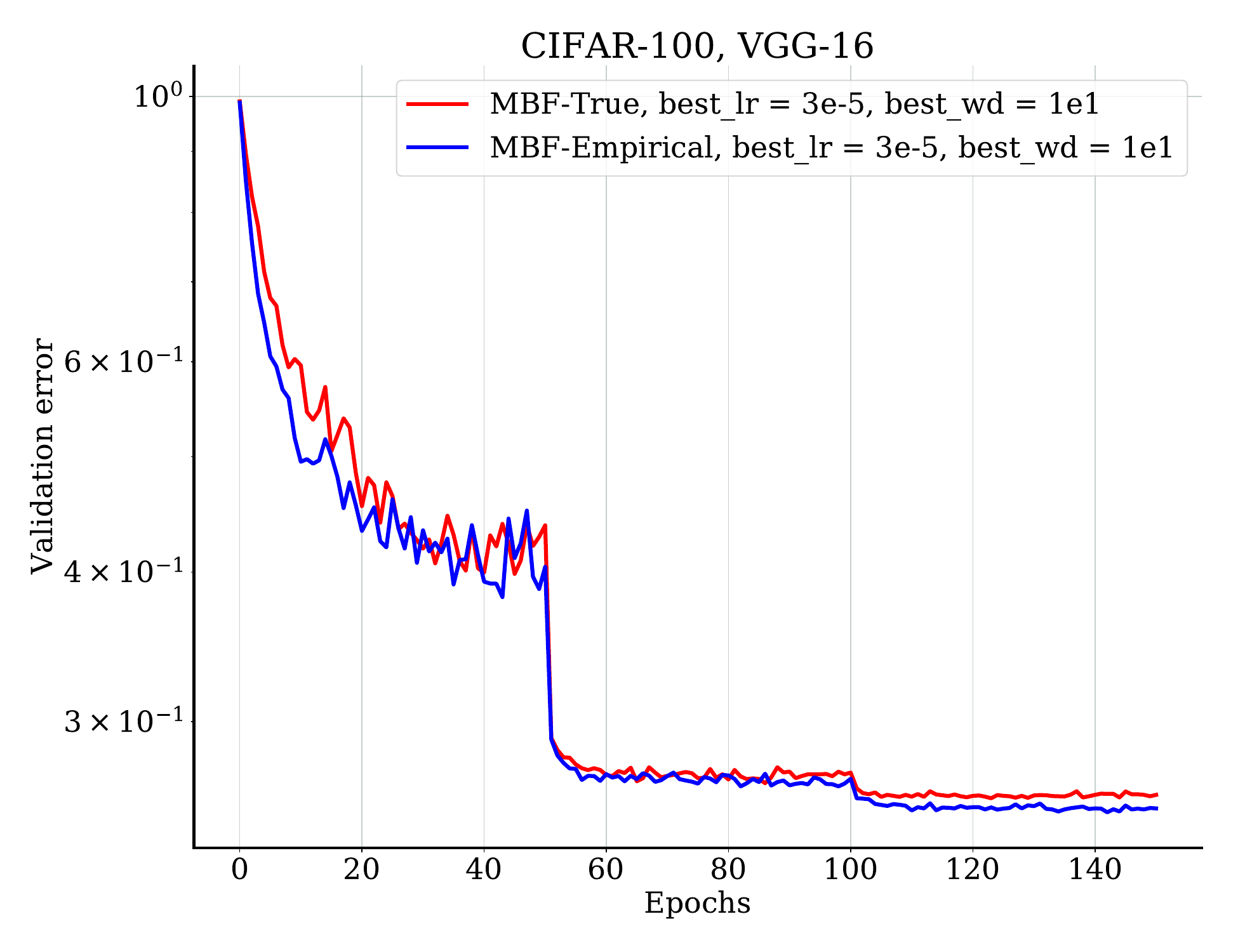}}\\
  \subfigure[\footnotesize SVHN, VGG-11]{\includegraphics[width=0.22\textwidth]{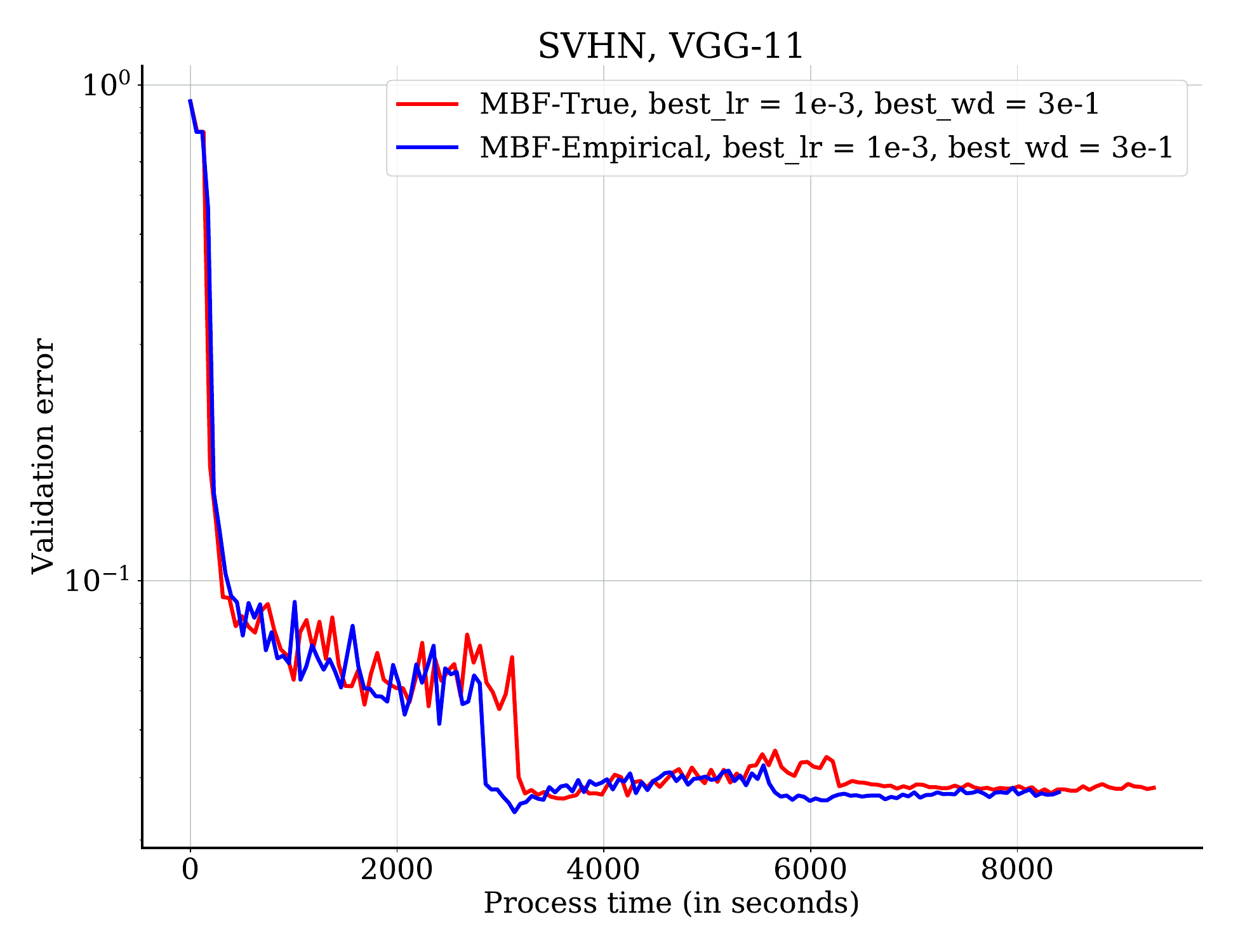} \quad \includegraphics[width=0.22\textwidth]{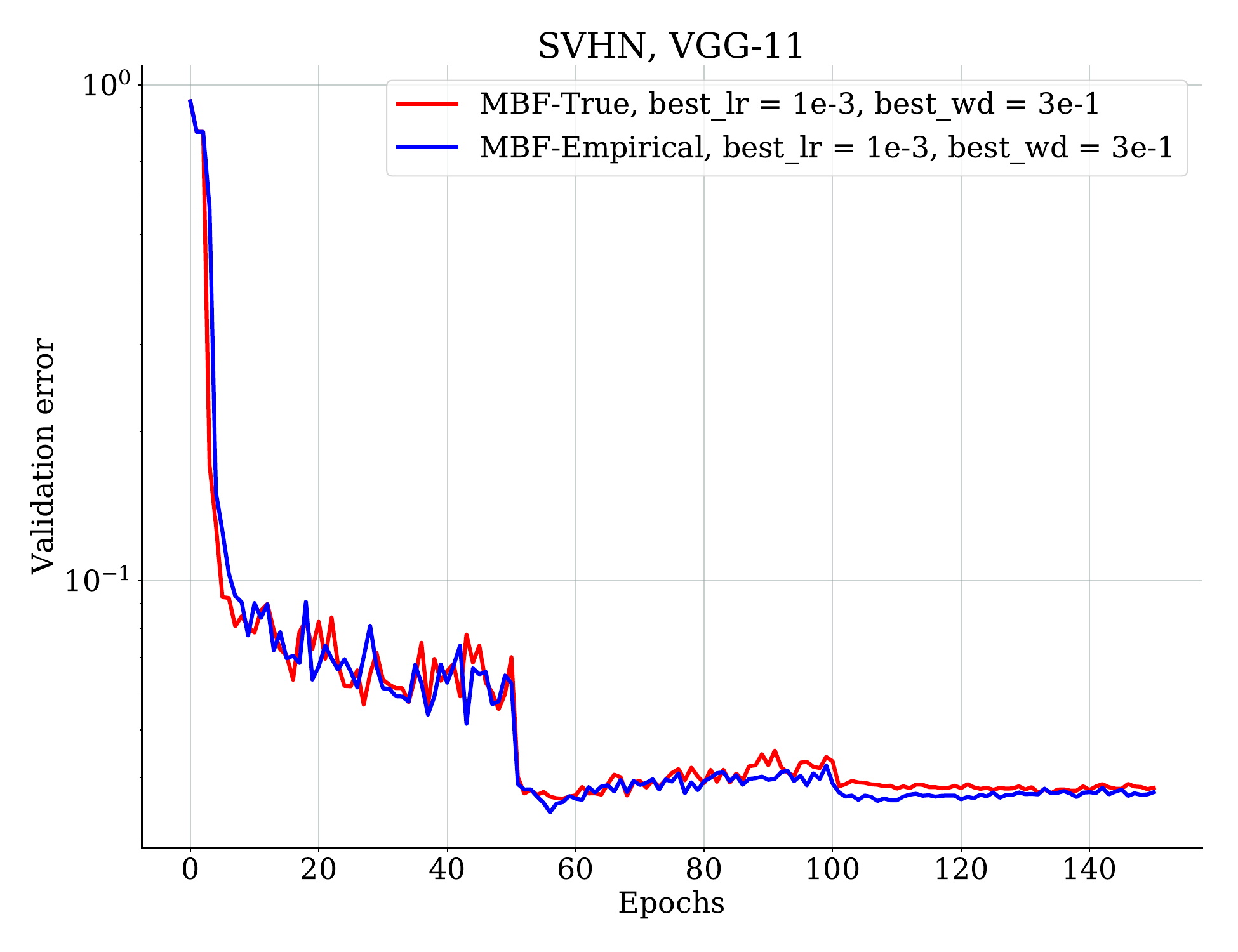}}
    \caption{
    Testing performance of MBF-True and MBF on three CNN problems.
    }
    \label{fig_cnn_true}
\end{figure}

The cosine similarity results reported in the main manuscript (see Figure 6 and related discussion) on the down-scaled MNIST suggest that the direction obtained by MBF and MBF-True behave similarly with respect the direction obtained by BDF. In this section, we compare the performance of MBF-True to MBF on the same Autoencoder problems (MNIST, FACES, CURVES) described in \ref{apx_exp_fcc} and the same CNN problems (CIFAR-10 + ResNet32, CIFAR-100 + VGG16, and SVHN + VGG11) described in \ref{apx_exp_ccn}. We 
used the same grid of parameters to tune MBF-True as the one described in \ref{apx_exp_fcc} and \ref{apx_exp_ccn}. We report in Figures \ref{fig_autoencoders_true} and \ref{fig_cnn_true} the training and validation errors obtained on these problems, as well as the best hyper-parameters for both methods in the legends. It seems that using the symmetric outer product of the empirical mini-batch gradient to update the mini-block preconditioner yields better results than using the mini-batch gradient from sampled data from the model's distribution to compute this inner product. 

We think this might be the case because MBF is closer to being an adaptive gradient methods, which also use the empirical gradient such as ADAGRAD and ADAM, rather than a second-order natural gradient method such as KFAC, where in the latter case using a sampled gradient yields better results than using the empirical data. Note that, when the mini-block sizes are 1, MBF becomes a diagonal preconditioning method like ADAM minus the square root operation.

\subsubsection{Spacial averaging on convolutional layers.}\label{apx_cnn_avg}

\begin{figure}[!h]
\centering
\subfigure[\footnotesize CIFAR-10, ResNet-32]{\includegraphics[width=0.22\textwidth]{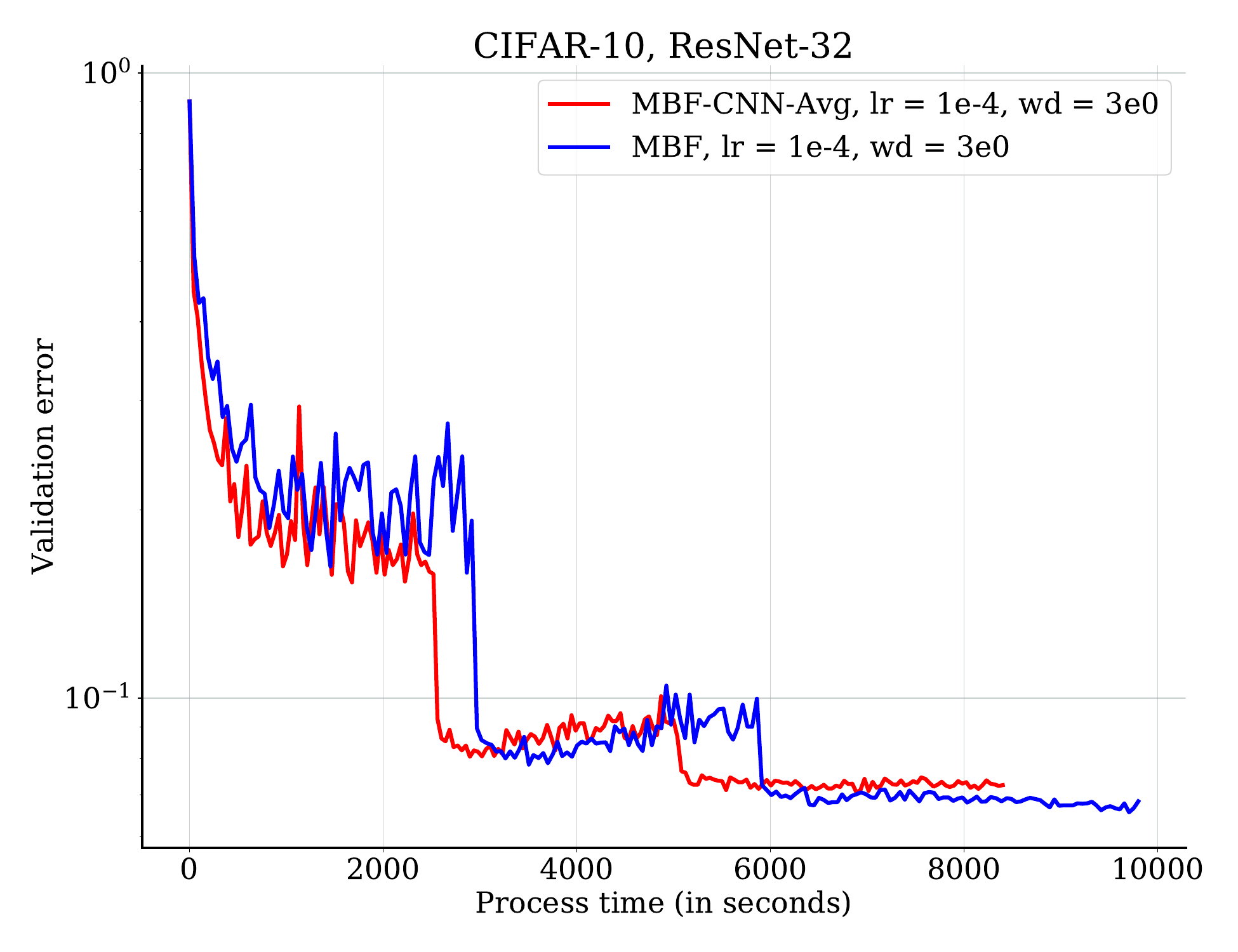} \quad \includegraphics[width=0.22\textwidth]{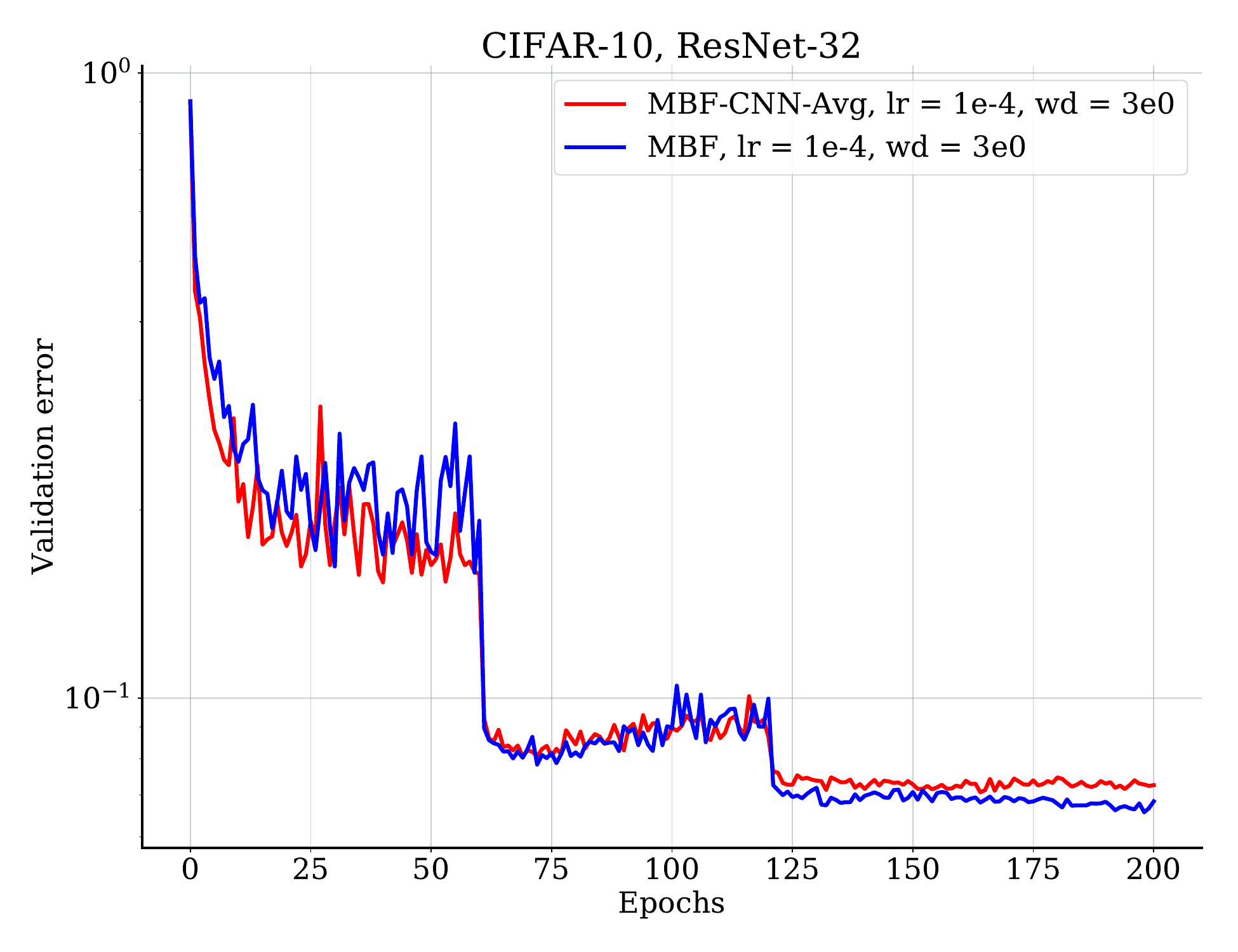}}\quad
  \subfigure[\footnotesize CIFAR-100, VGG-16]{\includegraphics[width=0.22\textwidth]{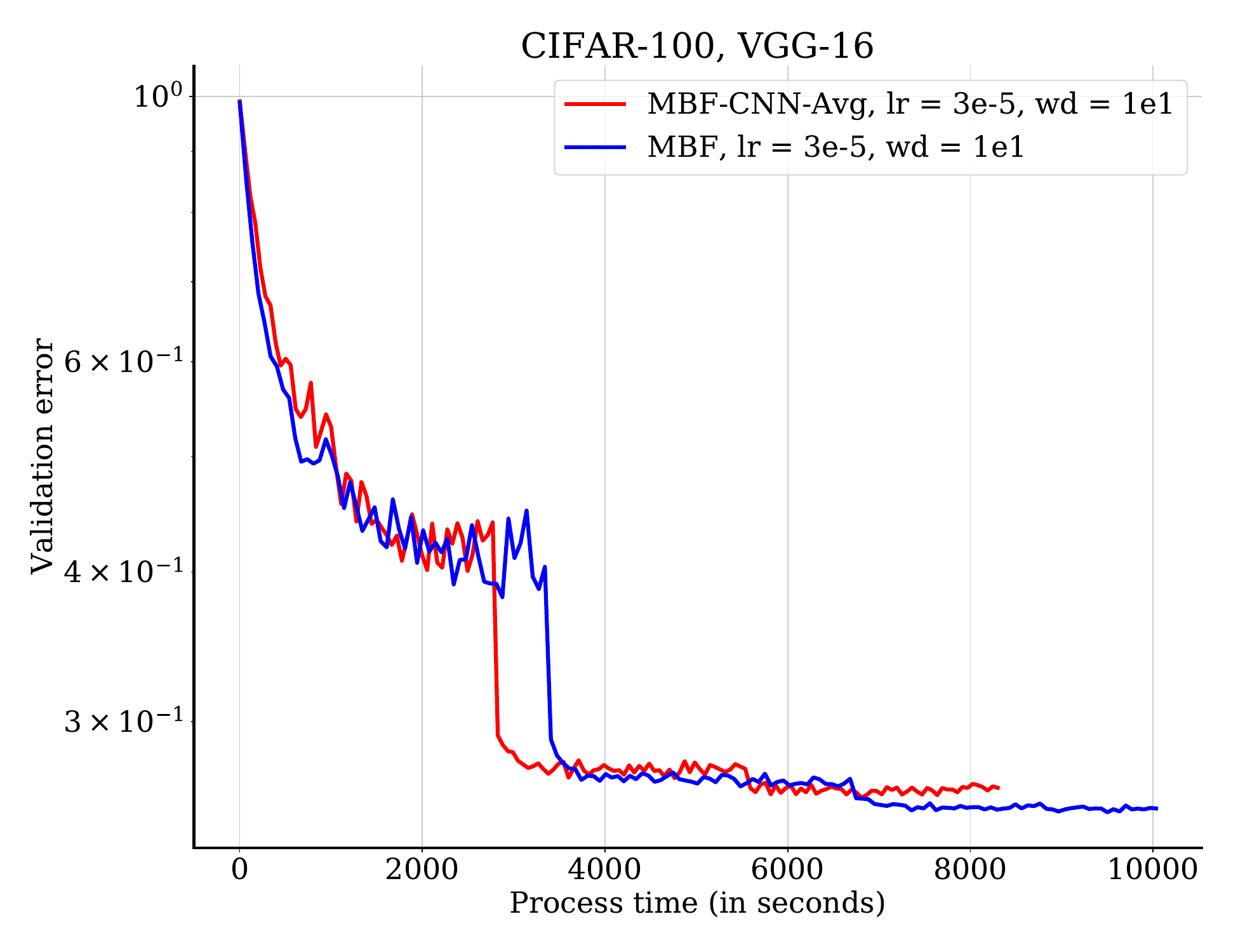} \quad \includegraphics[width=0.22\textwidth]{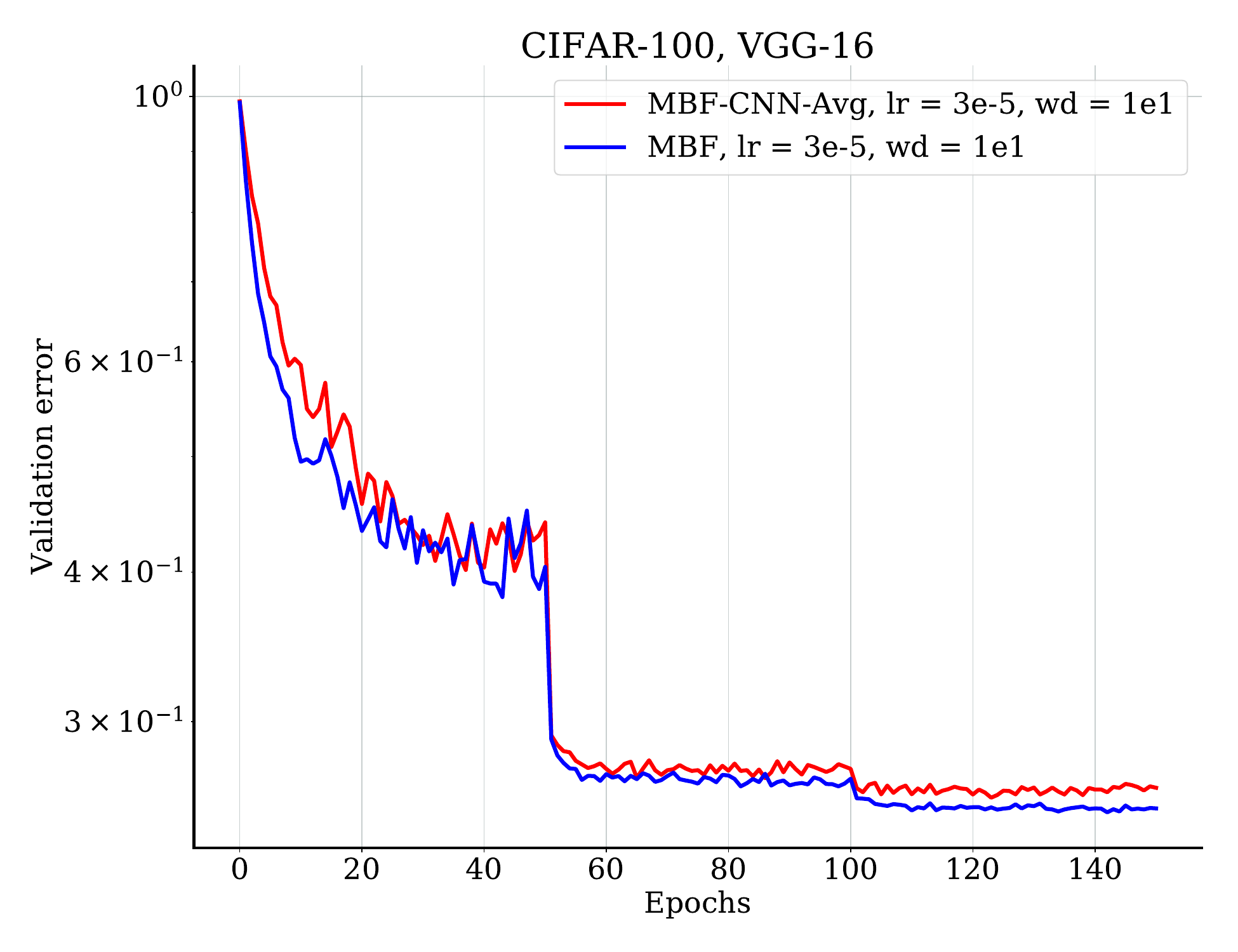}}\\
  \subfigure[\footnotesize SVHN, VGG-11]{\includegraphics[width=0.22\textwidth]{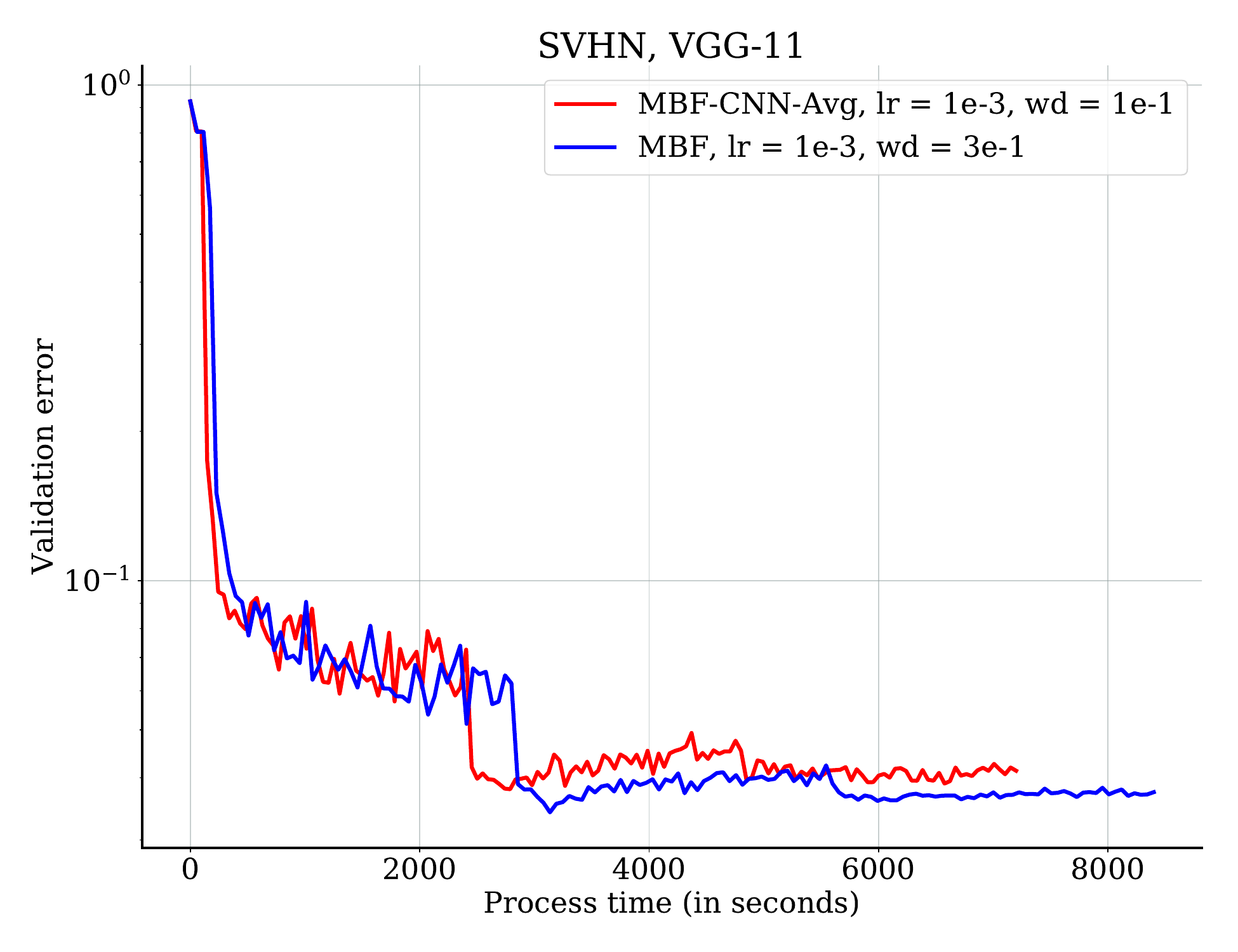} \quad \includegraphics[width=0.22\textwidth]{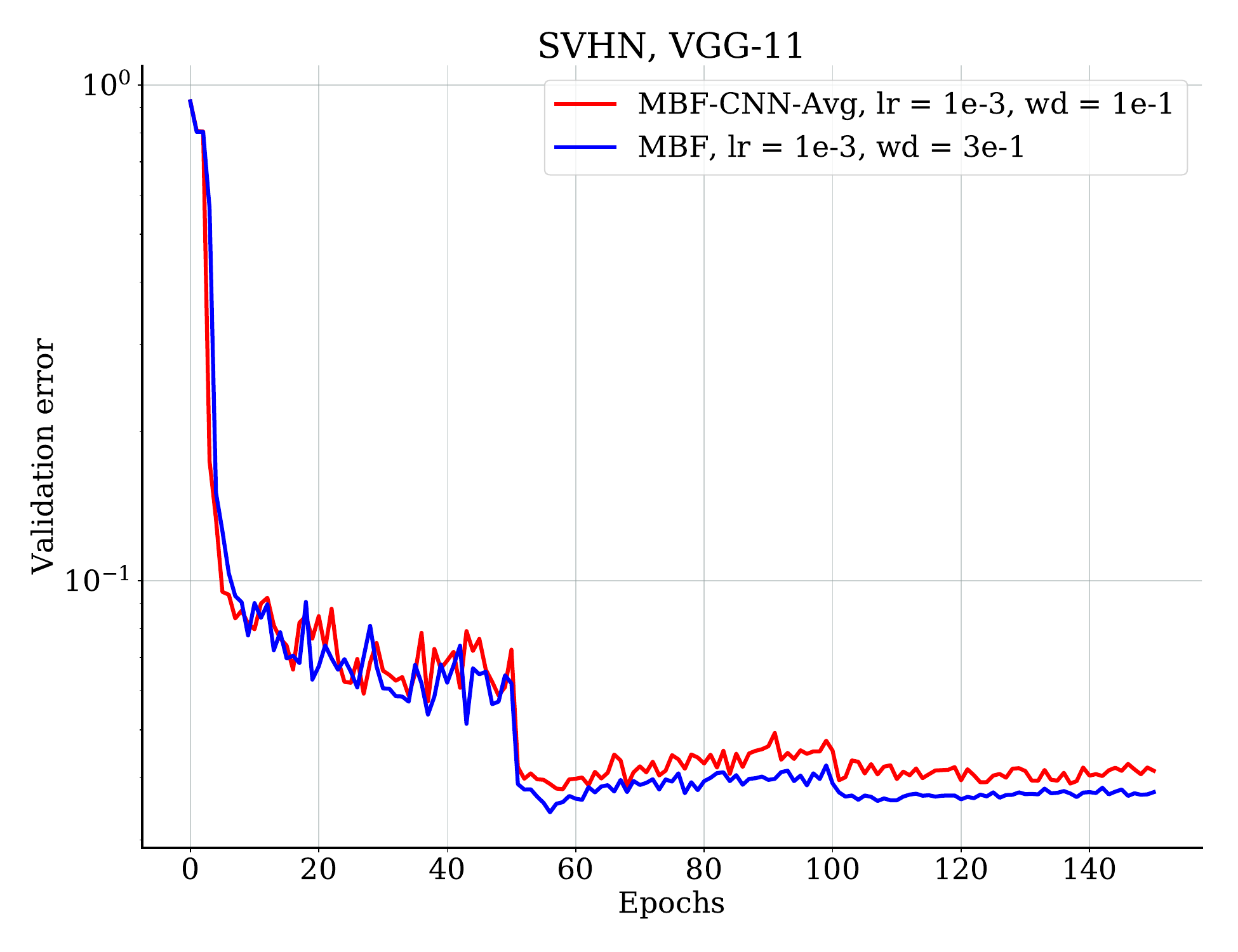}}
    \caption{
    Testing performance of MBF-CNN-Avg(MBF with spacial averaging applied to CNN layers) and MBF on three CNN problems.
    }
    \label{fig_cnn_avg}
\end{figure}

In this section, we compare the performance of MBF with spacial averaging applied to convolutional layers to MBF on the same three CNN problems (CIFAR-10 + ResNet-32, CIFAR-100 + VGG16, and SVHN + VGG11) described in \ref{apx_exp_ccn}. We used the same grid of parameters to tune MBF-CNN-Avg as the one described in \ref{apx_exp_ccn}. We report in Figure \ref{fig_cnn_avg} the validation errors obtained on these problems, as well as the best hyper-parameters for both methods in the legends. It seems that using the average of the kernel-wise mini-blocks to update the preconditioner yields slightly worse results than using the individual mini-blocks as preconditioner. We think this might be the case because the averaging over all mini-blocks results into a loss of curvarture information as the kernel-wise mini-blocks are small in size. Note that, when using the average mini-blocks, MBF will require less memory than adaptive first order methods such as ADAM.

\subsubsection{On the effect of the update frequencies $T_1, T_2$:} 

\begin{figure}[!h]
\centering
\subfigure[\footnotesize a) MNIST autoencoder, $T_2 =10$]{\includegraphics[width=0.8\textwidth]{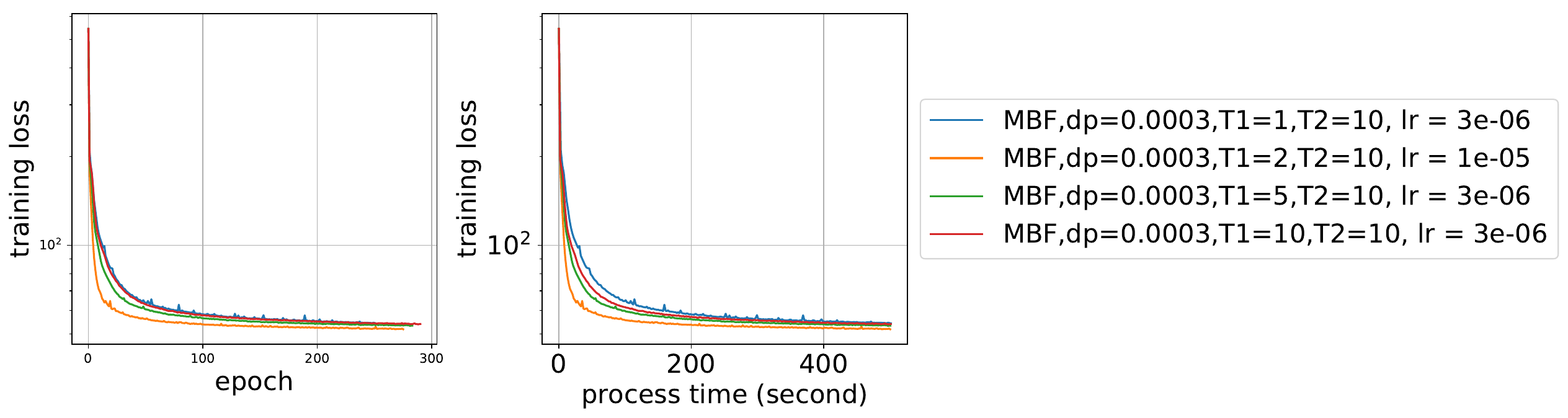}}\\
  \subfigure[\footnotesize b) MNIST autoencoder, $T_2 =20$]{\includegraphics[width=0.8\textwidth]{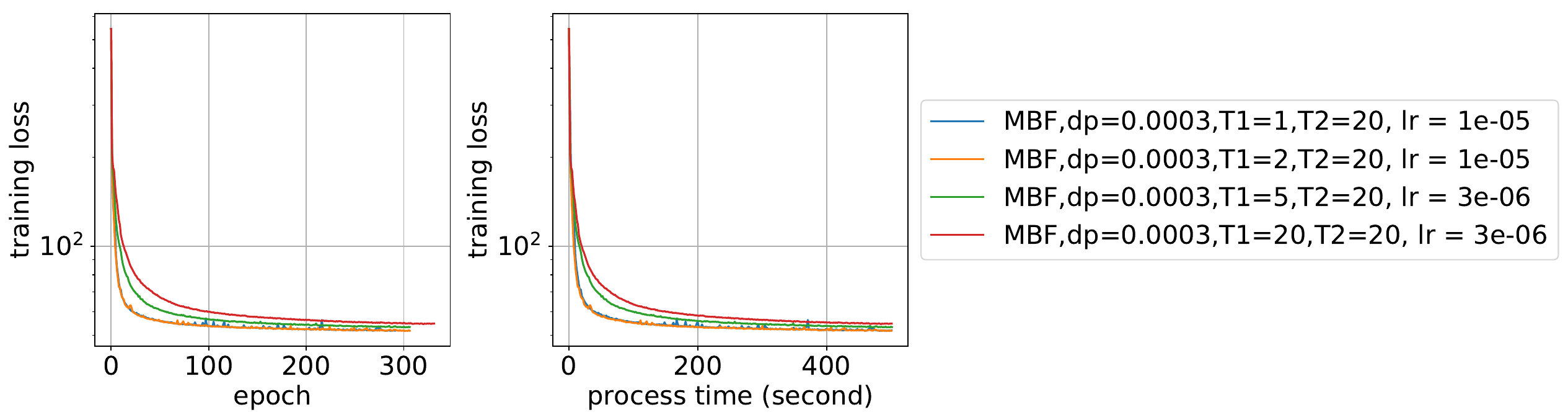}}\\
  \subfigure[\footnotesize c) MNIST autoencoder, $T_2 =25$]{\includegraphics[width=0.8\textwidth]{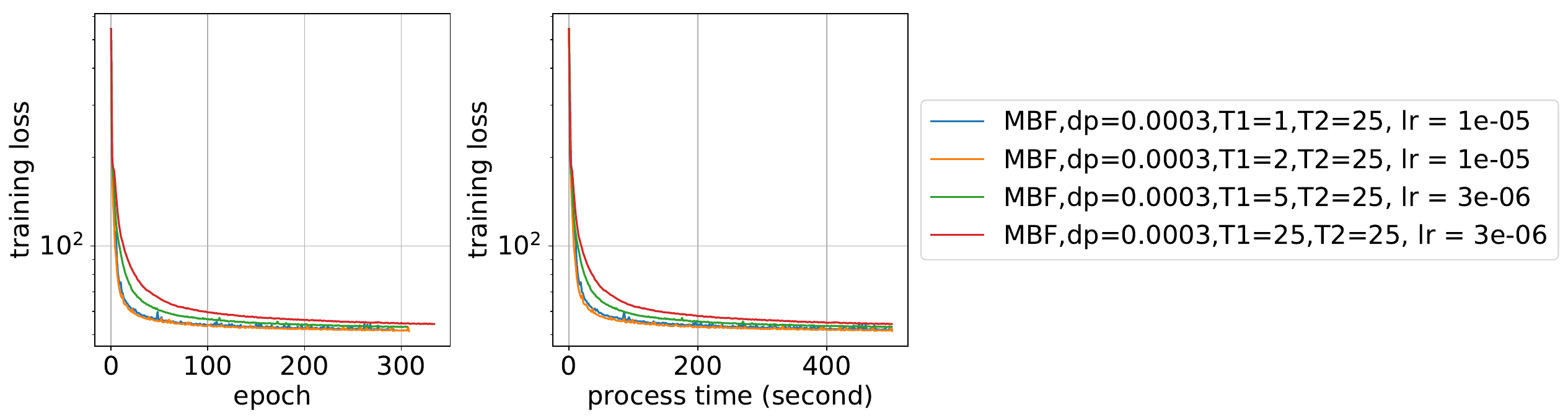}}
    \caption{
    Training performance of MBF on MNIST autoencoder problems for some combinations of $T_1, T_2$.
    }
    \label{fig_t1t2}
\end{figure}

We also explored the effect of the update frequencies $T_1, T_2$ for the mini-block preconditionners as used in Algorithm \ref{algo_MBF_full}. To be more specific, we tuned the learning rate for various combinations of $T_1, T_2$ depicted in Figure \ref{fig_t1t2}.
Comparing the performance of
Algorithm \ref{algo_MBF_full} for these different configurations, we can see that the effect of the frequencies $T_1, T_2$ on the final performance of MBF is minimal and the configurations $T_1, T_2 = (1,20)$, $T_1, T_2 = (2,25)$ seem to yield the best performance in terms of process time for autoencoder problems. 

\subsubsection{Additional inverse EMF heatmap illustrations}
\label{apx_more_motivation}
As mentioned in the main manuscript, we include here additional examples that illustrate that most of the weight in the inverse of the empirical Fisher matrix resides in the mini-blocks used in MBF. For convolutional layers, we trained a simple convolutional neural network, Simple CNN on Fashion MNIST \citep{xiao2017fashion}. The model is identical to the base model described in \cite{shallue2019measuring}. It consists of 2 convolutional layers with max pooling with 32 and 64 filters each and $5\times5$ filters with stride 1, “same” padding, and ReLU activation function followed by 1 fully connected layer. Max pooling uses a $2\times2$ window with stride 2. The fully connected layer has 1024 units. It does not use batch normalization. 

Figure \ref{heatmap_cnn_appendix} shows the heatmap of the absolute value of the inverse empirical Fisher corresponding to the second convolutional layer for channels 1, 16 and 32, which all use $64$ filters of size $5 \times 5$ (thus 64 mini-blocks of size $25 \times 25$ per channel). One can see that the mini-block (by filter) diagonal approximation is reasonable.

\begin{figure}[H]
  \centering
  \includegraphics[width=0.95\textwidth]{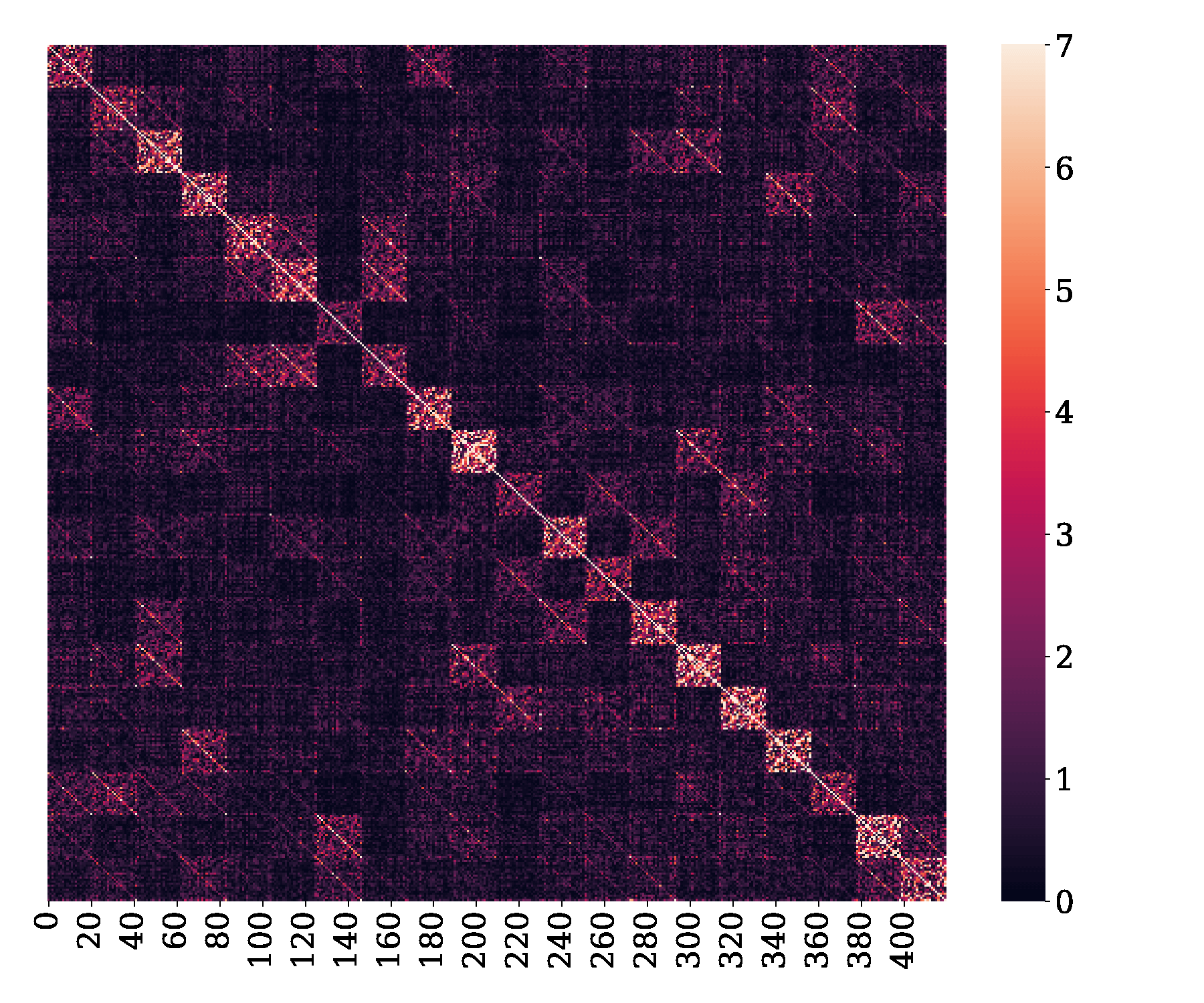}
  \caption{Absolute inverse EFM, second fully connected layer 20-20
}
\label{heatmap_fcc_apx}
\end{figure}

\begin{figure}[H]
  \centering
  \subfigure[\footnotesize Absolute inverse EFM for channel 1]{\includegraphics[width=0.4\textwidth]{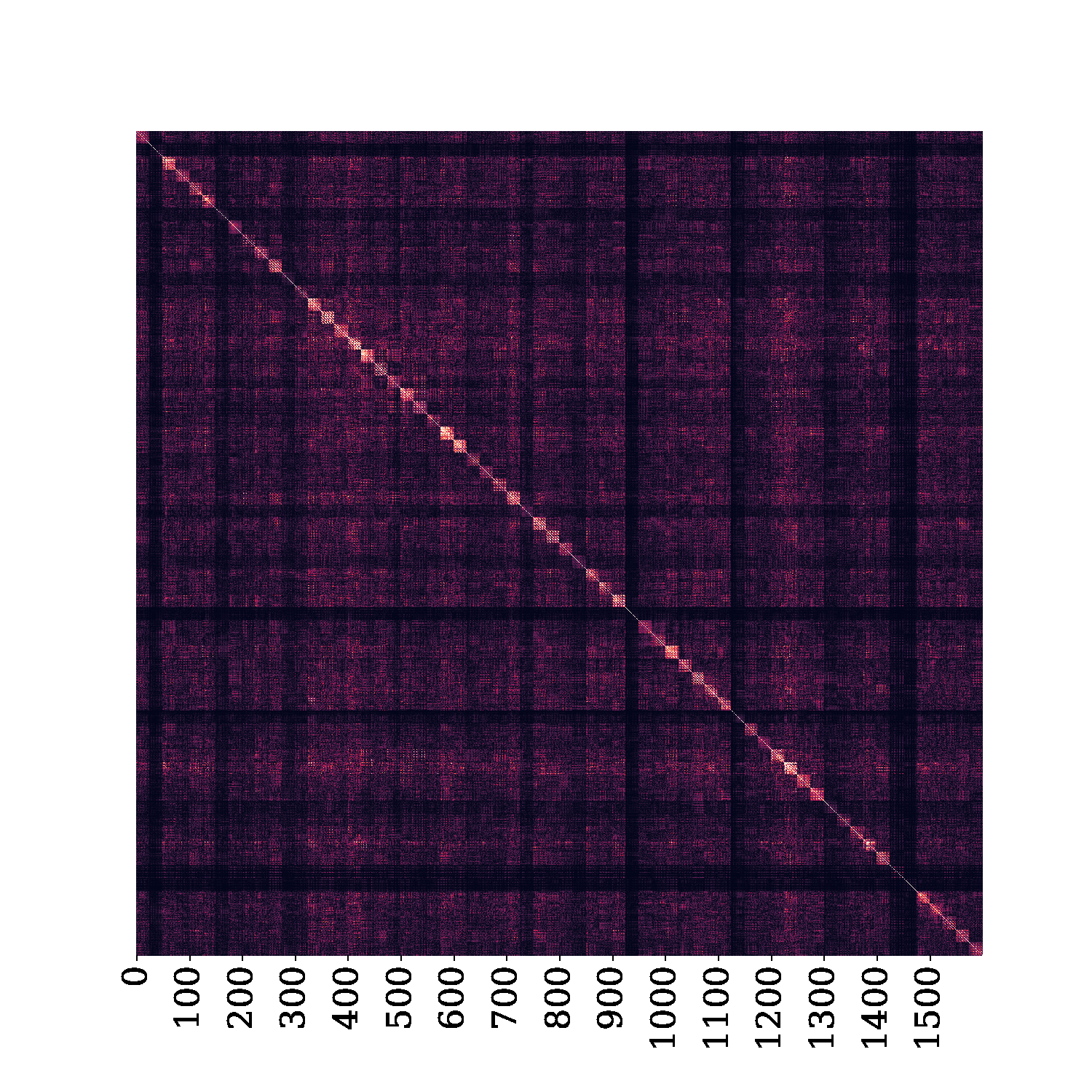}}\quad
  \subfigure[\footnotesize Zoom on the 20th to 30th blocks]{\includegraphics[width=0.4\textwidth]{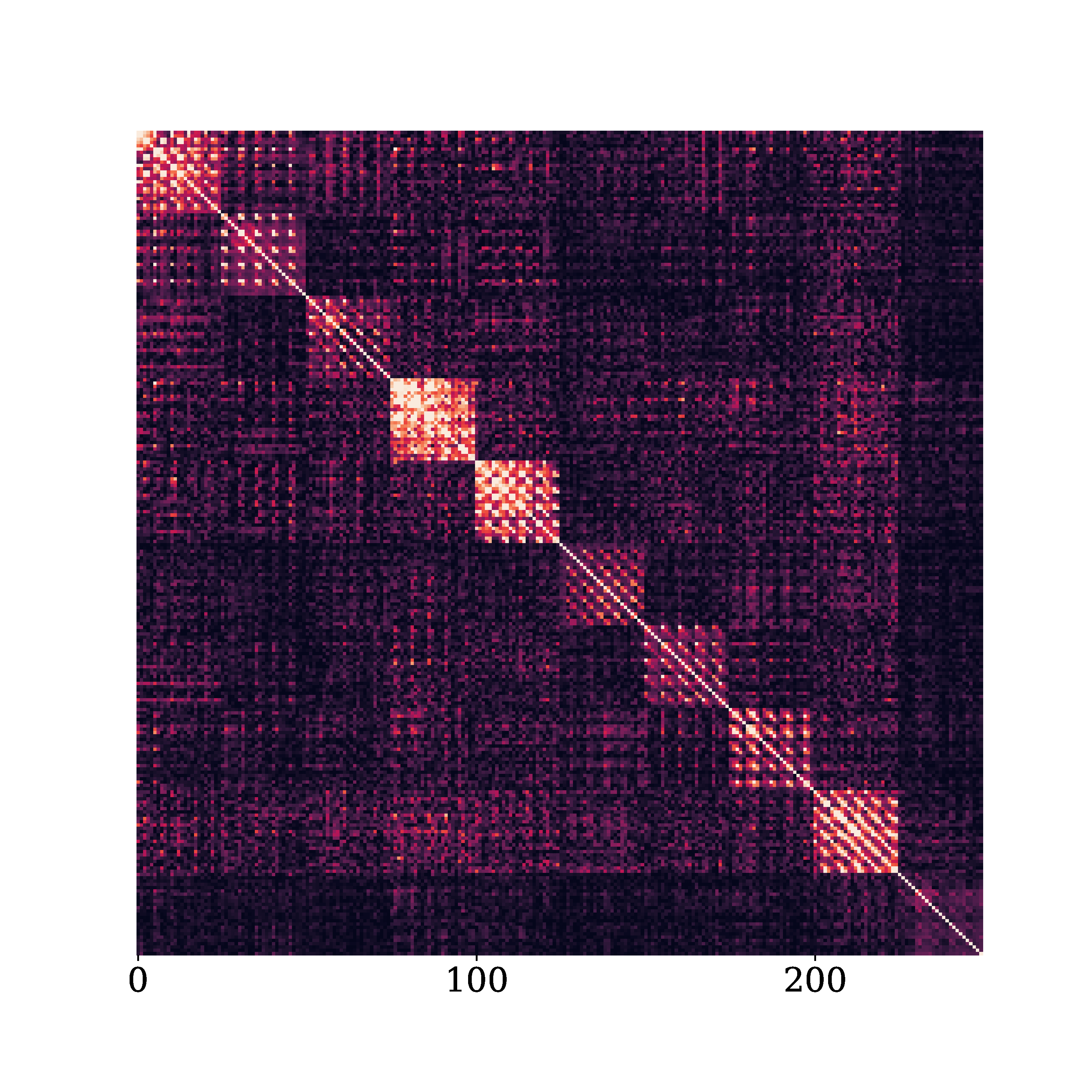}}\\
  \subfigure[\footnotesize Absolute inverse EFM for channel 16]{\includegraphics[width=0.4\textwidth]{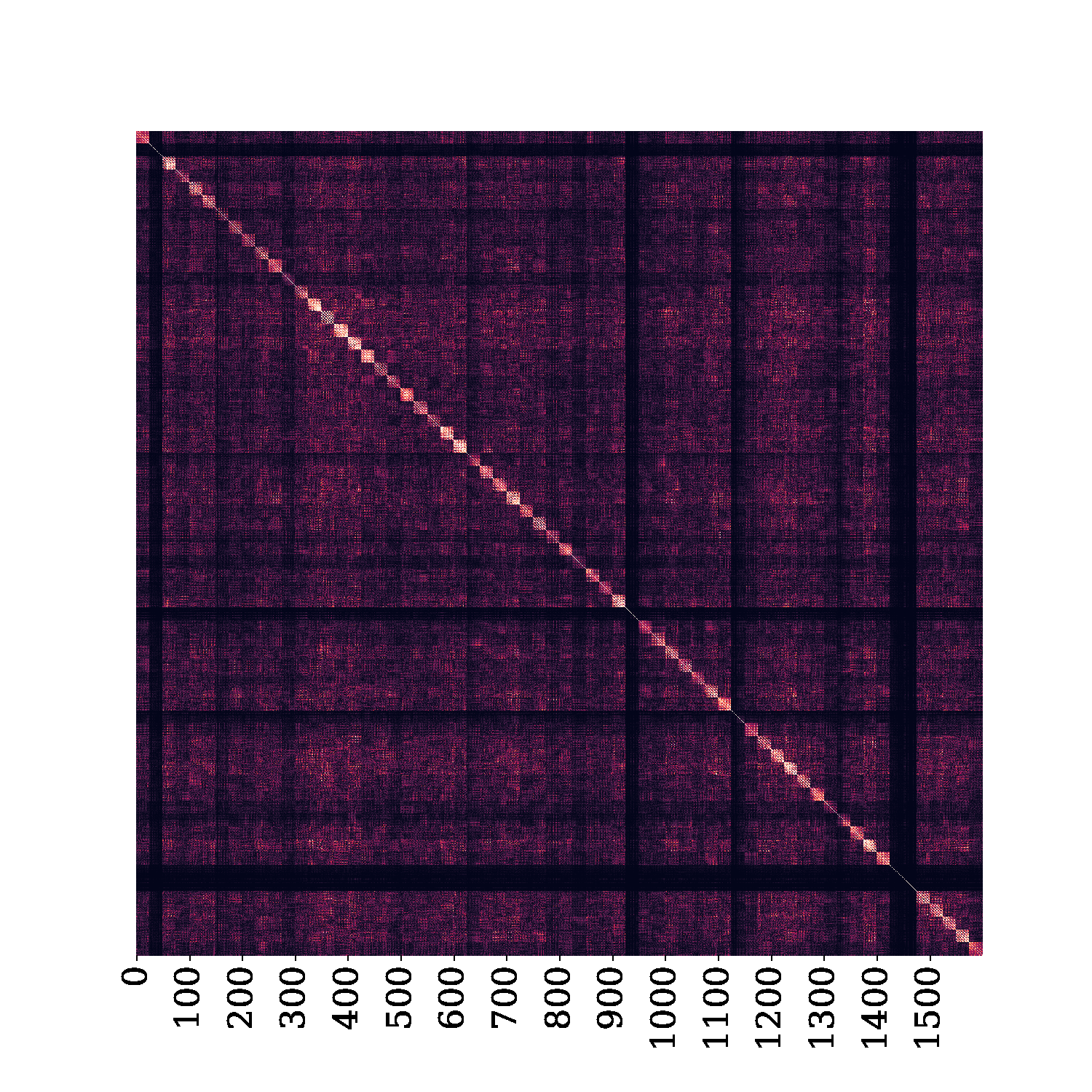}}\quad
  \subfigure[\footnotesize Zoom on the 20th to 30th blocks]{\includegraphics[width=0.4\textwidth]{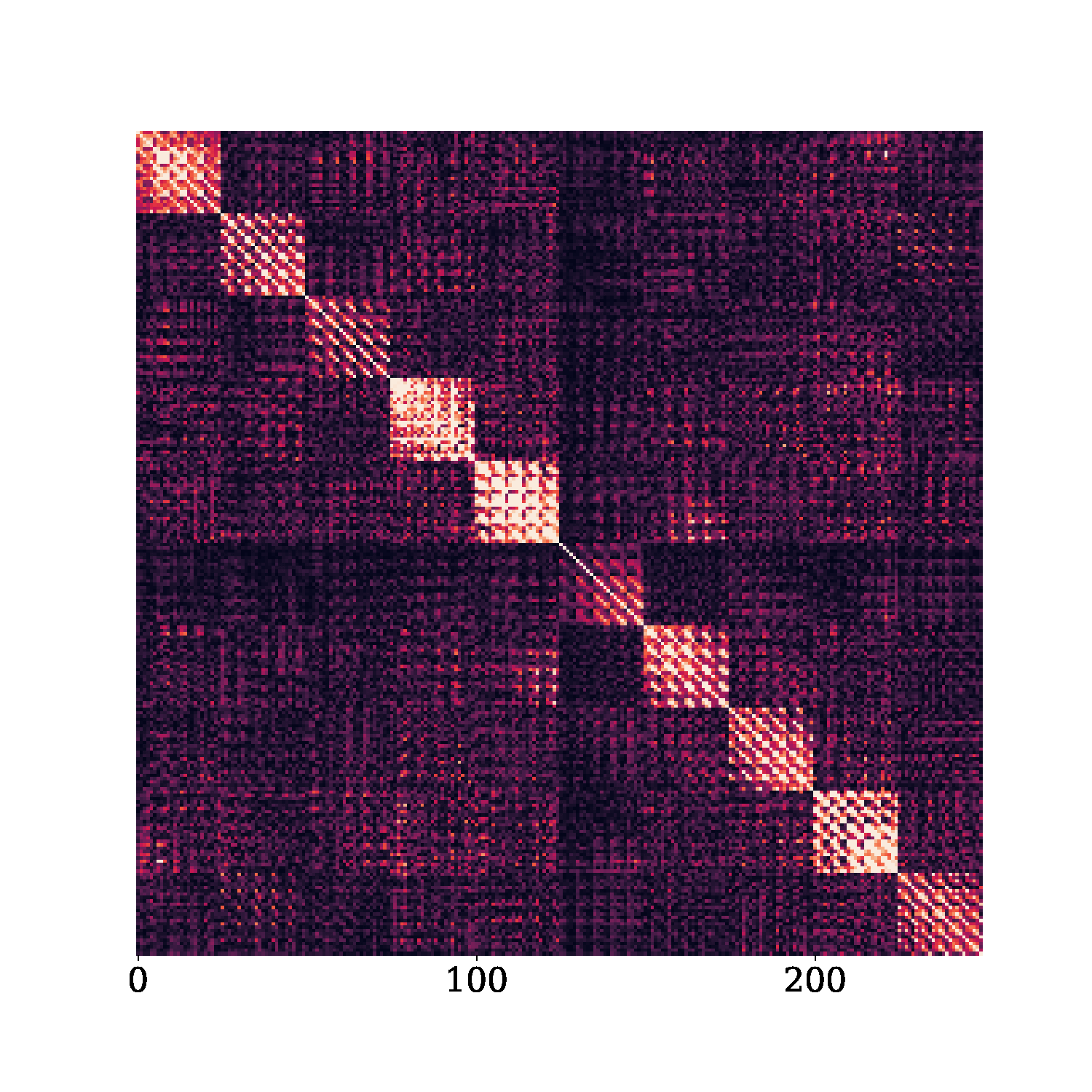}}\\
  \subfigure[\footnotesize Absolute inverse EFM for channel 32]{\includegraphics[width=0.4\textwidth]{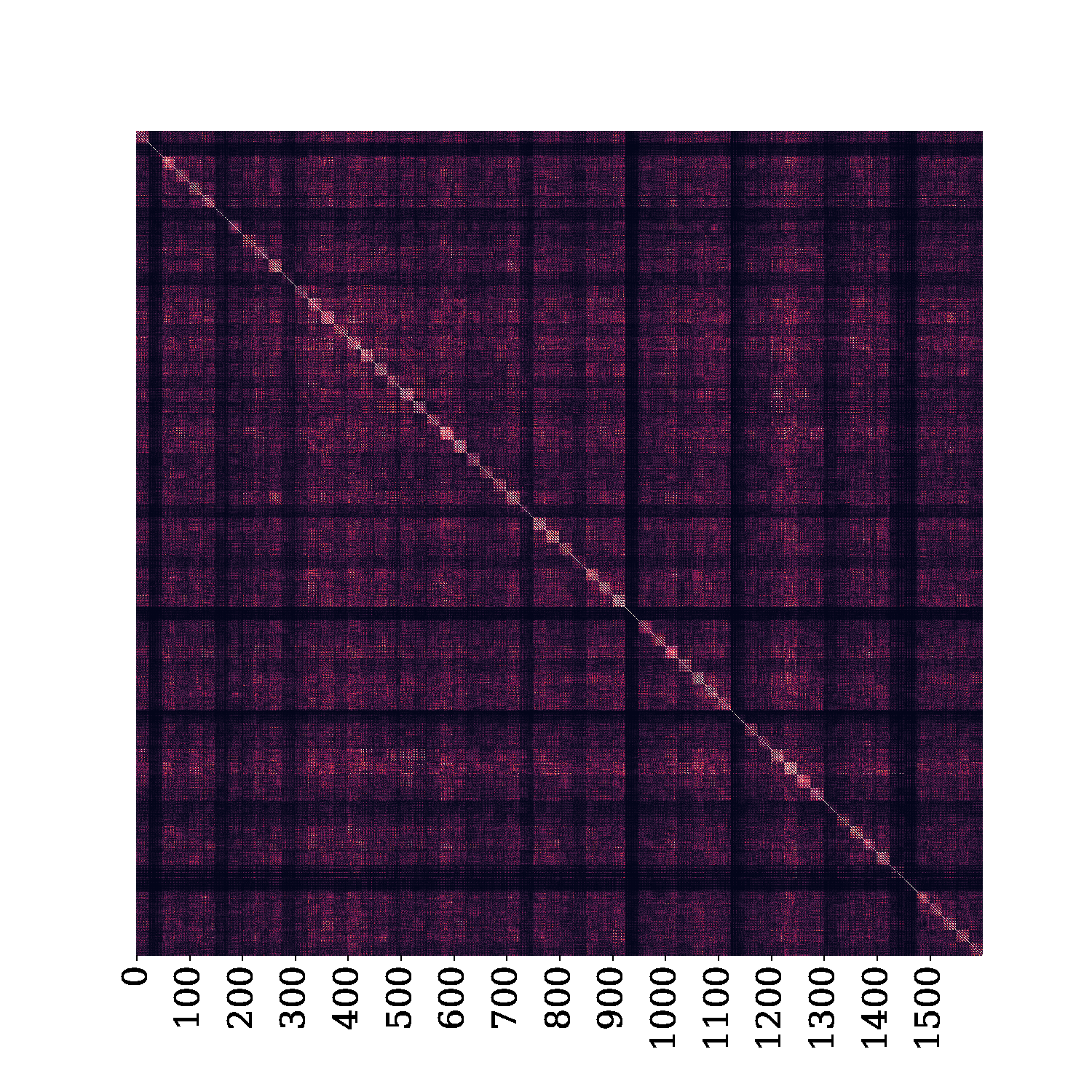}}\quad
  \subfigure[\footnotesize Zoom on the 20th to 30th blocks]{\includegraphics[width=0.4\textwidth]{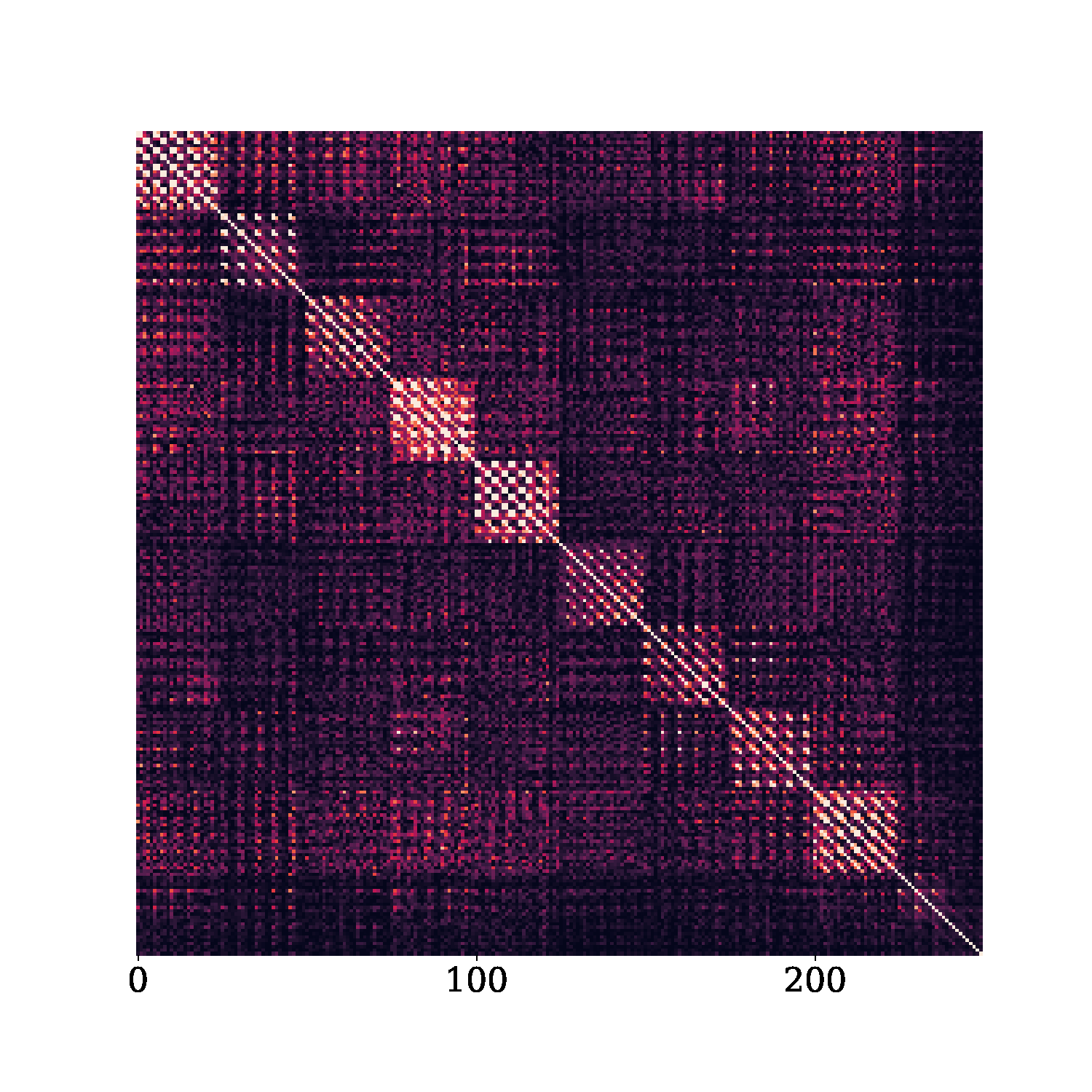}}\\
  \caption{
Absolute inverse of the empirical EFM after 10 epochs for the second convolutional layer of the Simple-CNN.
}
\label{heatmap_cnn_appendix}
\end{figure}

As mentioned in the manuscript, we illustrate the mini-block structure of the empirical Fisher matrix on a $7$-layer (256-20-20-20-20-20-10) feed-forward DNN using $\tanh$ activations, partially trained (after 50 epochs using SGD-m) to classify a $16 \times 16$ down-scaled version of MNIST that was also used in \citep{martens2015optimizing}. Figure \ref{heatmap_fcc_apx} shows the heatmap of the absolute value of the inverse empirical FIM for the second fully connected layers (including bias). One can see that the mini-block (by neuron) diagonal approximation is reasonable.

\subsection{Sensitivity to Hyper-parameters:}  

\subsection{MBF}

\begin{figure}[H]
\centering
\begin{minipage}{.33\textwidth}
  \centering
  \includegraphics[width=\textwidth]{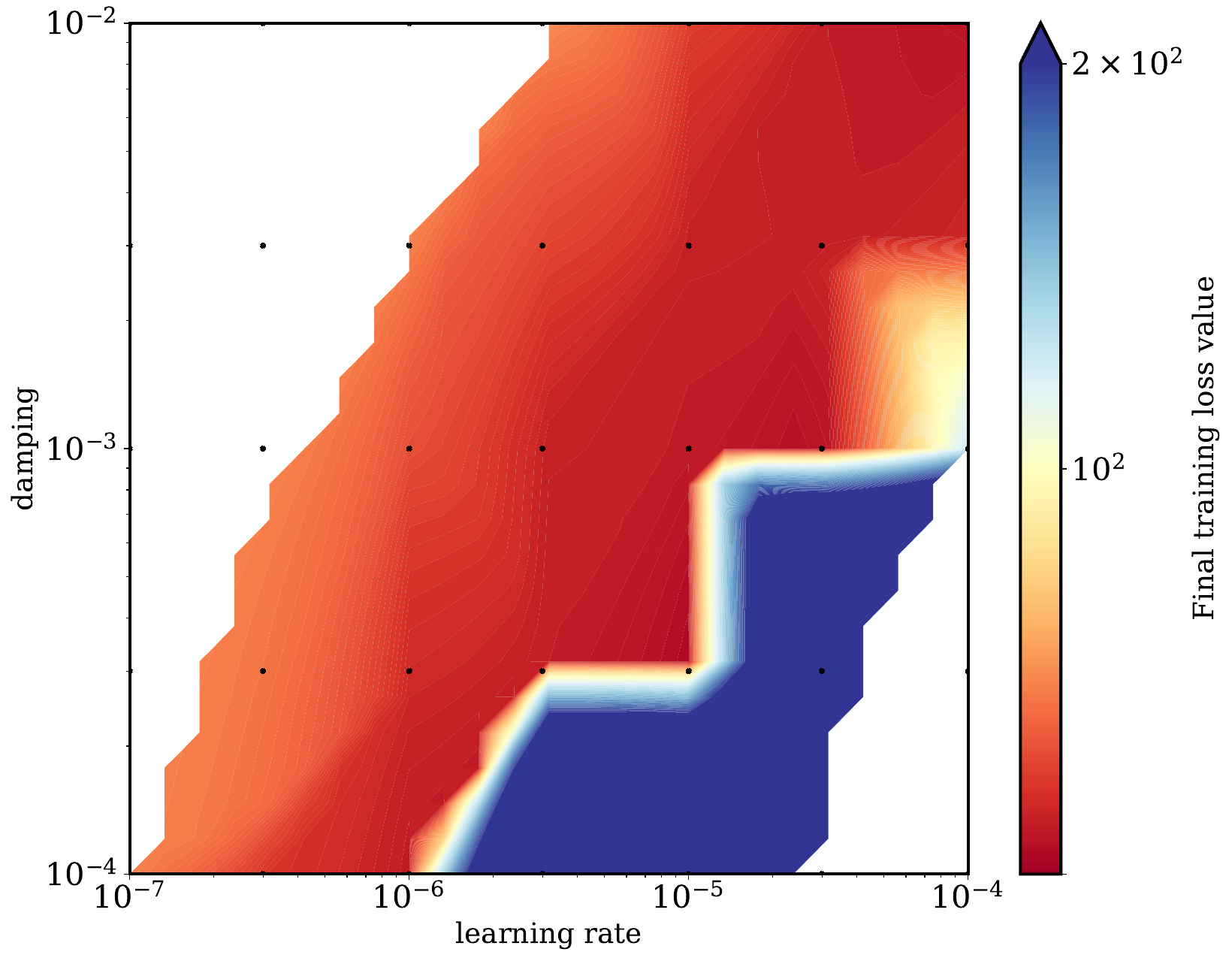}
\end{minipage}%
\begin{minipage}{.33\textwidth}
  \centering
  \includegraphics[width=\textwidth]{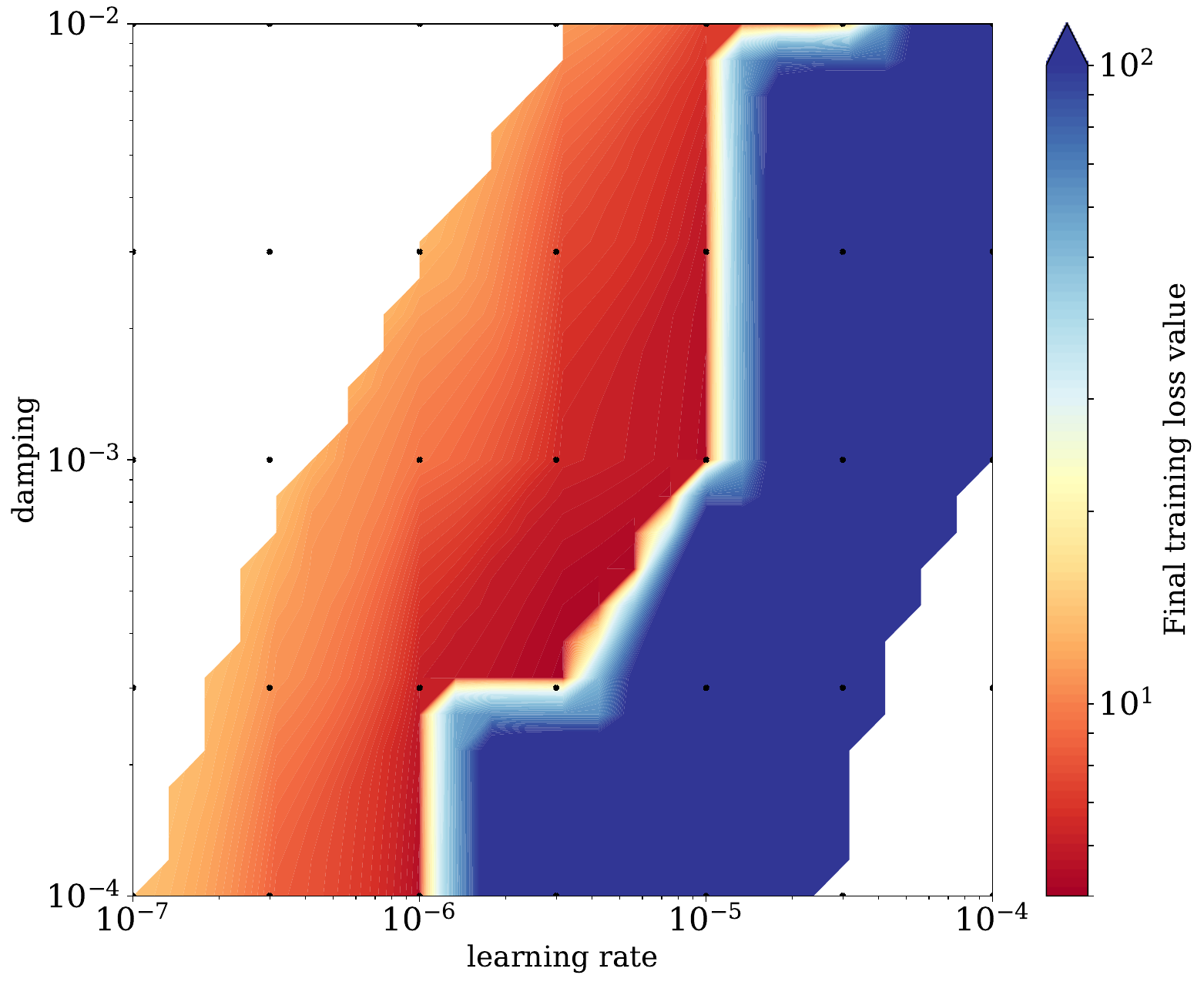}
\end{minipage}%
\begin{minipage}{.33\textwidth}
  \centering
  \includegraphics[width=\textwidth]{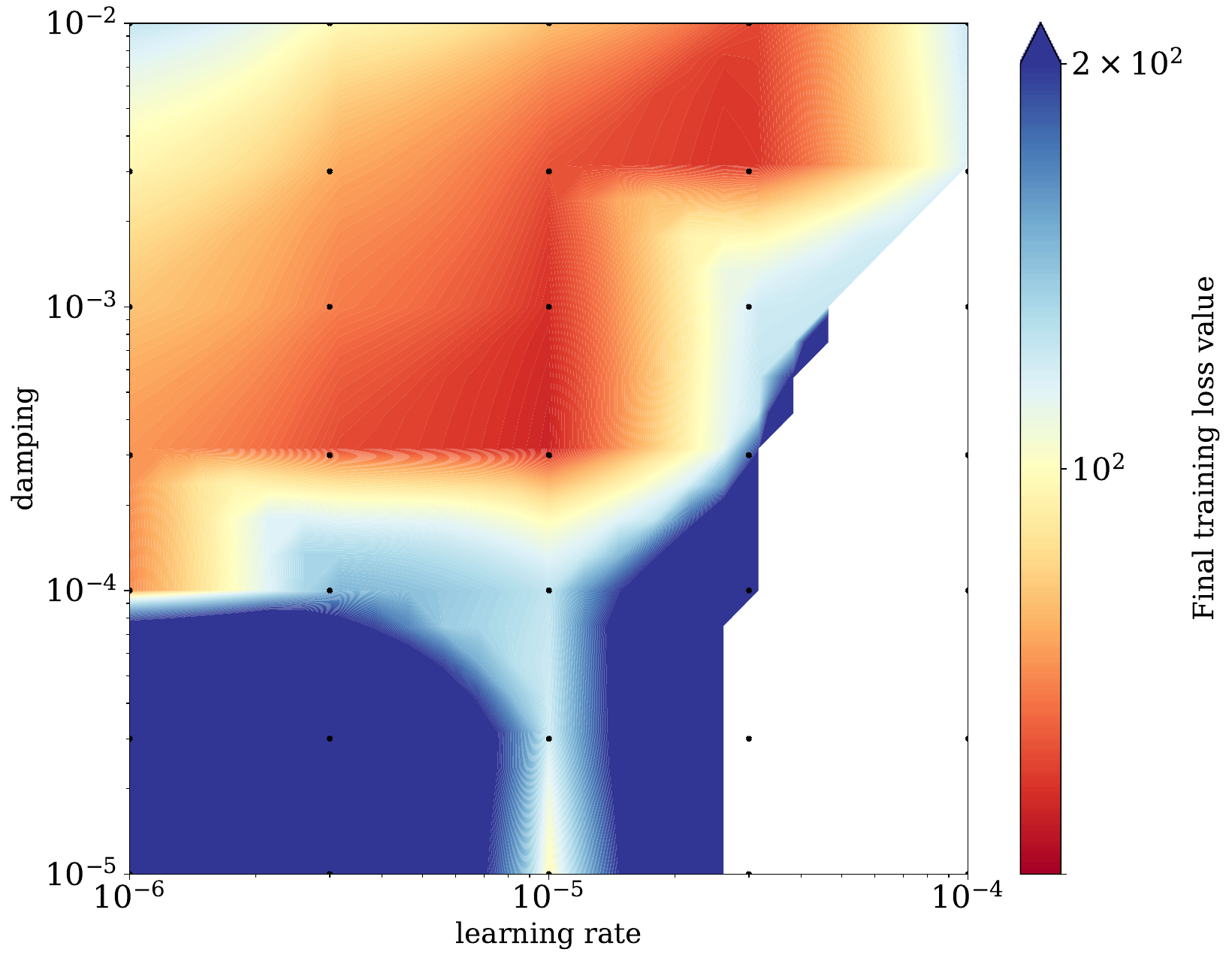}
\end{minipage}
\caption{
Landscape of the final training loss value w.r.t hyper-parameters (i.e. learning rate and damping) for MBF. The left, middle, right columns depict results for MNIST, FACES, CURVES, which are terminated after 500, 2000, 500 seconds (CPU time), respectively.
}
\label{fig_hyper}
\end{figure}

\subsection{KFAC}

\begin{figure}[H]
\centering
\begin{minipage}{.33\textwidth}
  \centering
  \includegraphics[width=\textwidth]{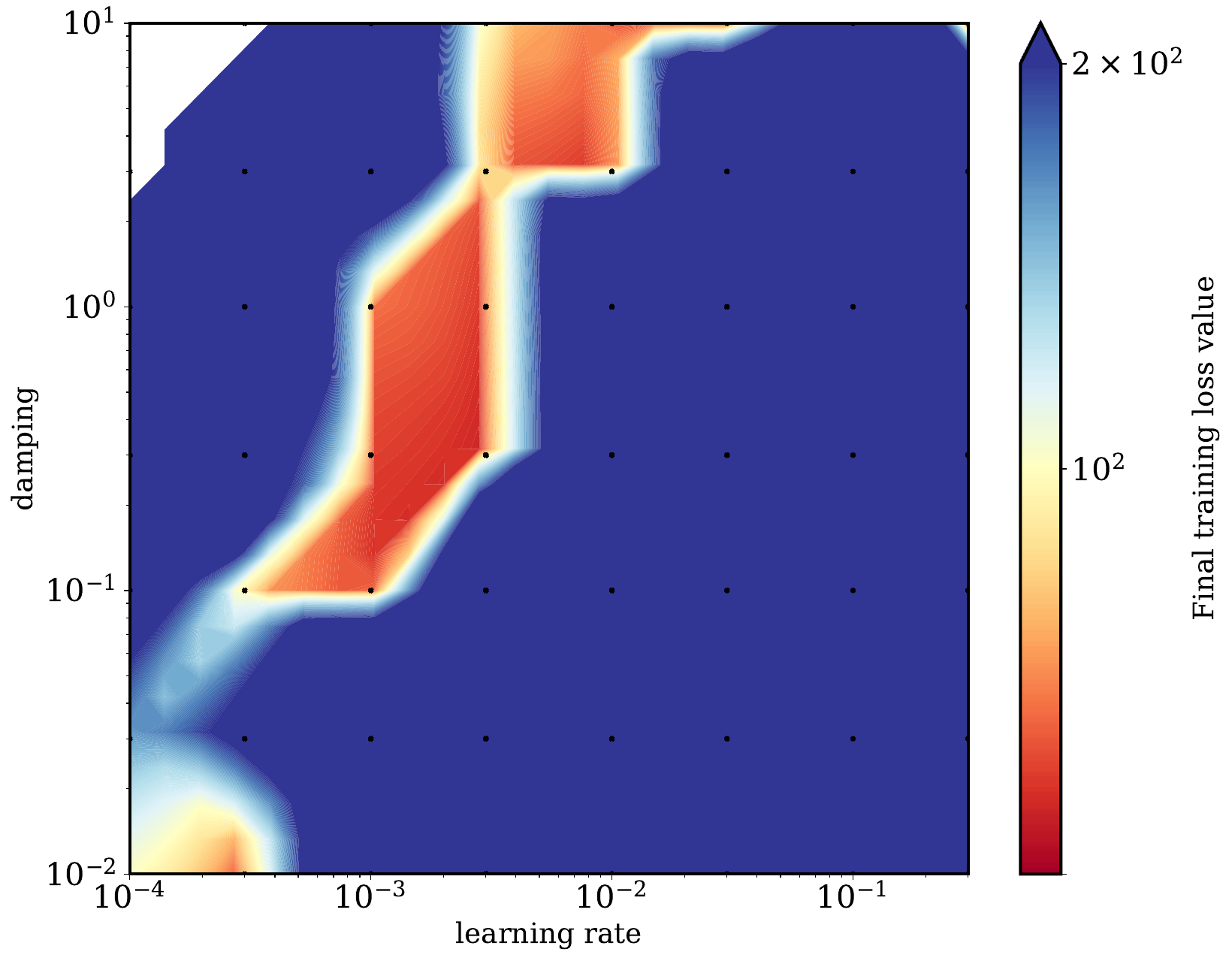}
\end{minipage}%
\begin{minipage}{.33\textwidth}
  \centering
  \includegraphics[width=\textwidth]{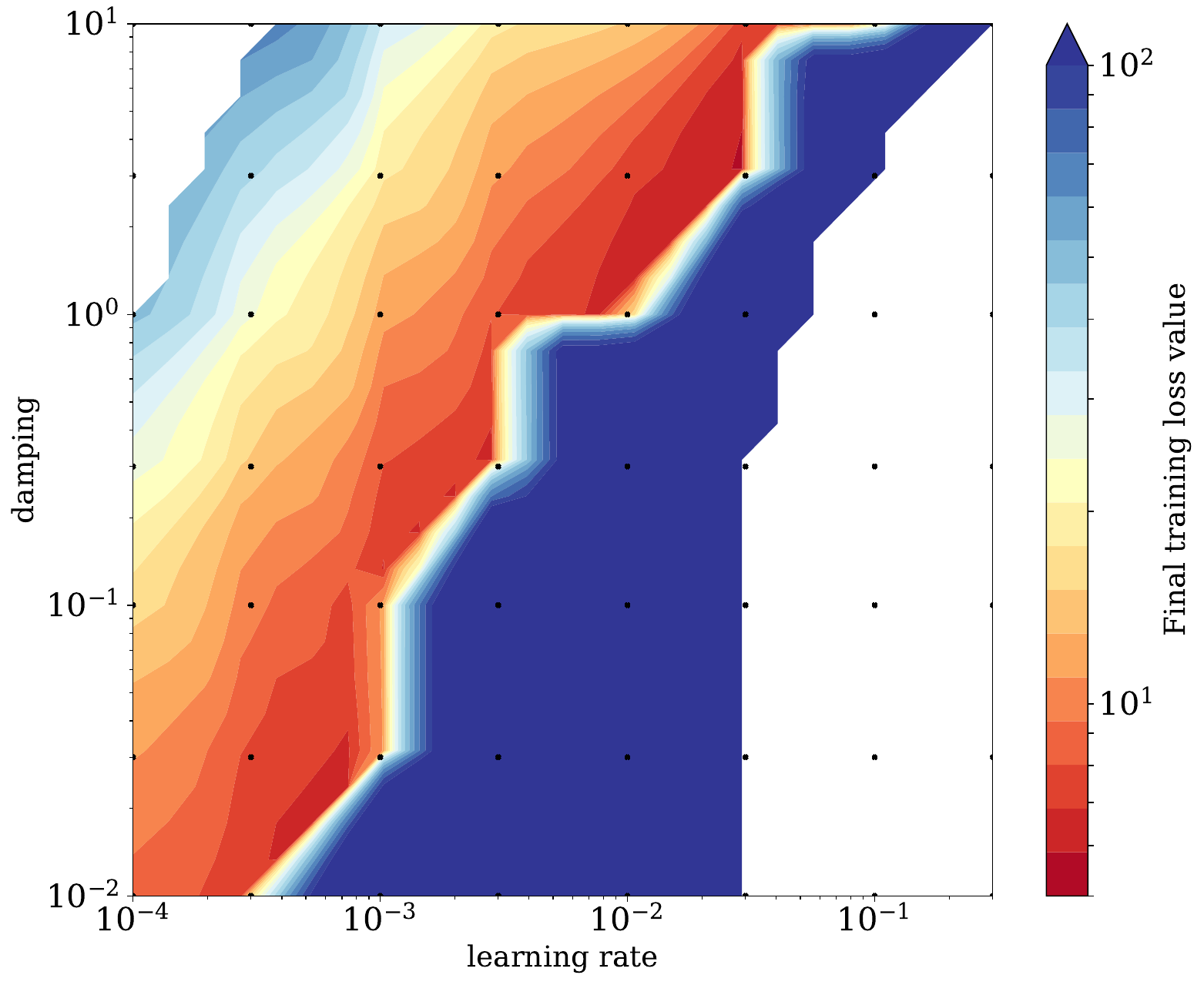}
\end{minipage}%
\begin{minipage}{.33\textwidth}
  \centering
  \includegraphics[width=\textwidth]{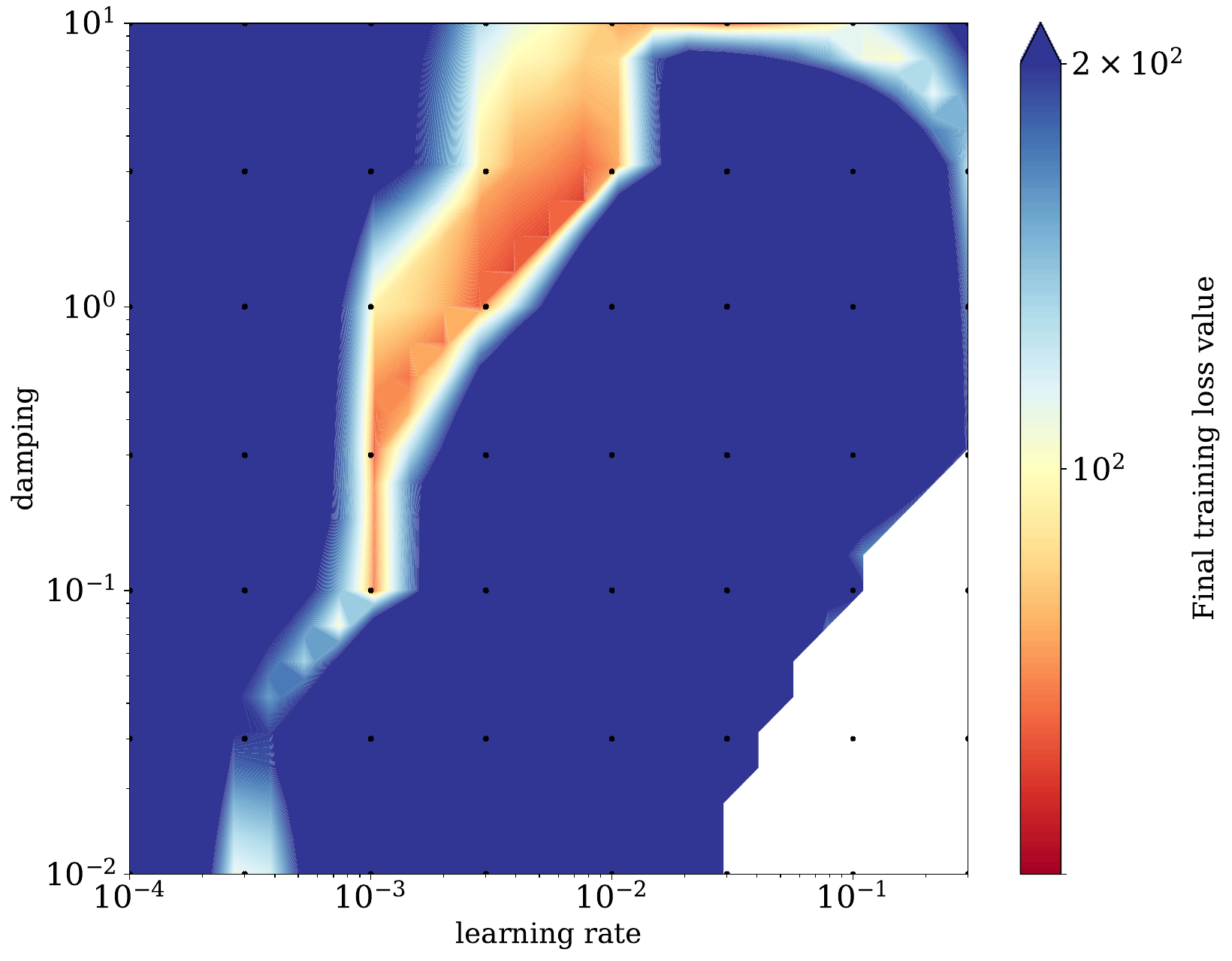}
\end{minipage}
\caption{
Landscape of the final training loss value w.r.t hyper-parameters (i.e. learning rate and damping) for KFAC. The left, middle, right columns depict results for MNIST, FACES, CURVES, which are terminated after 500, 2000, 500 seconds (CPU time), respectively.
}
\label{fig_hyper}
\end{figure}

\subsection{Training and testing plots:} 
\begin{figure}[H]
\centering
\subfigure[\footnotesize a) CIFAR-10, ResNet-32]{\includegraphics[width=0.45\textwidth]{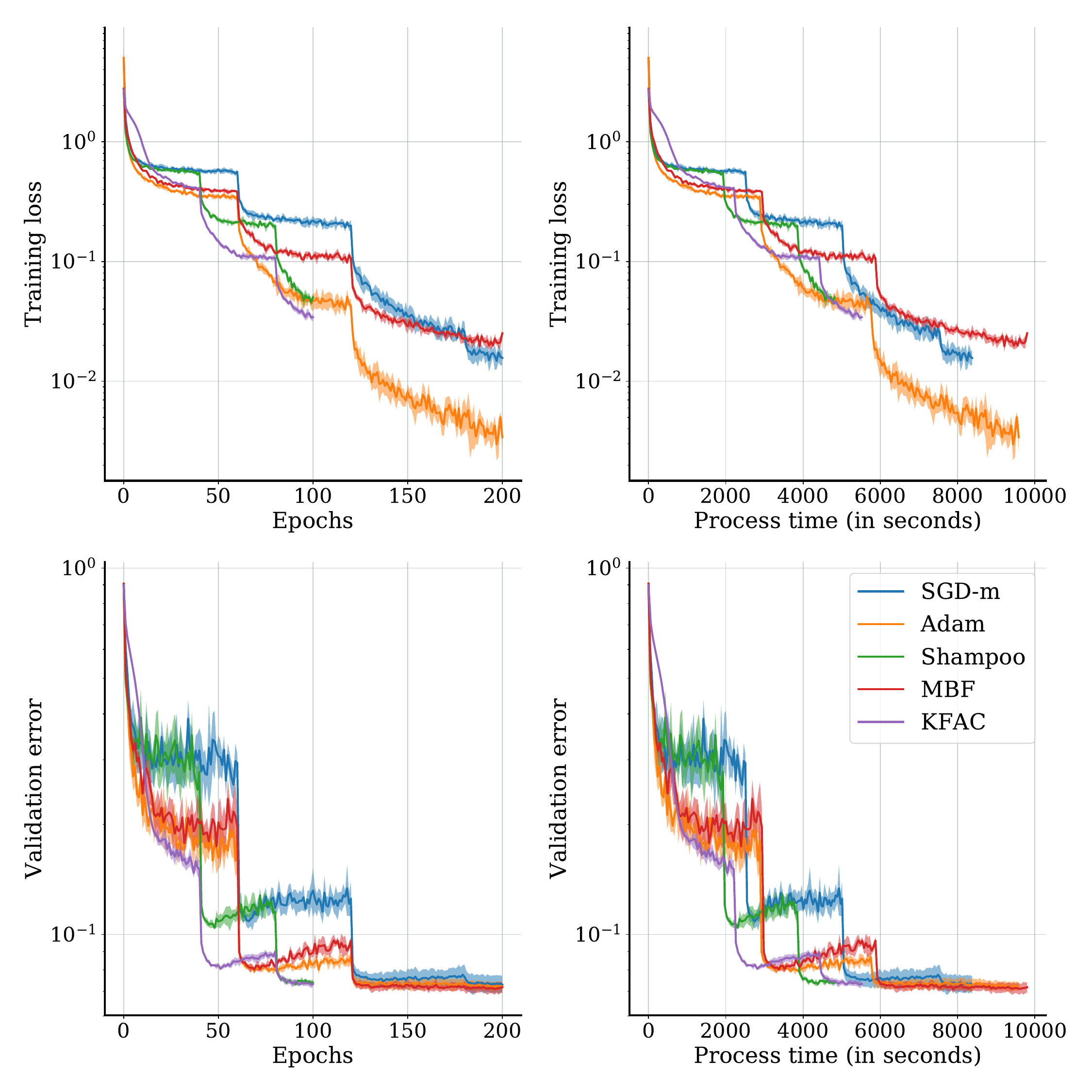}}\quad
  \subfigure[\footnotesize b) CIFAR-100, VGG16]{\includegraphics[width=0.45\textwidth]{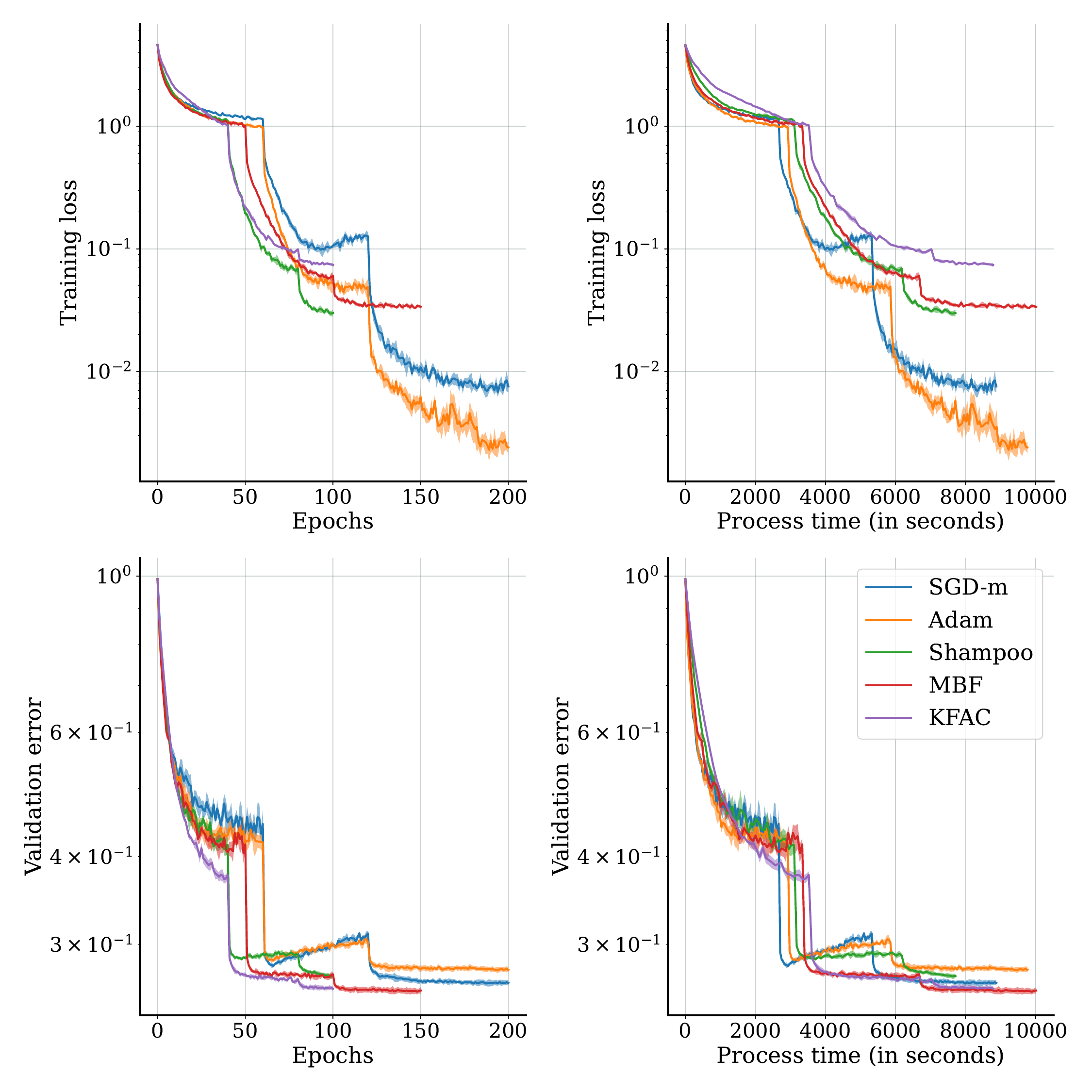}}\\
  \subfigure[\footnotesize c) SVHN, VGG11]{\includegraphics[width=0.45\textwidth]{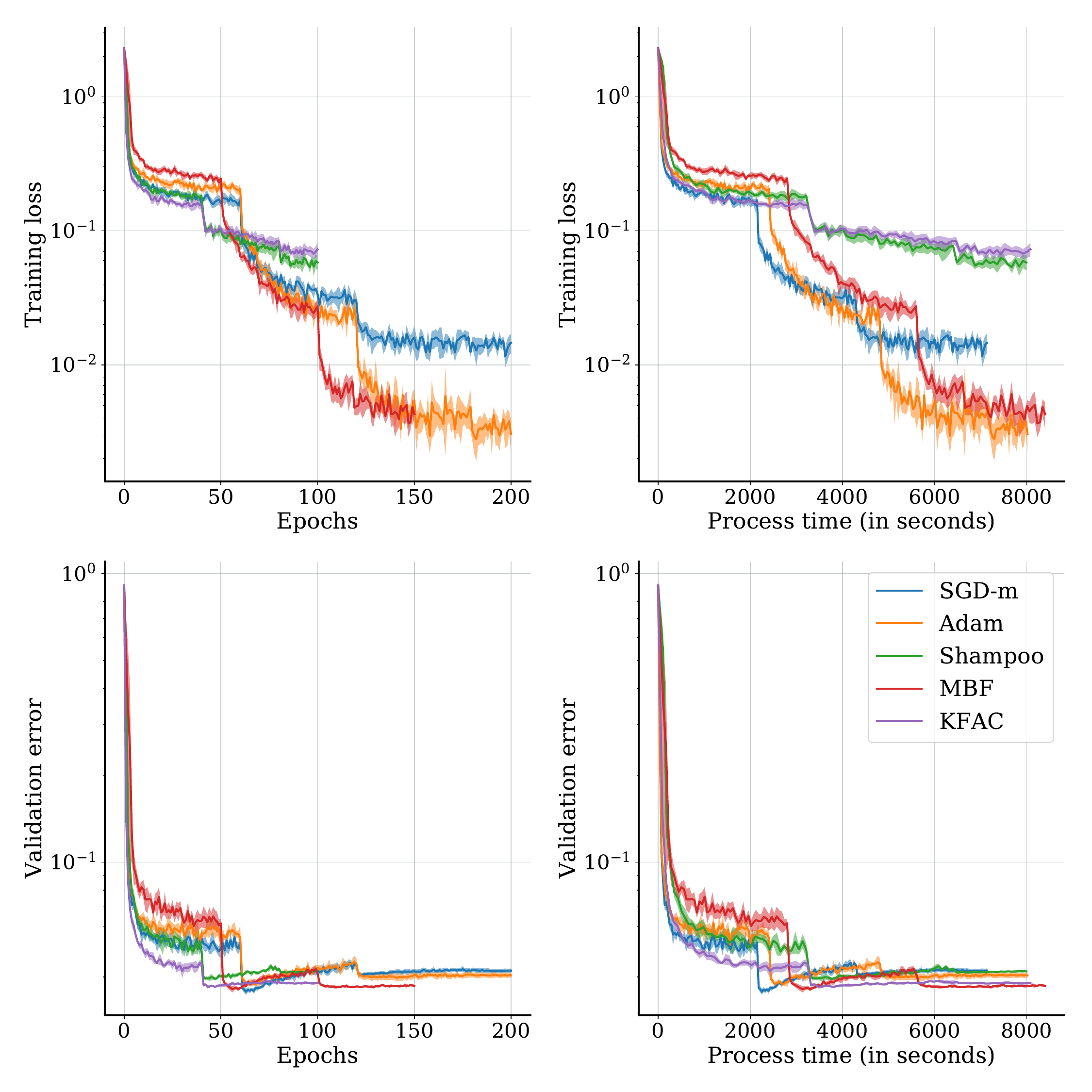}}
    \caption{
    Training and testing performance of MBF, KFAC, Shampoo, Adam, and SGD-m on three CNN problems.
    }
    \label{fig_cnns_add}
\end{figure}

\begin{figure}[H]
\centering
\subfigure[\footnotesize MNIST autoencoder]{\includegraphics[width=0.45\textwidth]{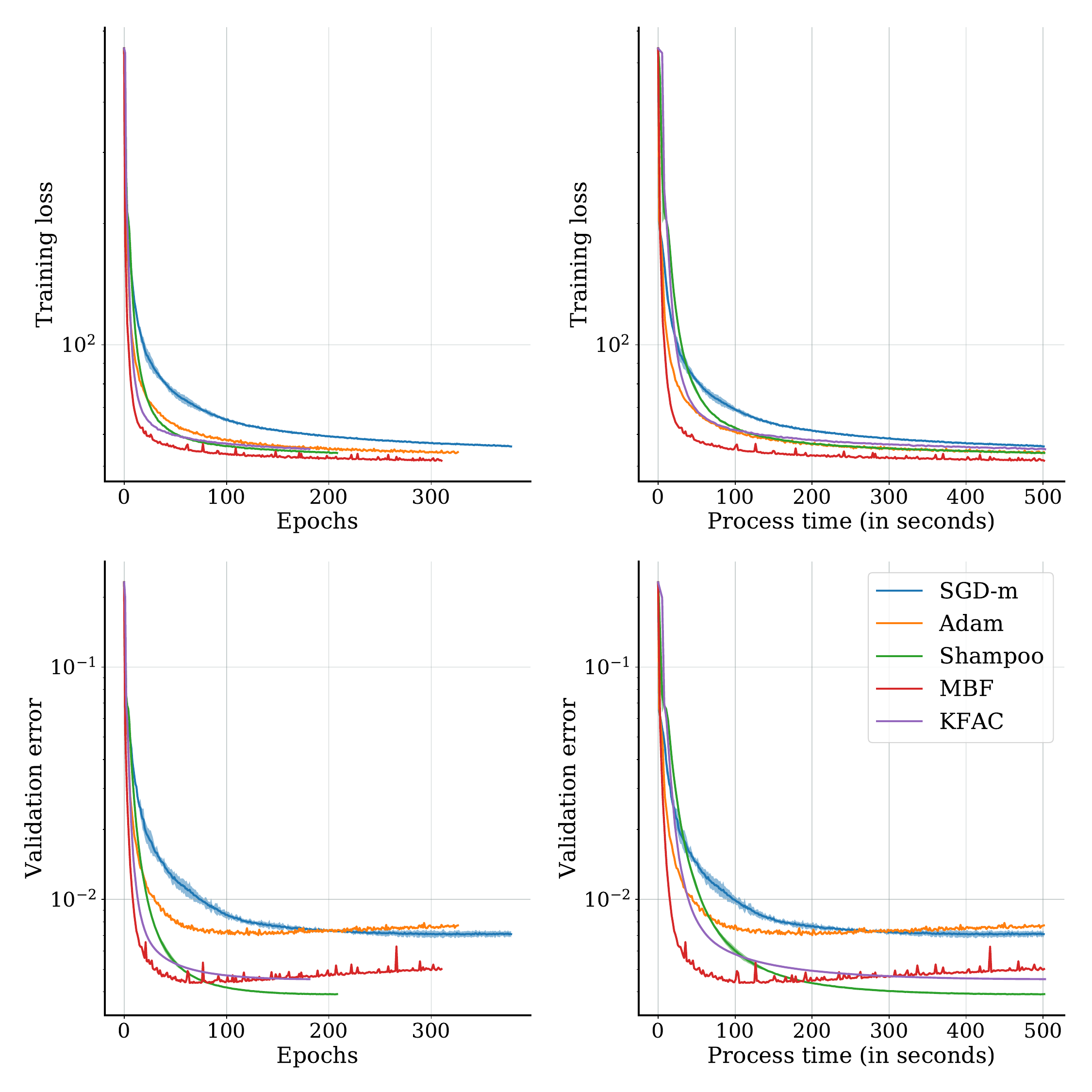}}\quad
  \subfigure[\footnotesize FACES autoencoder]{\includegraphics[width=0.45\textwidth]{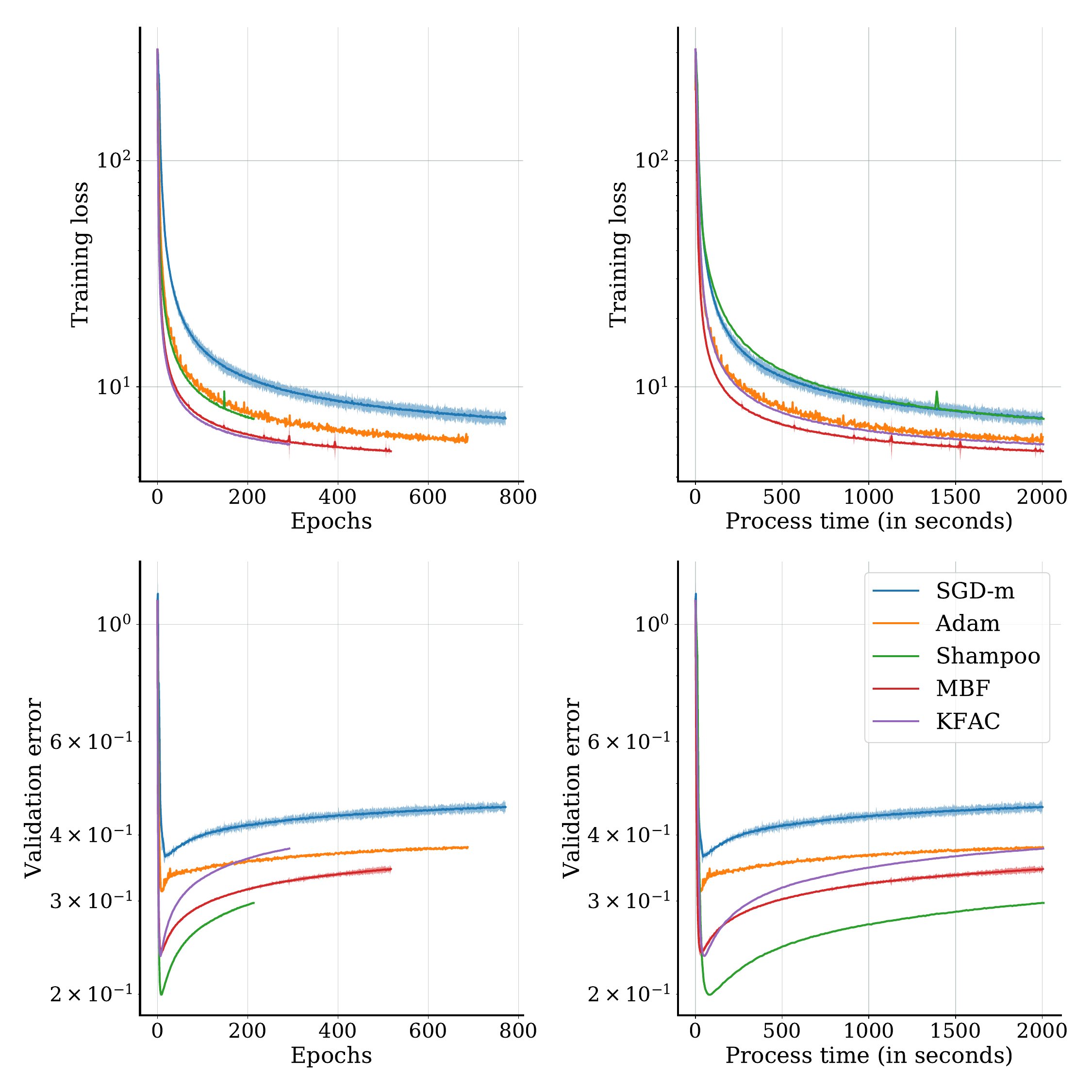}}\\
  \subfigure[\footnotesize CURVES autoencoder]{\includegraphics[width=0.45\textwidth]{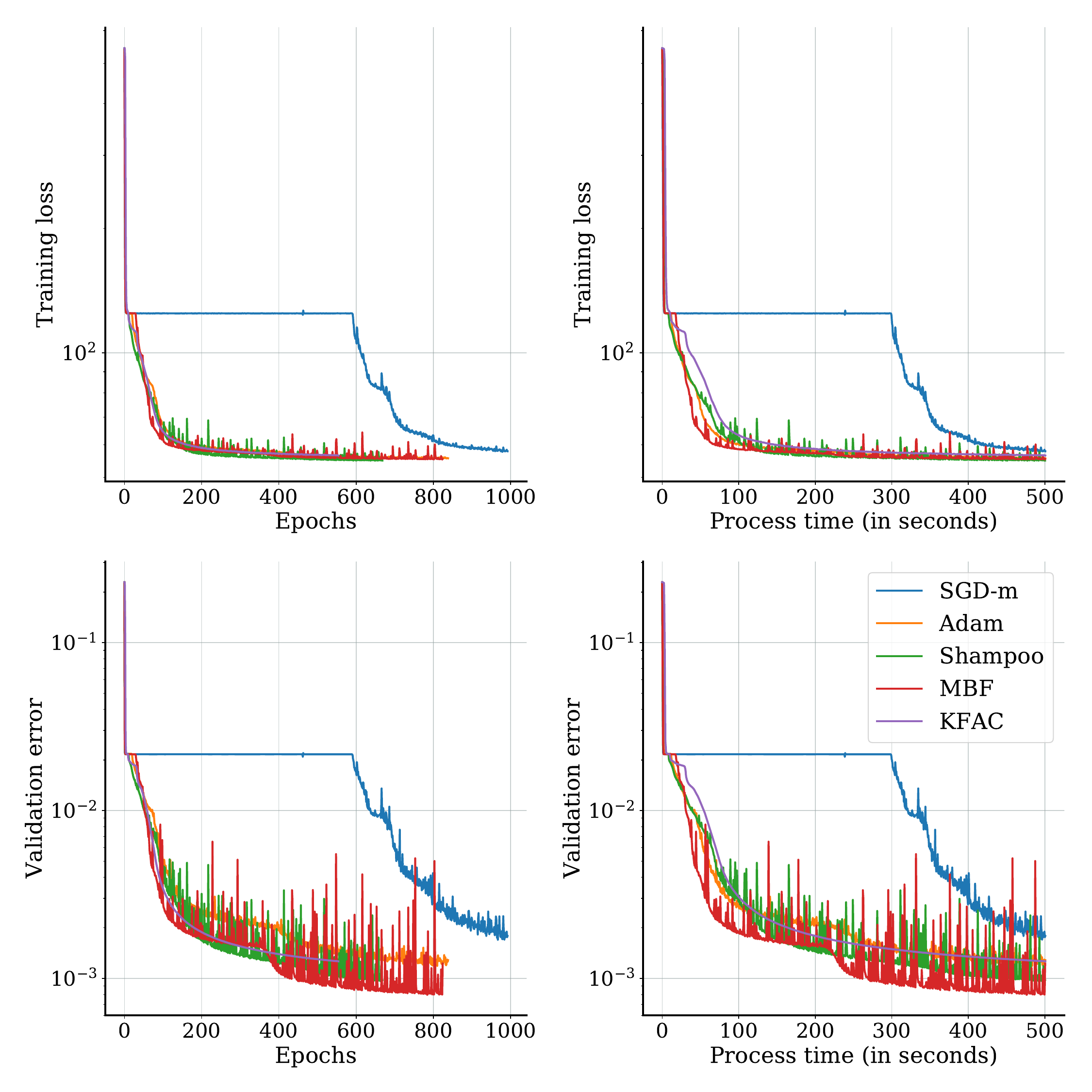}}
    \caption{
    Training and testing performance of MBF, KFAC, Shampoo, Adam, and SGD-m on three autoencoder problems.
    }
    \label{fig_autoencoders_add}
\end{figure}

For completeness, we report in Figures \ref{fig_cnns_add} and \ref{fig_autoencoders_add} both training and testing performance of the results plotted in Figures \ref{fig_cnns} and \ref{fig_autoencoders} in the main manuscript.
\newpage

\section{Limitations}\label{limitations}

We have explored using MBF in both Autoencoder and CNN problems. However, we believe it would be interesting to extend the method to other architectures such as RNNs and other sets of problems such as natural language processing (NLP) that predominately use Transformer models. It would also be interesting to extend our theoretical results to the fully stochastic case.

\end{document}